\documentclass{article} %
\usepackage{iclr2021_conference,times}

\usepackage{amsmath,amsfonts,bm}

\def\eqref#1{equation~\ref{#1}}

\def\1{\bm{1}}

\DeclareMathAlphabet{\mathsfit}{\encodingdefault}{\sfdefault}{m}{sl}
\SetMathAlphabet{\mathsfit}{bold}{\encodingdefault}{\sfdefault}{bx}{n}

\usepackage{subfig}
\usepackage{graphicx}
\usepackage{hyperref}
\usepackage{url}
\usepackage{algorithm}
\usepackage[noend]{algorithmic}
\usepackage{multicol}
\usepackage{nicefrac}
\usepackage{cleveref}

\title{Performance Weighting for Robust Federated Learning Against Corrupted Sources}

\author{Dimitris Stripelis \\
Information Sciences Institute \\
University of Southern California\\
Los Angeles, CA, USA \\
\texttt{\{stripeli\}@isi.edu}
\And
Marcin Abram \\
Department of Physics and Astronomy \\
\& Information Sciences Institute \\
University of Southern California\\
Los Angeles, CA, USA \\
\texttt{\{mjabram\}@usc.edu}
\And
Jos\'{e} Luis Ambite \\
Information Sciences Institute \\
University of Southern California\\
Los Angeles, CA, USA \\
\texttt{\{ambite\}@isi.edu}
}

\iclrfinalcopy %
\begin{document}

\maketitle

\begin{abstract}

Federated Learning has emerged as a dominant computational paradigm for distributed machine learning. Its unique data privacy properties allow us to collaboratively train models while offering participating clients certain privacy-preserving guarantees. However, in real-world applications, a federated environment may consist of a mixture of benevolent and malicious clients, with the latter aiming to corrupt and degrade federated model's performance. Different corruption schemes may be applied such as model poisoning and data corruption. Here, we focus on the latter, the susceptibility of federated learning to various data corruption attacks.
We show that the standard global aggregation scheme of local weights is inefficient in the presence of corrupted clients. To mitigate this problem, we propose a class of task-oriented performance-based methods computed over a distributed validation dataset with the goal to detect and mitigate corrupted clients. Specifically, we construct a robust weight aggregation scheme based on geometric mean and demonstrate its effectiveness under random label shuffling and targeted label flipping attacks.

\end{abstract}

\section{Introduction} \label{sec:Introduction}
Federated Learning is considered a standardized distributed machine learning paradigm due to its inherent privacy-preserving guarantees. In principal, federated learning relaxes the need to move the data out of their original location and instead pushes the training of the model down to the learning site \citep{mcmahan2017communication,yang2019federated}. Each learning site (learner) maintains a local model that is trained on its own private dataset and only shares the trained parameters with the rest of the federation. A trustworthy server is responsible to aggregate all the local models and compute the global (community) model.  

Although federated learning is privacy-preserving since it operates on aggregated parameters, it is susceptible to data corruption/poisoning attacks that can originate at the learning sites \citep{lyu2020threats,MAL-083}. Inadvertent alterations on the integrity of the local training data (e.g., human annotation errors, systematic mislabeling) or targeted adversarial data corruptions can severely damage the performance of the federation with dire consequences on the final outcome. To circumvent these challenges, new defensive mechanisms need to be developed that are capable to improve the resiliency of the federated learning systems against such attacks.

In this work, we present a new Performance Weighting scheme for robust federated learning in the presence of corrupted data sources that can be used both as a defensive as well as a detection mechanisms for a range of data poisoning attacks. We use the local validation dataset of each learner as a testbed to evaluate the performance of the local models and measure their weight in the community model. We empirically show that our weighting scheme is robust to data poisoning attacks even if majority of training and validation data are corrupted.

\section{Background \& Related Work} \label{sec:BackgroundRelatedWork}
The widespread adoption of federated learning has spurred research on adversarial attacks that aim to tamper with the learning process of the federation, such as backdoor attacks \citep{bagdasaryan2020backdoor,wang2020attack,xie2019dba,ananda2019can}, membership inference attacks \citep{truex2019demystifying,nasr2019comprehensive} and poisoning attacks. Poisoning attacks can be toxonomized into two subcategories, model poisoning and data poisoning. In model poisoning, the attacker modifies the learning process by introducing adversarial gradients and parameter updates \citep{wei2021covert,fang2020local,bhagoji2019analyzing,pillutla2019robust}, while in the case of data poisoning the attacker aims to manipulate the training data by directly modifying the training examples \citep{tolpegin2020data,fung2018mitigating,shen2016auror}.  

Our focus is on the data poisoning (corruption) attacks and specifically label flipping \citep{steinhardt2017certified,shen2016auror,xiao2015support,xiao2012adversarial,biggio2011support} and label shuffling \citep{li2021auto}. In our learning environments, we consider static adaptability \citep{MAL-083,pillutla2019robust}, where the data poisoning attack is considered fixed throughout the federated execution, as well as continuous participation rate where the attacker continuously participates in the learning process; a situation frequently observed in cross-silo federated settings.

It has been shown that in centralized environments, deep learning can be robust even to extreme degrees of corruption~\cite{Rolnick2017}, such as label shuffling exceeding 90\%. However, when considering a federated learning environment, the federation can be strongly affected by corruption when using standard aggregation schemes such as federated average (FedAvg). As we also show in this work, there is a substantial degradation of the community model's performance even at the presence of very low corruption levels (e.g., 30\% label shuffling). To address this problem, we are proposing a different aggregation scheme that empirically offers much improved robustness to various data corruption attacks.

Our proposed Performance Weighting scheme builds upon the work of \cite{stripelis2020accelerating}. Similar approaches were investigated by \cite{zhao2020shielding,wang2020model}. Specifically, in \cite{zhao2020shielding} the authors aggregate the local models into sub-models and delegate their evaluation to learners that have a similar data distribution with the aggregated model, while in \cite{wang2020model} the authors evaluate the local models against a validation dataset that is hosted at the central server. However, both proposed approaches used the validation-based accuracy as a detection mechanism to discard corrupted learners from the federation, whereas in our work we keep the corrupted models in the federation with a downgraded contribution value. We show empirically that by not discarding corrupted models during federated training (as in~\cite{tolpegin2020data}), our method leads to similar or even better performance than in the case of total exclusion of the corrupted models.

\section{Performance Weighting Mechanism}

In a federated learning environment with $N$ participating learners, the goal is to find a set of optimal model parameters by jointly optimizing the global objective function $f(w)=\sum_{k=1}^{N}\frac{p_k}{\mathcal{P}}F_k(w)$, with $F_k$ denoting the local objective function of learner $k$ weighted by factor $p_k$ and normalized to $[0,1]$ through the sum of all weighting factors, $\mathcal{P}=\sum_k^{N}{p_k}$. Every learner optimizes its local function $F_k$ by minimizing the empirical risk over its local training dataset, $D_{k}^{T}$.

In the case of Federated Average (FedAvg), the contribution value $p_k$ is equal to the size of the local training dataset $p_k = |D_{k}^{T}|$. However, in the case of Performance Weighting the contribution value is estimated by evaluating the performance of a learner's local model against a validation dataset. Similar to the distributed validation weighting scheme (DVW) proposed in \citet{stripelis2020accelerating}, we assign a performance score to the local model of each learner by evaluating its local model against the validation dataset of every other learner. In particular, every learner splits its local dataset into two disjoint datasets, a \textit{training} and a \textit{validation}, and reserves the validation dataset, $D_{k}^{V}$, throughout the federated execution for evaluating other learners' models. Note, that in FedAvg the learners train with all available examples, meaning $D_{k}^{V}=\emptyset$. Since in the case of Performance Weighting $D_{k}^{V} \neq \emptyset$, consequently the size of the training dataset is this case is necessarily smaller comparing to the FedAvg since the validation dataset never becomes part of the training. 
 
On the grounds that a federation may consist of learners with diverse amounts of data and the number of training examples per class may not be balanced, the validation dataset is assembled through~\emph{stratified sampling}. The validation dataset needs to accurately represent the training data distribution of all learners, and therefore if the dataset is generated by randomly sampling examples, then the prediction tasks (e.g., classes in classification, score ranges in regression) could be over- or under-represented and would not reflect the underlying training distribution of the learner. To address this, in our work, every learner constructs its local validation dataset by sampling 5\% out of the training examples from each class.

\begin{figure*}
  \begin{minipage}[htpb]{0.47\textwidth}
    \begin{algorithm}[H]
        \caption{\texttt{Performance Weighting.}}
        \label{alg:DVW}
        \begin{algorithmic}
            \renewcommand{\algorithmicrequire}{\textbf{Controller executes:}}
            \REQUIRE
            \FOR{$t = 0, \dots, T-1$}
                 \FOR{each learner $k \in N$}
                    \STATE $w_k = \textsc{ClientOpt}(w_c, \epsilon)$
                    \STATE $p_k = \textsc{Eval}(w_k)$
                 \ENDFOR
                 \STATE {$w_c = \sum_{k=1}^{N}\frac{p_{k}}{\mathcal{P}}w_k$ with $\mathcal{P}=\sum_{k}^N p_k$}
            \ENDFOR
            
            \STATE
            \renewcommand{\algorithmicrequire}{\textbf{\textsc{ClientOpt($w_t, \epsilon$):}}}
            \REQUIRE        
            \STATE $\mathcal{B} \leftarrow$ Split $\epsilon * D_k^{T}$ in batches of size $\beta$
            \FOR{$b \in \mathcal{B}$}
                \STATE {$w_{t+1} = w_{t} - \eta\nabla F_k(w_t;b)$}
            \ENDFOR
            \STATE Return $w_{t+1}$
            
            \STATE
            \renewcommand{\algorithmicrequire}{\textbf{\textsc{Eval($w$):}}}
            \REQUIRE
                \STATE $\mbox{\it CM} = 0_{C,C}$  \COMMENT{Confusion matrix CxC}
                \FOR{each learner $k \in N$ \textbf{in parallel}}
                    \STATE $\mbox{\it CM} = \mbox{\it CM} + \textsc{Evaluator}_{k}(w)$
                \ENDFOR
                \STATE $\textsc{Score} = {\mbox{\textit{fn}}}(\mbox{\it CM})$ %
                \STATE Return $\textsc{Score}$
        \end{algorithmic}
    \end{algorithm}
  \end{minipage}
  \hfill
  \begin{minipage}[htpb]{0.5\textwidth}
    \includegraphics[width=1\textwidth]{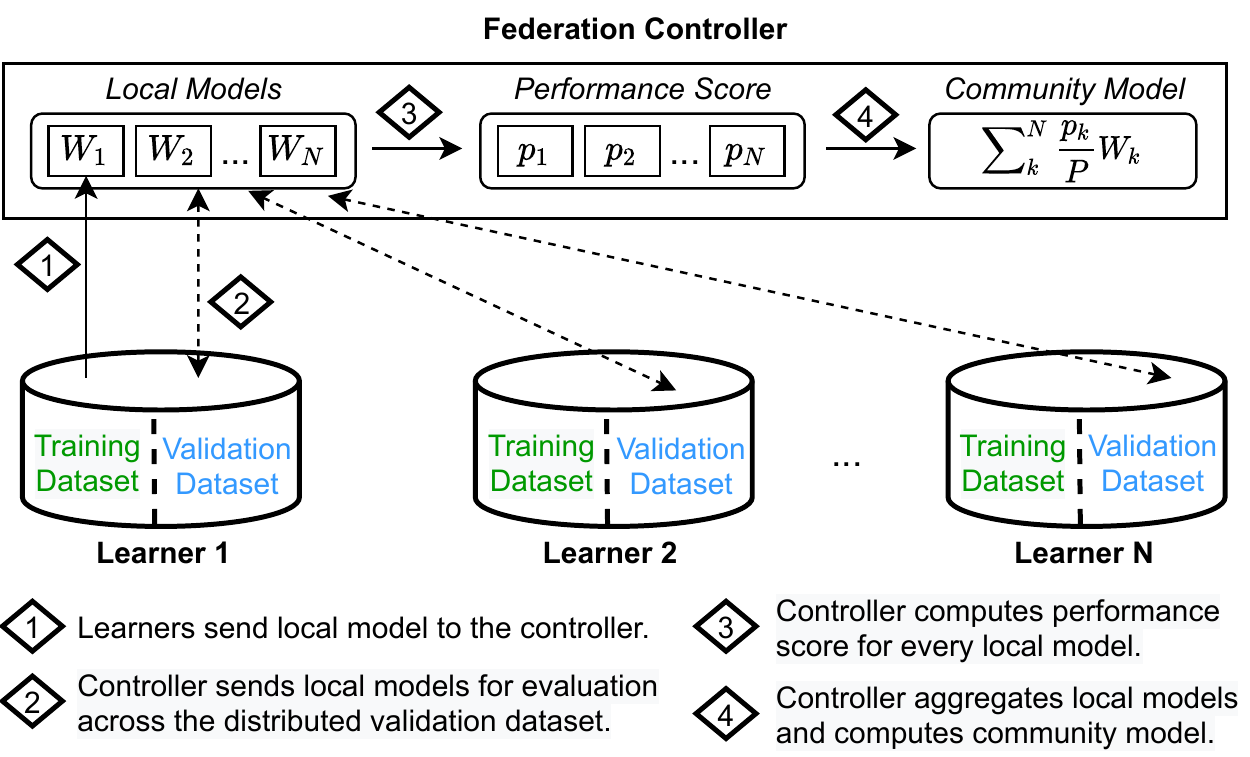}
    \caption{Execution pipeline of the performance weighting aggregation scheme. Initially, learners train locally on their local dataset, the controller receives the trained models and sends them for evaluation to the evaluation service of every participating learner. The controller aggregates all models based on their performance score and computes the new community model. With the computation of the new community model the new federation round begins.}
    \label{fig:DVW_ExecutionFLowDiagram}
  \end{minipage}
  \hfill
\end{figure*}

\paragraph{Execution Pipeline.} At the start of the federation all learners receive from the controller the original model state and train on their local dataset for an assigned number of iterations (see~\textsc{ClientOpt} procedure in Alg.~\ref{alg:DVW}; note that input parameter $\epsilon$ denotes local epochs). Upon completing their local training learners send their model back to the controller and request a community update. When the controller receives the local model ($w_k$) from all learners (i.e., synchronous execution), it aggregates all local models to compute the new global (community) model ($w_c$) and sends it back to the learners to continue training. When Performance Weighting is applied, prior to aggregating the local models, the controller evaluates every local model against the validation dataset of every participating learner in order to determine the performance weighting factor of each individual model (see~\textsc{Eval} procedure in Alg.~\ref{alg:DVW}). To accomplish this, the controller sends the local model of each learner to the evaluator service of every other learner and accumulates the respective evaluation metrics from all services. Currently, we focus only on classification tasks (not regression) and the controller combines the confusion matrices from every evaluator service into a cumulative confusion matrix. We also assume that every learner truthfully evaluates the received models. Thus, we operate in the regime of "forgetful but honest" participants.

\paragraph{Performance Scores.} Once the final matrix is accumulated, the controller can compute a range of classification metrics (see \textit{fn(CM)} in Alg.\ref{alg:DVW}) \citep{ferri2009experimental}. We are currently evaluating Micro- and Macro-Average Accuracy, as well as Geometric Mean. Specifically, Micro-Accuracy is defined as $DVW_{acc}^{\mu}=\frac{TP_\mathrm{C}}{\#Examples}$, with $TP_\mathrm{C}$ denoting the total number of true positives over the validation examples of all classes $C$. Macro-Accuracy is defined as $DVW_{acc}^{M}=\frac{\sum_{i}^{C}a_i}{C}$, with $a_i$ denoting the accuracy of class $i$, which is equal to $a_i=\frac{TP_i}{m_i}$ with $TP_i$ and $m_i$ being the total number of true positives and validation examples of class $i$, respectively. Similarly, for multi-class evaluation, the Geometric Mean is computed as the geometric average of the partial accuracy of each class, $DVW^{GMean}=\sqrt[C]{\Pi_{i}^{C} a_{i}}$. In cases where learners might not have training examples for a particular class then the accuracy value for this class is going to be 0 and as a result the Geometric Mean value will also be 0. To account for these learning scenarios, we add a small arbitrary correction value, $\varepsilon=0.001$.
For instance, if for three classes $\{0,1,2\}$ a classifier has respective accuracies $\{0.76, 0.84, 0\}$, then its Geometric Mean value will be equal to $\sqrt[3]{0.76 * 0.84 * 0.001}=0.086$.

\section{Attack Modes}
We explore two different data poisoning attacks, label shuffling and targeted label flipping, similar to the works of \citet{li2021auto,tolpegin2020data}. In the case of label shuffling, the labels of the local dataset of a learner are randomly scrambled, while targeted label flipping attack refers to the case where a corrupted learner flips the label of examples that correspond to a particular source class to a target class. The goal of both attacks is to tamper the training process of the federation. In our learning environments, we poison both training and validation examples and we evaluate the effectiveness of the attacks on IID data distributions over both homogeneous (Uniform) and heterogeneous (PowerLaw) amounts of data across learners. The success of each attack is computed on the top-1 accuracy of the community model on the test set. Uniform refers to the case where learners have an equal number of local examples, while PowerLaw refers to the case where the number of examples is rightly skewed. We also investigate the contribution value of each learner in the federation for each performance weighting scheme, as well as the performance per class against the examples of the distributed validation dataset. In our current work, we explore the robustness of the community model against the two attack modes on the CIFAR-10 domain over an increasing number of corrupted learners. In our experiments, we consider 10 learners in total. Details regarding the federated model execution and the experiments for every attack in every environment can be found in \cref{appendix:FederatedModelExecution,appendix:uniform_label_shuffling,appendix:targeted_label_flipping}.

\begin{figure*}[t]
\begin{minipage}{1\textwidth}
  \captionsetup[subfigure]{justification=centering}
  \subfloat[Label Shuffling \newline Uniform \& IID]{
    \includegraphics[width=.482\textwidth]{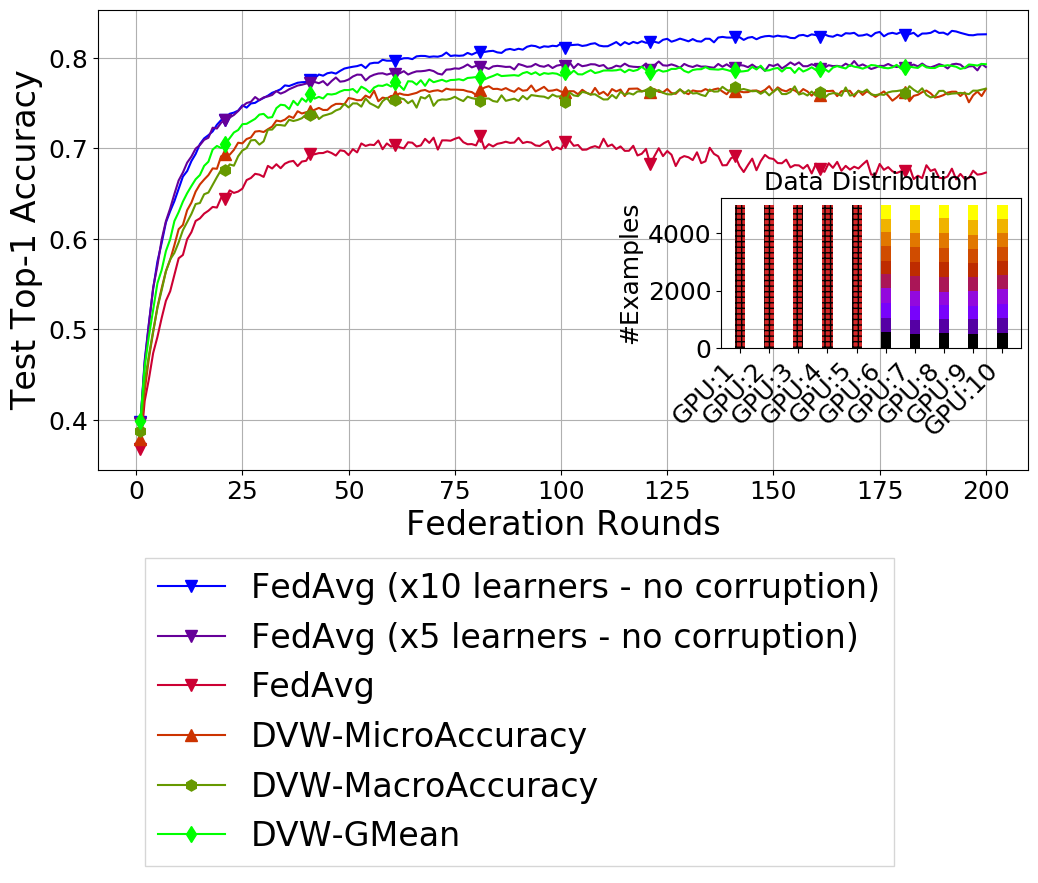}
    \label{subfig:MainPaper_Cifar10_UniformIID_UniformLabelShuffling_5Learners_FederationConvergence}
  }
  \subfloat[Label Shuffling \newline PowerLaw \& IID]{
    \includegraphics[width=.482\textwidth]{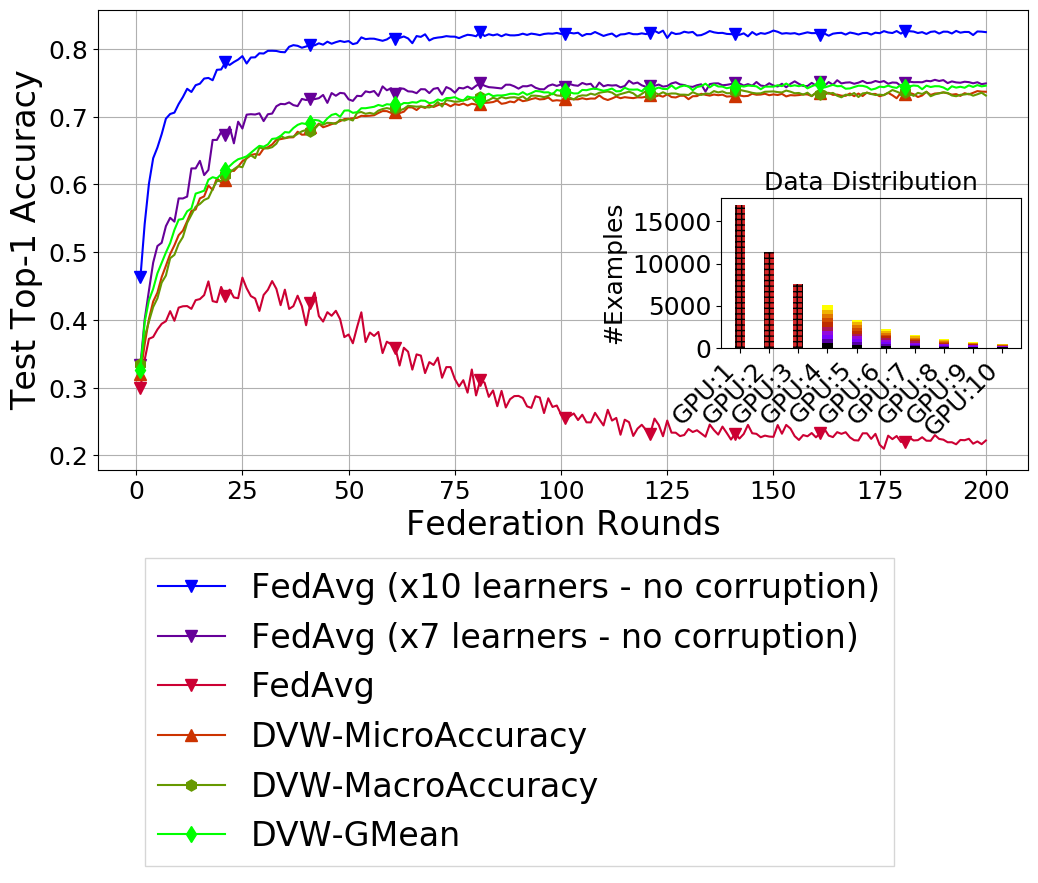}
    \label{subfig:MainPaper_Cifar10_PowerLawIID_UniformLabelShuffling_5Learners_FederationConvergence}
  }
  
  \subfloat[DVW-GMean \newline (5 corrupted learners)]{
    \includegraphics[width=.482\textwidth]{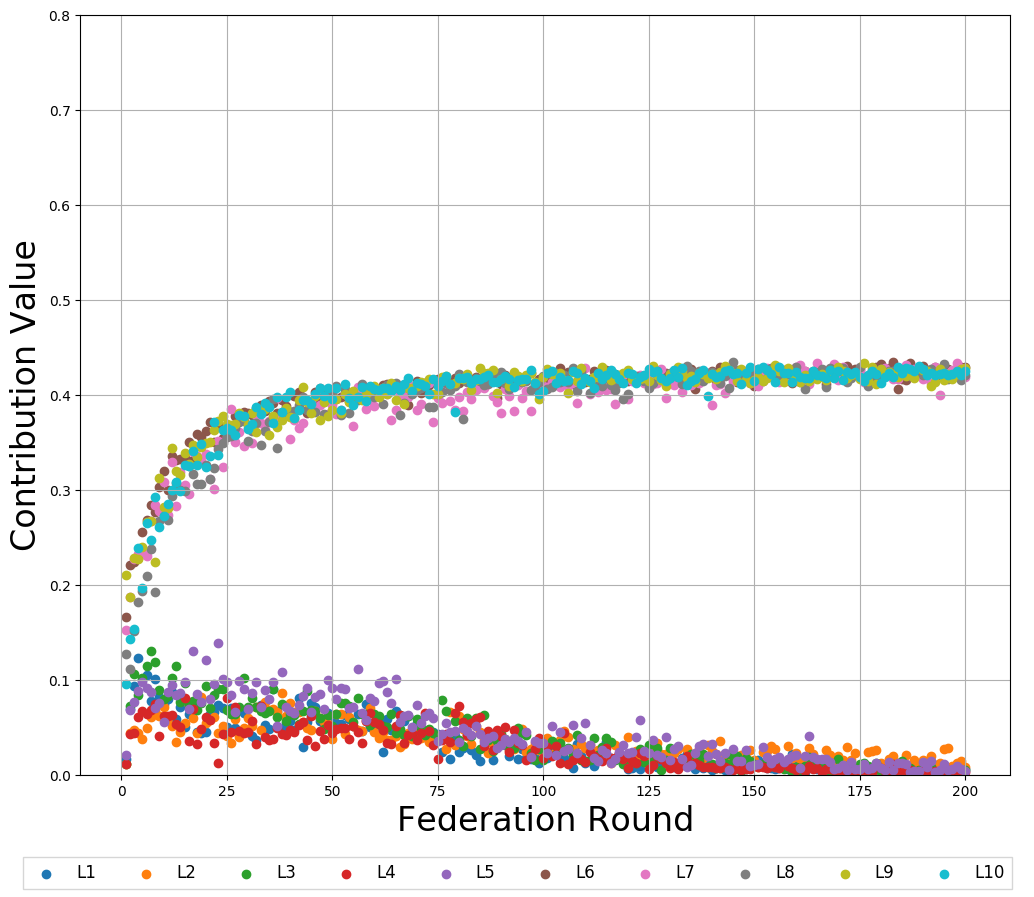}
    \label{subfig:MainPaper_Cifar10_UniformIID_UniformLabelShuffling_5Learners_ContributionValue_DVWGMean}
  }
  \subfloat[DVW-GMean \newline (3 corrupted learners)]{
    \includegraphics[width=.482\textwidth]{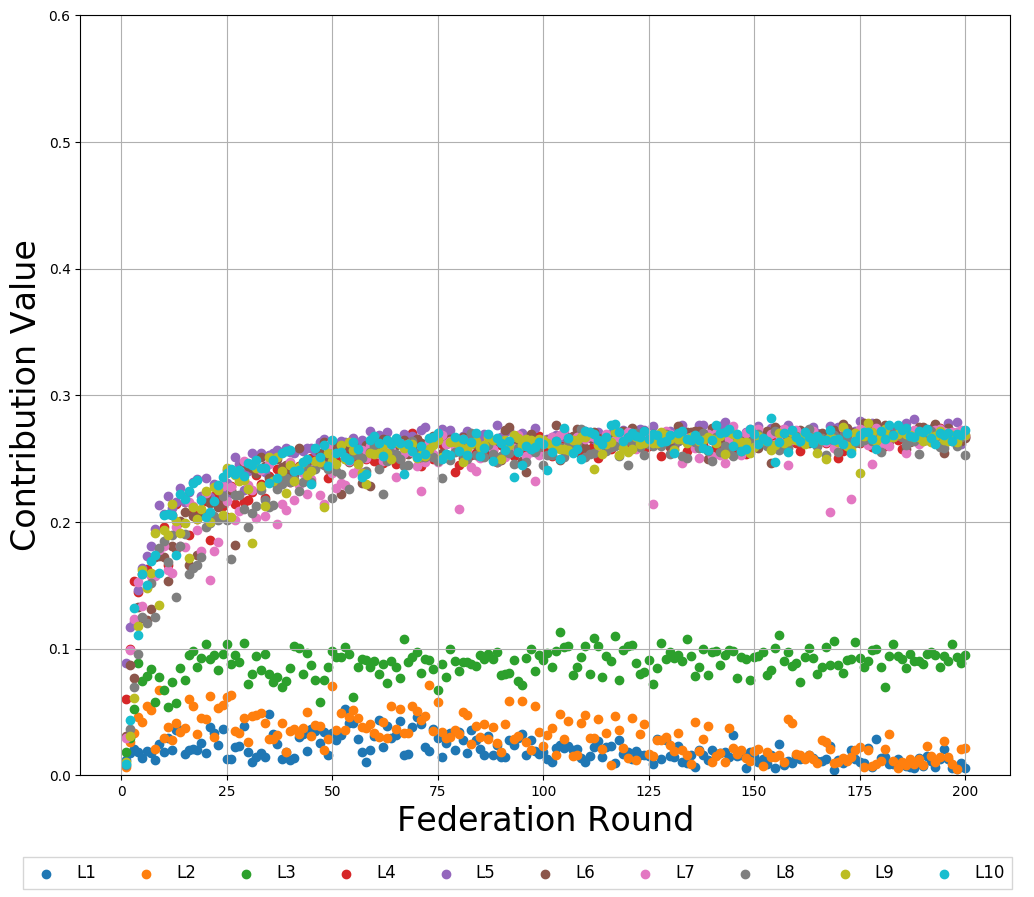}
    \label{subfig:MainPaper_Cifar10_PowerLawIID_UniformLabelShuffling_3Learners_ContributionValue_DVWGMean}
  }

  \caption{Top row presents the convergence of the federation for the Label Shuffling attack mode and bottom row presents the corresponding per-learner contribution value for the Geometric Mean weighting scheme.}
\end{minipage}
\end{figure*}

\paragraph{Uniform Label Shuffling} For every corrupted learner, we shuffle the labels of its local training dataset based on a discrete uniform distribution, $\mathcal{U}\{0,9\}$. For the \textit{Uniform \& IID} learning environment, we investigate five different degrees of corruption, i.e. $\{1,3,5,6,8\}$ corrupted learners or $\{10\%, 30\%, 50\%, 60\%, 80\%\}$ totally corrupted data. Here, we only show the federation convergence for 5 corrupted learners (Figure \ref{subfig:MainPaper_Cifar10_UniformIID_UniformLabelShuffling_5Learners_FederationConvergence}) and the per-learner contribution value for the Geometric Mean scheme (Figure \ref{subfig:MainPaper_Cifar10_UniformIID_UniformLabelShuffling_5Learners_ContributionValue_DVWGMean}).

As we can see, our proposed Geometric Mean scheme (DVW-GMean) outperforms the standard Federated Average (FedAvg) by a large margin. DVW-GMean outperforms also all tested alternative weighting schemes, namely DVM-MicroAccuracy and DVW-MacroAccuracy. Note, that ``FedAvg (x10 learners -- co corruption)'' indicates the case without any corruption, and is shown here only to illustrate the overall impact of the data poisoning attack. ``FedAvg (x5 learners -- co corruption)'' indicates a case where we manually removed all corrupted learner and we leave only those with unaffected datasets. This represents the case where a corruption detection mechanism (e.g., PCA~\cite{tolpegin2020data}) is used to exclude corrupted learners from the federation at the beginning of federated training. As we see, our Geometric Mean scheme approaches ``FedAvg (x5 learners -- co corruption)''results quickly, converging after 100 federation rounds. This demonstrates the resilience of our aggregation scheme to label shuffling attacks. 

We included a detail analysis of the remaining cases (e.g., different corruption levels) in ~\cref{appendix:uniform_label_shuffling_uniform_iid}.

In the case of the \textit{Power Law \& IID} environment due to the data amount heterogeneity across learning sites, we corrupt learners starting from the head of the data distribution and going rightwards; with the head owning the majority of the training examples. We consider only three corruption levels, i.e. $\{1,3,5\}$ corrupted learners or $\{33\%, 71\%, 88\%\}$ totally corrupted data. Here, we show the federation convergence for 3 corrupted learners (Figure \ref{subfig:MainPaper_Cifar10_PowerLawIID_UniformLabelShuffling_5Learners_FederationConvergence}) and the contribution value per learner for the Geometric Mean scheme (Figure \ref{subfig:MainPaper_Cifar10_PowerLawIID_UniformLabelShuffling_3Learners_ContributionValue_DVWGMean}).

Here, the difference between the proposed DVW-GMean and FedAvg is even greater. Using standard FedAvg, even though it is initially performing, it quickly degrades as training progresses to a test accuracy close to 0.25, which is only marginally better than choosing classes at random. In contrast, DVW-GMean again quickly converges to the level of ``FedAvg (x5 learners -- co corruption)'', showing that the only effect of the corruption is only a moderate increase of the training time.

We included a detail analysis of the remaining cases in~\cref{appendix:uniform_label_shuffling_powerlaw_iid}, where we also included detail confusion matrices.

\begin{figure*}[t]
\begin{minipage}{1\textwidth}
  \captionsetup[subfigure]{justification=centering}
  \subfloat[Label Flipping \newline Uniform \& IID]{
    \includegraphics[width=.482\textwidth]{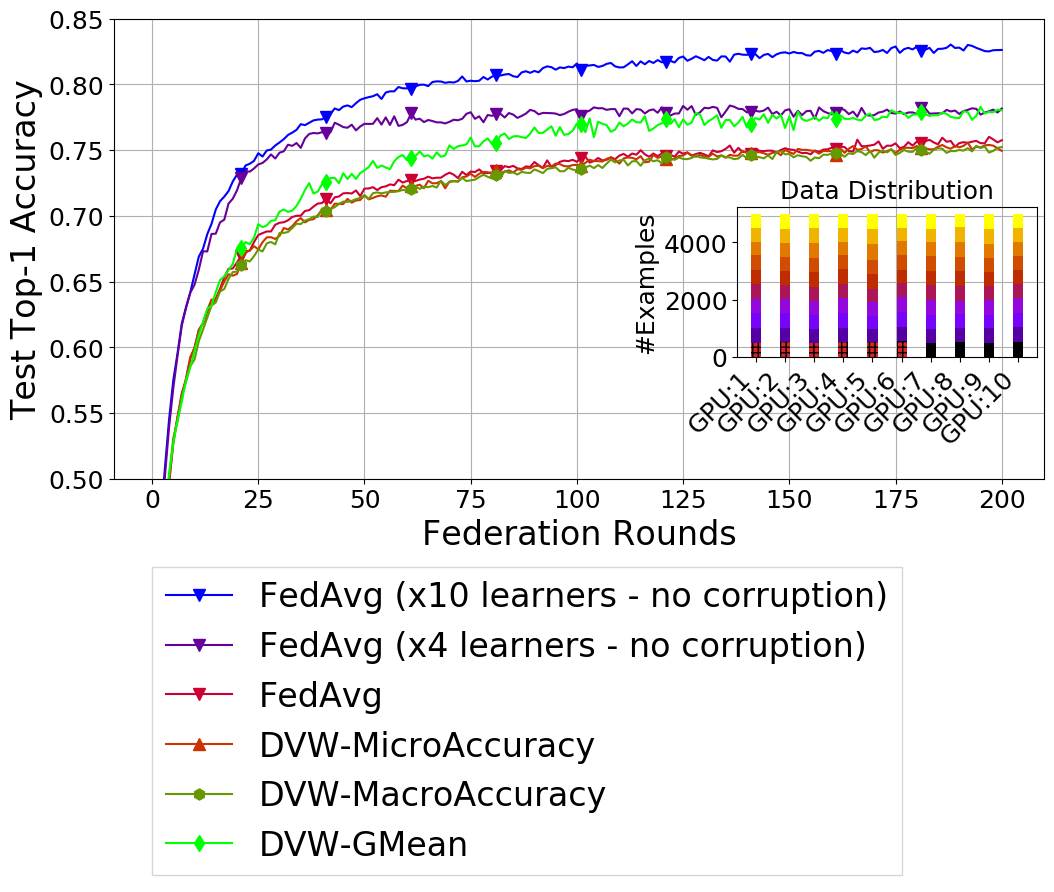}
    \label{subfig:MainPaper_Cifar10_UniformIID_TargetedLabelFlipping_6Learners_FederationConvergence}
  }
  \subfloat[Label Flipping \newline PowerLaw \& IID]{
    \includegraphics[width=.482\textwidth]{plots/TargetedlabelFlipping/PoliciesConvergence/Cifar10_PowerLawIID_TargetedLabelFlipping0to2_3Learners_L1L2L3_VanillaSGD_PoliciesConvergence}
    \label{subfig:MainPaper_Cifar10_PowerLawIID_TargetedLabelFlipping_3Learners_FederationConvergence}
  } 
  
  \subfloat[DVW-GMean \newline (6 corrupted learners)]{
    \includegraphics[width=.482\textwidth]{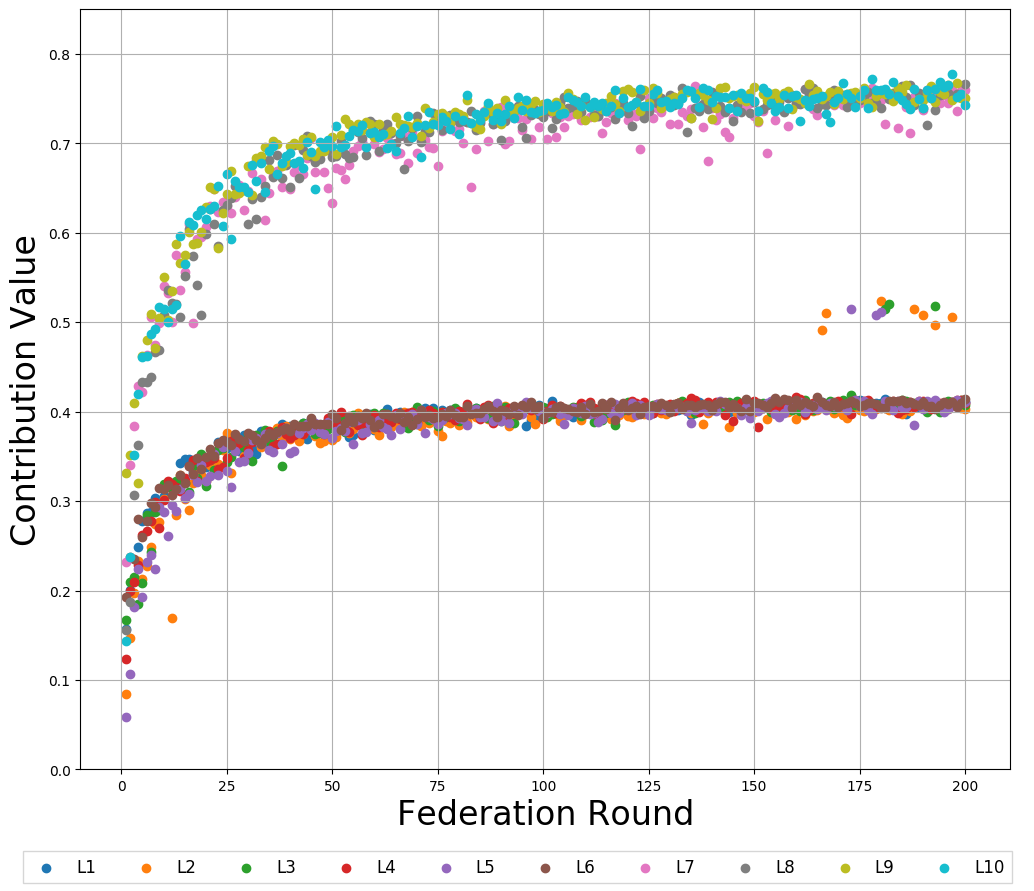}
    \label{subfig:MainPaper_Cifar10_UniformIID_TargetedLabelFlipping0to2_6Learners_ContributionValue_DVWGMean}
  }
  \subfloat[DVW-GMean \newline (3 corrupted learners)]{
    \includegraphics[width=.482\textwidth]{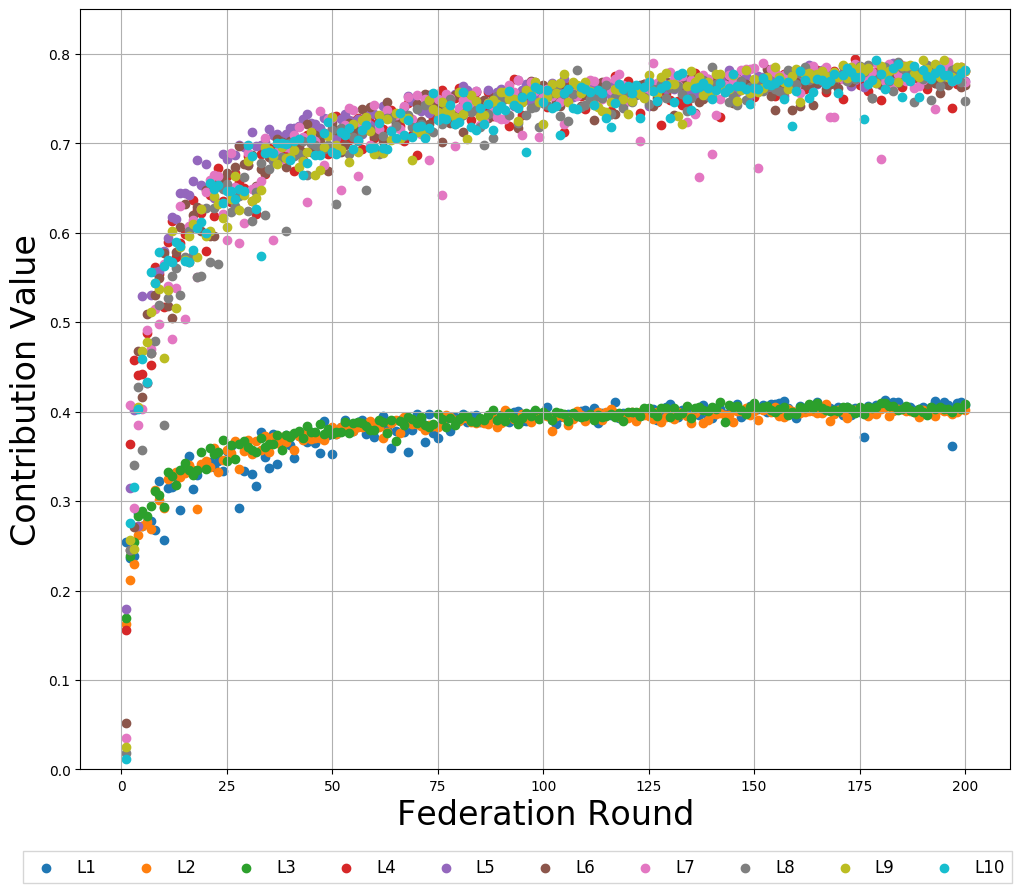}
    \label{subfig:MainPaper_Cifar10_PowerLawIID_TargetedLabelFlipping0to2_3Learners_ContributionValue_DVWGMean}
  }  
  \caption{Top row presents the convergence of the federation for the Label Flipping attack mode and bottom row presents the corresponding per-learner contribution value for the Geometric Mean weighting scheme.}
\end{minipage}
\end{figure*}

\paragraph{Targeted Label Flipping} In our learning environments, we inspect the performance of the federation when flipping airplane examples to birds ($0 \rightarrow 2$) over an increasing number of corrupted learners. We selected this corruption since airplanes and birds are relatively close in the feature space and hence are harder to detect. For the \textit{Uniform \& IID} environment, we investigate five different degrees of corruption, i.e. $\{1,3,5,6,8\}$ corrupted learners or $\{10\%, 30\%, 50\%, 60\%, 80\%\}$ class-degree corruption. In Figs. \ref{subfig:MainPaper_Cifar10_UniformIID_TargetedLabelFlipping_6Learners_FederationConvergence} and \ref{subfig:MainPaper_Cifar10_UniformIID_TargetedLabelFlipping0to2_6Learners_ContributionValue_DVWGMean} we show the corruption for 6 learners and the contribution value of each learner in the federation for the Geometric Mean scheme.

As we can see, our proposed Geometric Mean scheme (DVW-GMean) outperforms all alternative schemes, including the standard Federated Average (FedAvg). Similar to the cases discussed earlier, DVW-GMean converges to the level represented by ``FedAvg (x5 learners -- co corruption)'', indicating a very good resilience to the corruption attack. The only visible effect of the corruption is a slightly slower convergence. While ``FedAvg (x5 learners -- co corruption)'' levels off after about 100 federation rounds, DVW-GMean needs about 200 rounds to achieve the same accuracy level.

We included analysis of the remaining cases in~\cref{appendix:targeted_label_flipping_uniform_iid}.

In the case of \textit{PowerLaw \& IID}, given the data amount heterogeneity, we corrupt learners starting from the head of the data distribution. We consider three corruption levels, i.e. $\{1,3,5\}$ corrupted learners or $\{34\%, 72\%, 88\%\}$ class-degree corruption. In Figs. \ref{subfig:MainPaper_Cifar10_PowerLawIID_TargetedLabelFlipping_3Learners_FederationConvergence} and \ref{subfig:MainPaper_Cifar10_PowerLawIID_TargetedLabelFlipping0to2_3Learners_ContributionValue_DVWGMean} we present the results for 3 corrupted learners.

In this case, DVW-GMean not only outperforms all alternative schemes, but also outperforms the the smaller federation of 5 learners with no corruption, ``FedAvg (x5 learners -- co corruption)''. This phenomenon demonstrates that our scheme is more beneficial compared to detecting and excluding corrupted sources~\cite{tolpegin2020data}, since it can leverage the information stored in the corrupted sources to train a model of increased robustneness and performance.

The analysis of the remaining cases are included in~\cref{appendix:targeted_label_flipping_powerlaw_iid}.

\section{Discussion \& Future Work}

In our work we have shown that the standard Federation Average aggregation scheme is sensitive to data corruption attacks. We have proposed the Performance Weighting mechanism as an alternative method and we have shown that an aggregation scheme based on geometric mean has much improved resilience to random label shuffling and targeted label flipping. We have also shown that using our scheme we still benefit from the partially corrupted sites, achieving in some situations better results than just excluding them from the federation.

So far, in our work we have focused on IID learning environments. Currently we are working to extend our Performance Weighting mechanism to non-IID environments. We are also investigating how to evaluate the learners' local models in a privacy-preserving manner to protect from any potential information leakage and prevent adversaries from crafting more sophisticated attacks. For instance, Gaussian noise or weight precision trimming could be two promising protective mechanisms. As it was also experimentally shown in \citet{zhao2020shielding}, by adding a small variance on the model weights, the accuracy of the model can be preserved. Finally, we are also looking into approaches that can reduce the number of evaluation requests (models exchanged) required for assigning a performance score to every local model by sampling a subset of the available learners based on the importance of their local dataset, \citet{wang2019measure}.

\section{Acknowledgements}
This research was supported in part by the Defense Advanced Research Projects Activity (DARPA) under contract HR0011\-2090104, and in part by the National Institutes of Health (NIH) under grants U01AG068057 and RF1AG051710.  The views and conclusions contained herein are those of the authors and should not be interpreted as necessarily representing the official policies or endorsements, either expressed or implied, of DARPA, NIH, or the U.S. Government.

\bibliography{main}
\bibliographystyle{iclr2021_conference}

\appendix
\section{Federated Learning Environment Data Distribution}
In our experiments we expect all learners to hold data from all target classes (IID) but with dissimilar amounts of data. Specifically, we consider two assignment cases\footnote{The data partitioning script can be found at: \url{https://www.dropbox.com/s/71pml8nne0g5q2m/data_partitioning_ops.py?dl=0}}. In the first case, \textit{Uniform}, the learners have an equal number of examples, while in the second case, \textit{PowerLaw}, the number of examples is rightly skewed. More precisely, the total number of training examples per learner in the Uniform \& IID case is 5000, while in the PowerLaw and \& IID case, the exact number of data points per learner is distributed as follows: $\{L1:16964, L2:11314, L3:7537, L4:5023, L5:3348, L6:2232, L7:1488, L8:992, L9:661, L10:441\}$.

\begin{figure}[htpb]
\centering
  \subfloat[Uniform \& IID]{
    \includegraphics[width=0.45\linewidth]{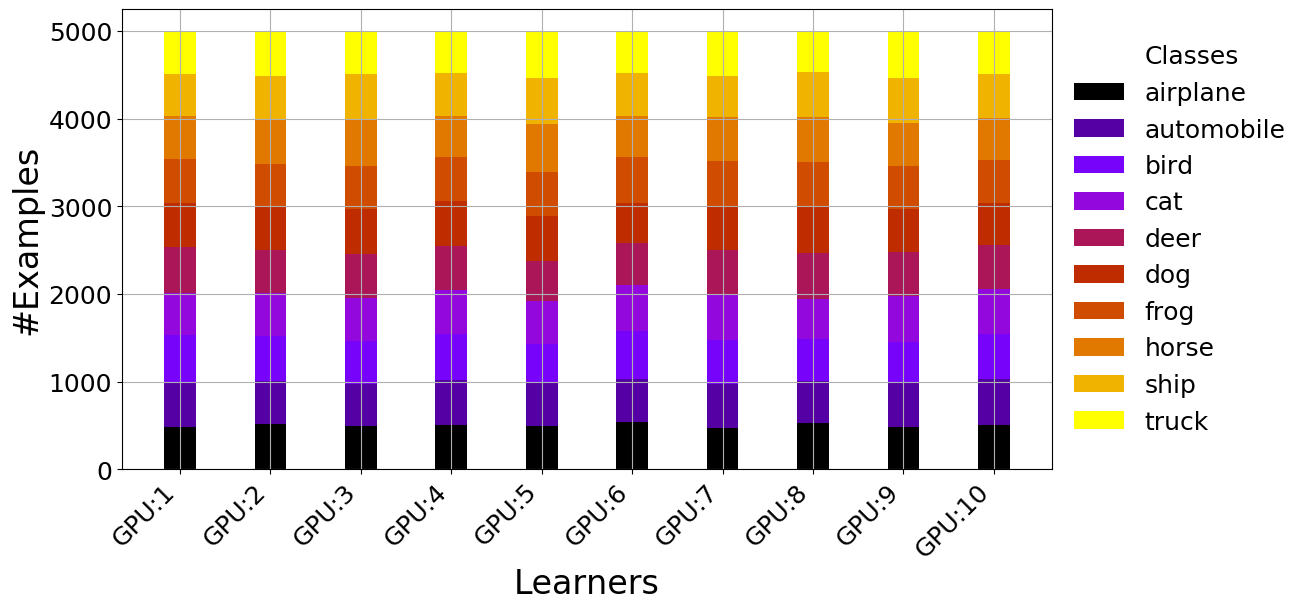}
    \label{subfig:cifar10_uniformiid_datadistribution_withclasses}
  }
  \subfloat[PowerLaw \& IID]{
    \includegraphics[width=0.45\linewidth]{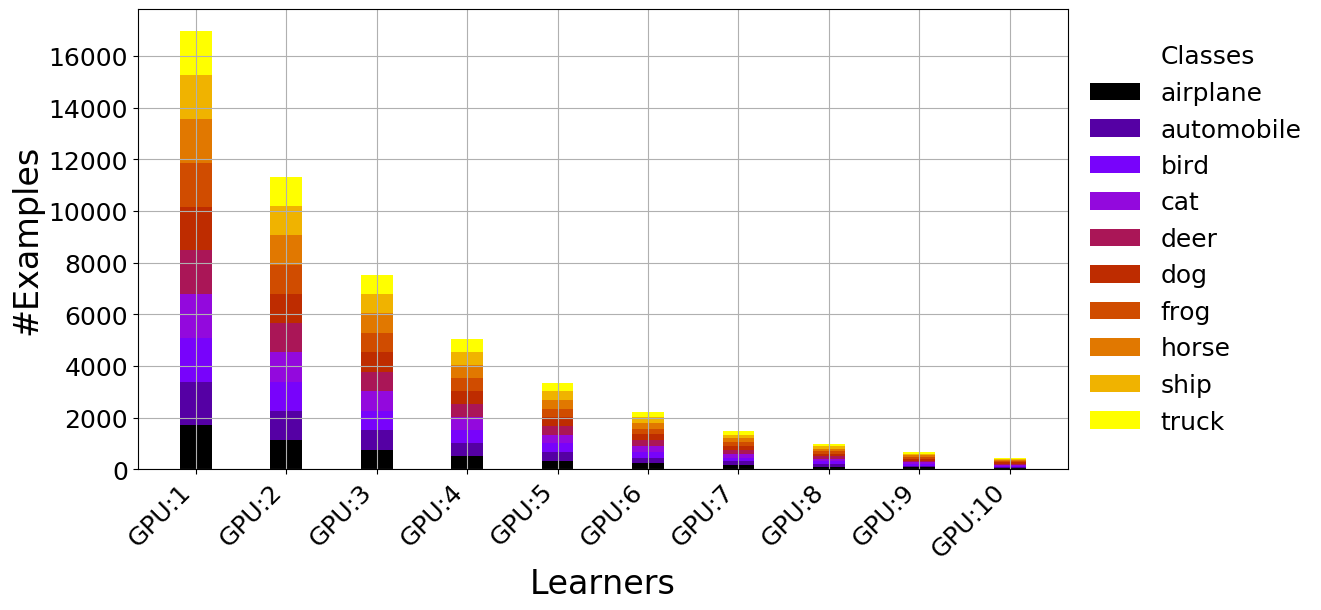}
    \label{subfig:cifar10_powerlawiid_datadistribution_withclasses}
  }
  
  \caption{CIFAR-10 federated learning environments (no corruption).}
  \label{fig:cifar10_learningenvironmetns_datadistributions}
\end{figure}

\section{Federated Model Execution} \label{appendix:FederatedModelExecution} 
For CIFAR-10 we used a 2-CNN model\footnote{Cifar-10:\url{https://github.com/tensorflow/models/tree/r1.13.0/tutorials/image/cifar10}}. Every learner trained its local model using SGD with learning rate $\eta=0.05$ and batch size $\beta=100$ for 4 local epochs in-between federation rounds and learners shared all trainable weights (i.e., kernels and biases) with the federation controller. The federated environment was developed on the Tensorflow library v.1.13.1 and all experiments were run on a single server equipped with 8~GeForce GTX 1080~Ti graphics cards of 10~GB RAM each.

\section{Uniform Label Shuffling}\label{appendix:uniform_label_shuffling}

\subsection{Uniform \& IID} \label{appendix:uniform_label_shuffling_uniform_iid}

\subsubsection{Federation Convergence} \label{appendix:uniform_label_shuffling_uniform_iid_federation_convergence}

\begin{figure}[htpb]
\centering
  \subfloat[1 corrupted learner]{
    \includegraphics[width=0.45\linewidth]{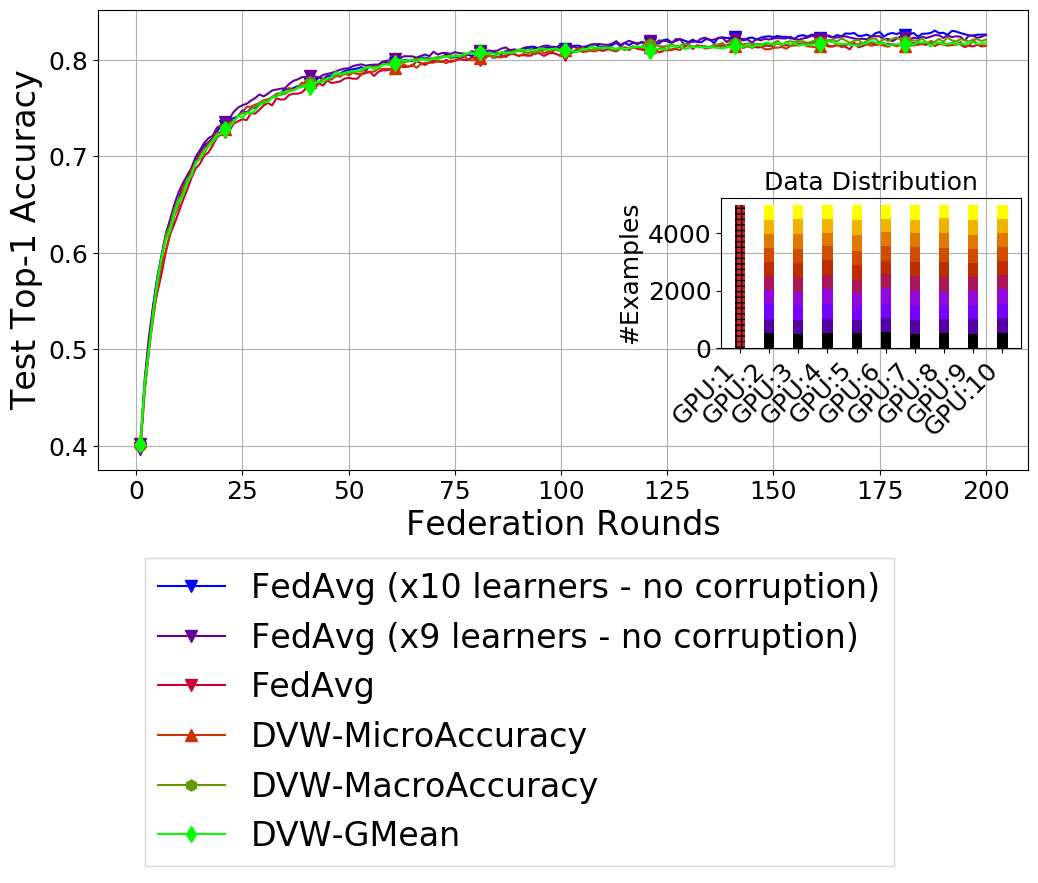}
    \label{subfig:cifar10_uniformiid_uniformlabelshuffling_convergence_1learner}
  }
  \subfloat[3 corrupted learners]{
    \includegraphics[width=0.45\linewidth]{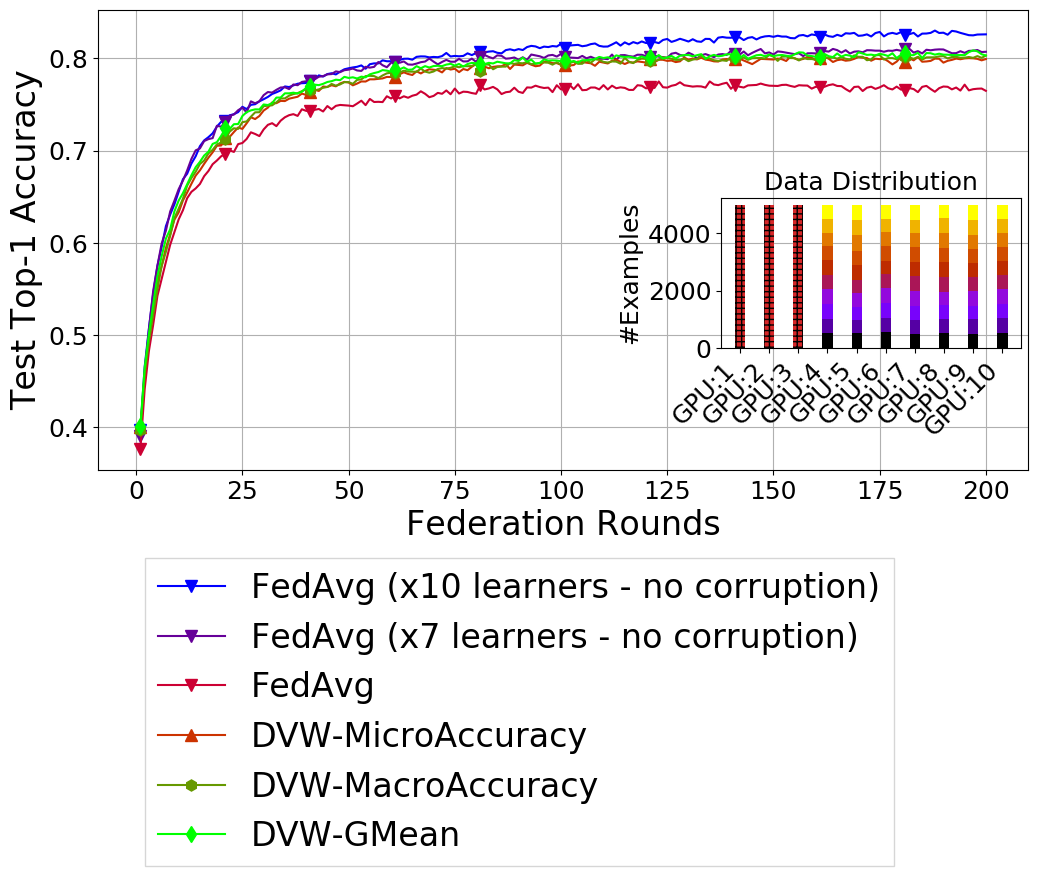}
    \label{subfig:cifar10_uniformiid_uniformlabelshuffling_convergence_3learners}
  }
  
  \subfloat[5 corrupted learners]{
    \includegraphics[width=0.45\linewidth]{plots/UniformLabelShuffling/PoliciesConvergence/Cifar10_UniformIID_UniformLabelShuffling_5Learners_L1L2L3L4L5_VanillaSGD_PoliciesConvergence.png}
    \label{subfig:cifar10_uniformiid_uniformlabelshuffling_convergence_5learners}
  }
  \subfloat[6 Corrupted learners]{
    \includegraphics[width=0.45\linewidth]{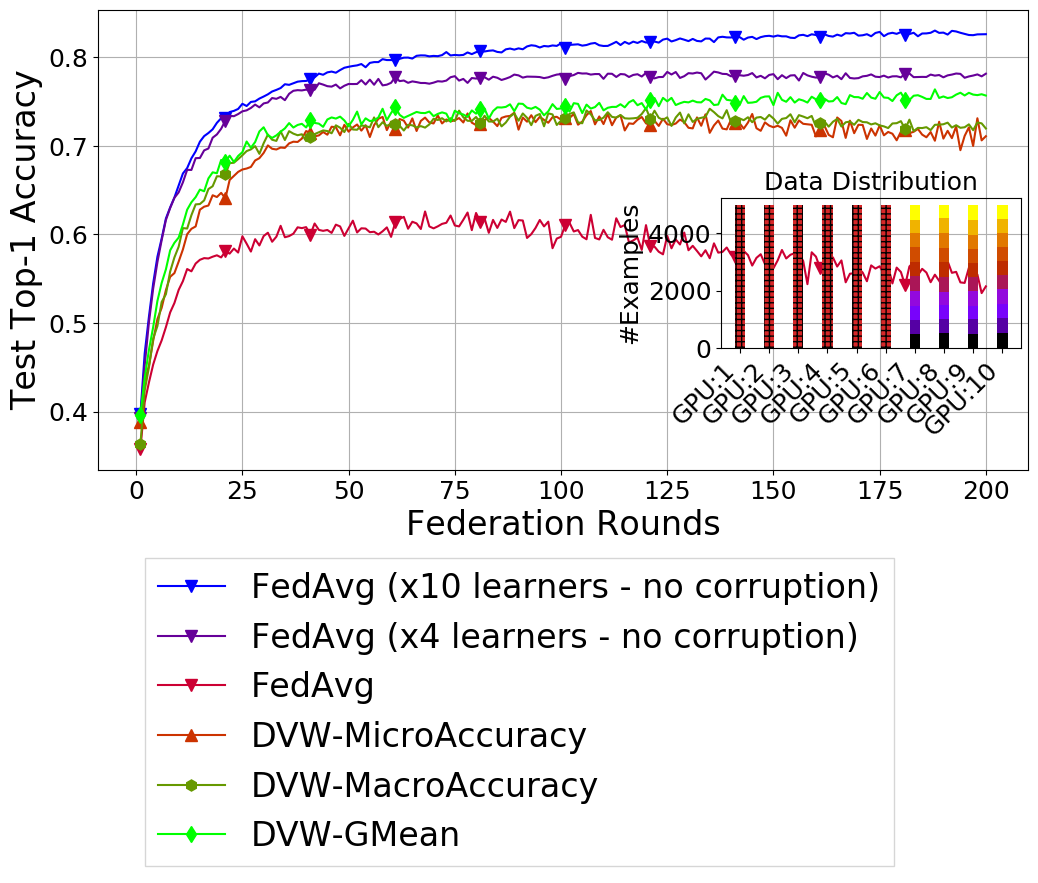}
    \label{subfig:cifar10_uniformiid_uniformlabelshuffling_convergence_6learners}
  }
  
  \subfloat[8 corrupted learners]{
  \centering
    \includegraphics[width=0.45\linewidth]{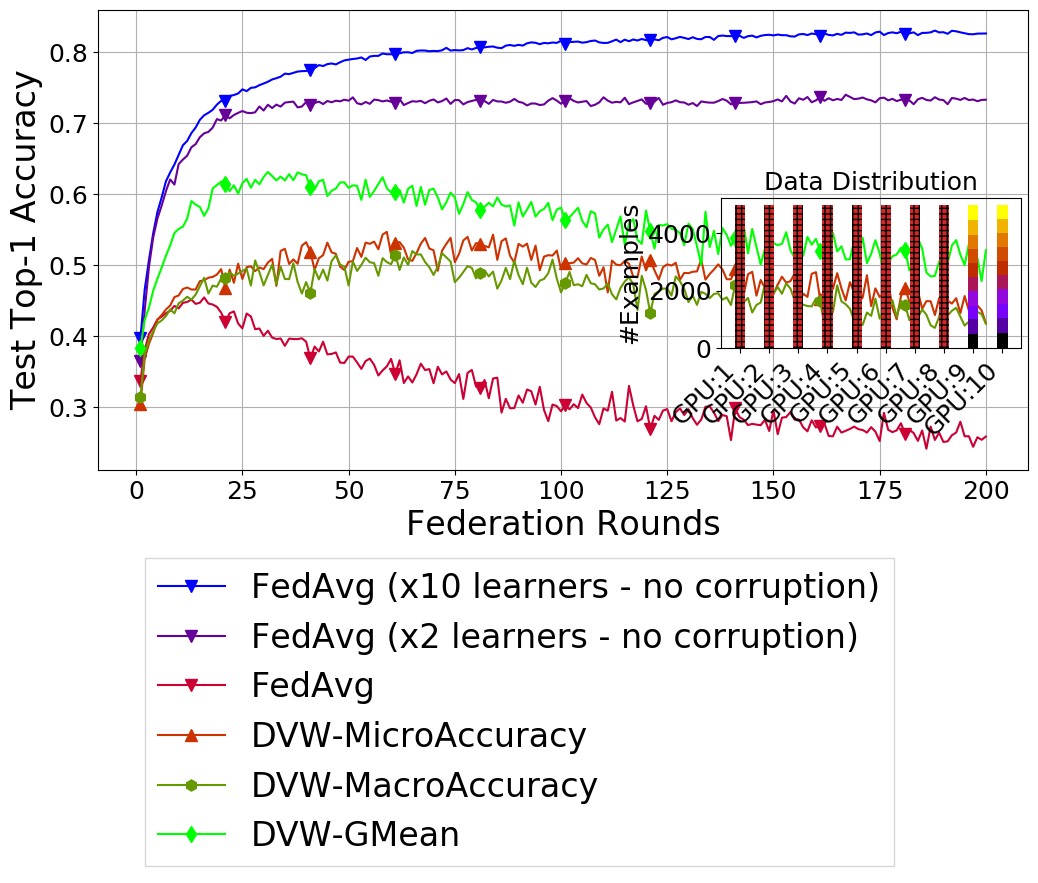}
    \label{subfig:cifar10_uniformiid_uniformlabelshuffling_convergence_8learners}
  }
  
  \caption{\textbf{Federation convergence for the Uniform Label Shuffling data poisoning attack in the Uniform \& IID learning environment.} Federation performance is measured on the test top-1 accuracy over an increasing number of corrupted learners. Corrupted learners are marked with the red hatch within the data distribution inset. For every learning environment, we include the convergence of the federation with no corruption (10 honest learners) and with exclusion of the corrupted learners (x honest learners). We also present the convergence of the federation for FedAvg (baseline) and different performance aggregation schemes, Micro-Accuracy, Macro-Accuracy and Geometric-Mean. Geometric Mean demonstrates the highest resiliency against label shuffling poisoning.}
  \label{fig:cifar10_uniformiid_uniformlabelshuffling_federation_convergence}
\end{figure}

Figure \ref{fig:cifar10_uniformiid_uniformlabelshuffling_federation_convergence} demonstrates the federation convergence rate for every learning environment for the uniform label shuffling attack mode. It is evident that FedAvg fails to learn as the number of corrupted learners increases. Our Performance Weighting scheme can improve the resiliency of the federation by suppressing the contribution of the corrupted learners. As it is also shown, in some cases (e.g. 50\% corruption, five corrupted learners) the performance of the Geometric Mean is very close to the federated environment when corrupted learners are already excluded from the global aggregation.

\newpage %

\subsubsection{Learners Contribution Value} \label{appendix:uniform_label_shuffling_uniform_iid_learners_contribution_value}

In \cref{fig:cifar10_uniformiid_uniformlabelshuffling_contributionvalue_1learner,fig:cifar10_uniformiid_uniformlabelshuffling_contributionvalue_3learners,fig:cifar10_uniformiid_uniformlabelshuffling_contributionvalue_5learners,fig:cifar10_uniformiid_uniformlabelshuffling_contributionvalue_6learners,fig:cifar10_uniformiid_uniformlabelshuffling_contributionvalue_8learners} we demonstrate the contribution/weighting value of each learner in the federation for every learning environment for the three different performance metrics. While all metrics are able to distinguish corrupted from non-corrupted learners from very early stages of the federated training, the Geometric Mean also progressively downgrades the performance of the corrupted learners, which in turn proves to be beneficial for the overall performance of the federation.

\begin{figure}[htpb]

  \subfloat[DVW-MicroAccuracy]{
    \includegraphics[width=0.33\linewidth]{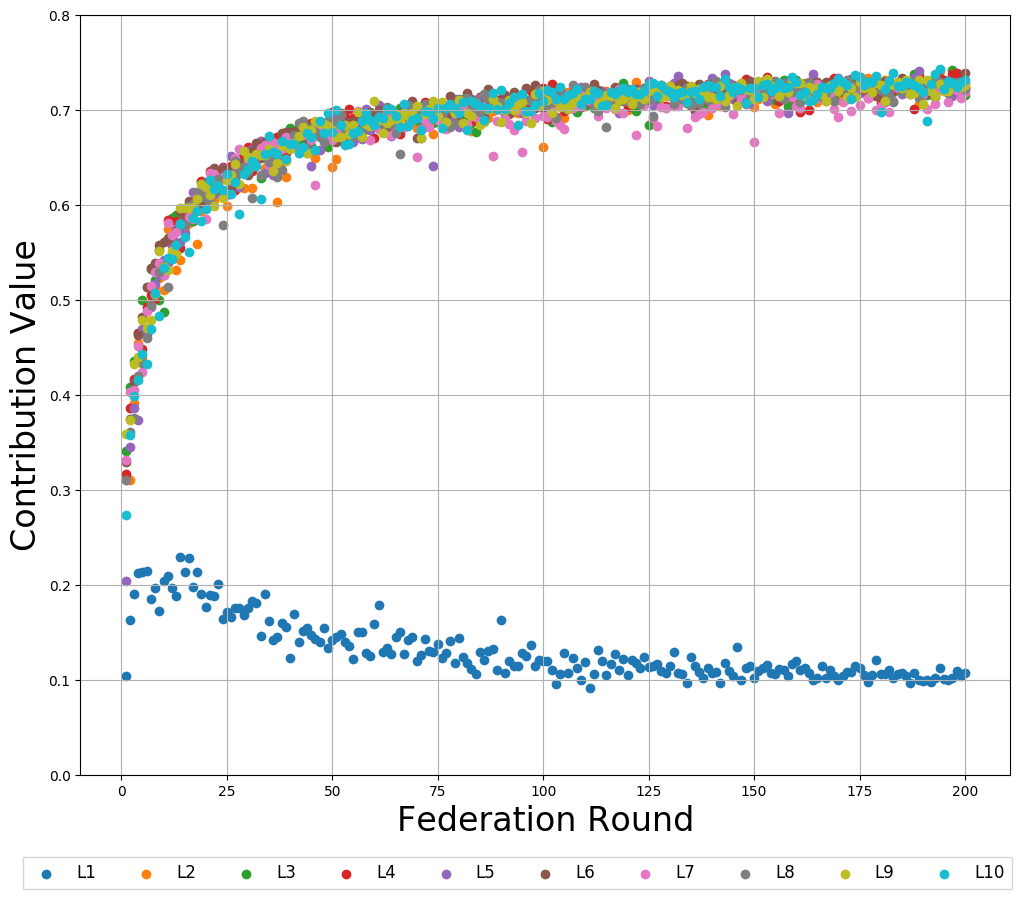}
    \label{subfig:cifar10_uniformiid_uniformlabelshuffling_microaccuracy_contributionvalue_1learner}
  }
  \subfloat[DVW-MacroAccuracy]{
    \includegraphics[width=0.33\linewidth]{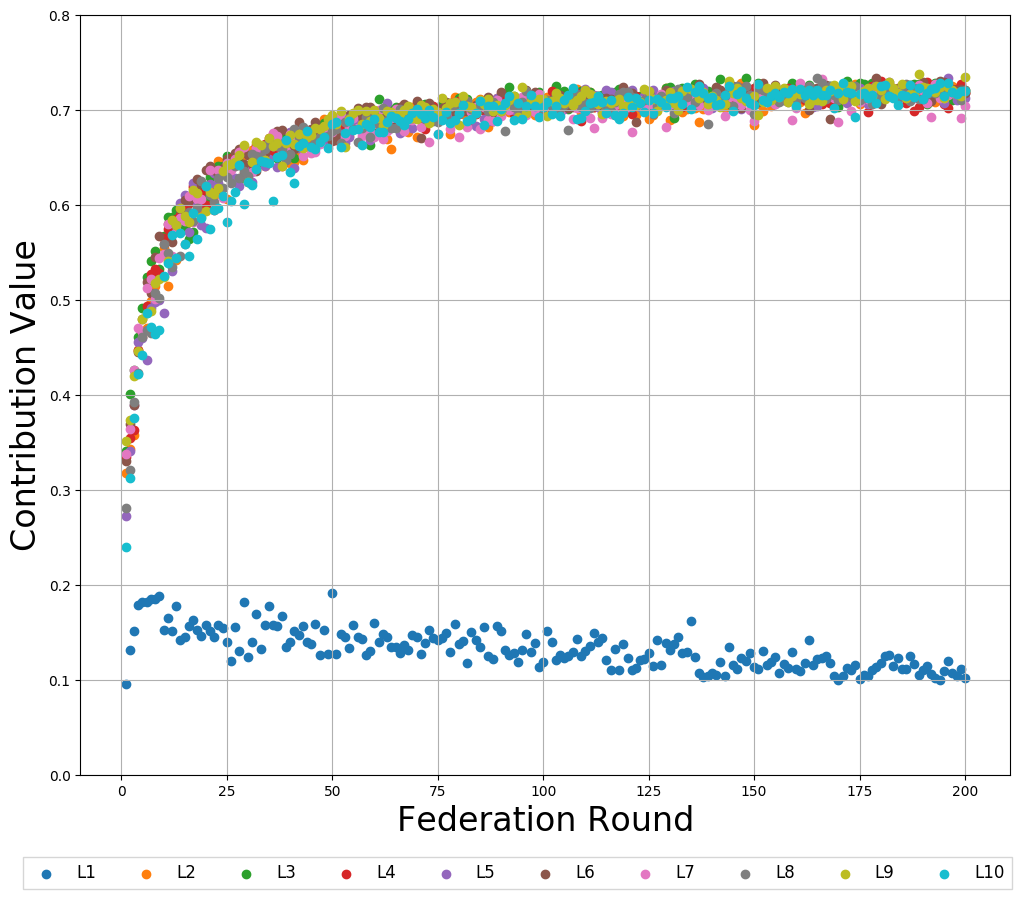}
    \label{subfig:cifar10_uniformiid_uniformlabelshuffling_macroaccuracy_contributionvalue_1learner}
  }
  \subfloat[DVW-GMean]{
    \includegraphics[width=0.33\linewidth]{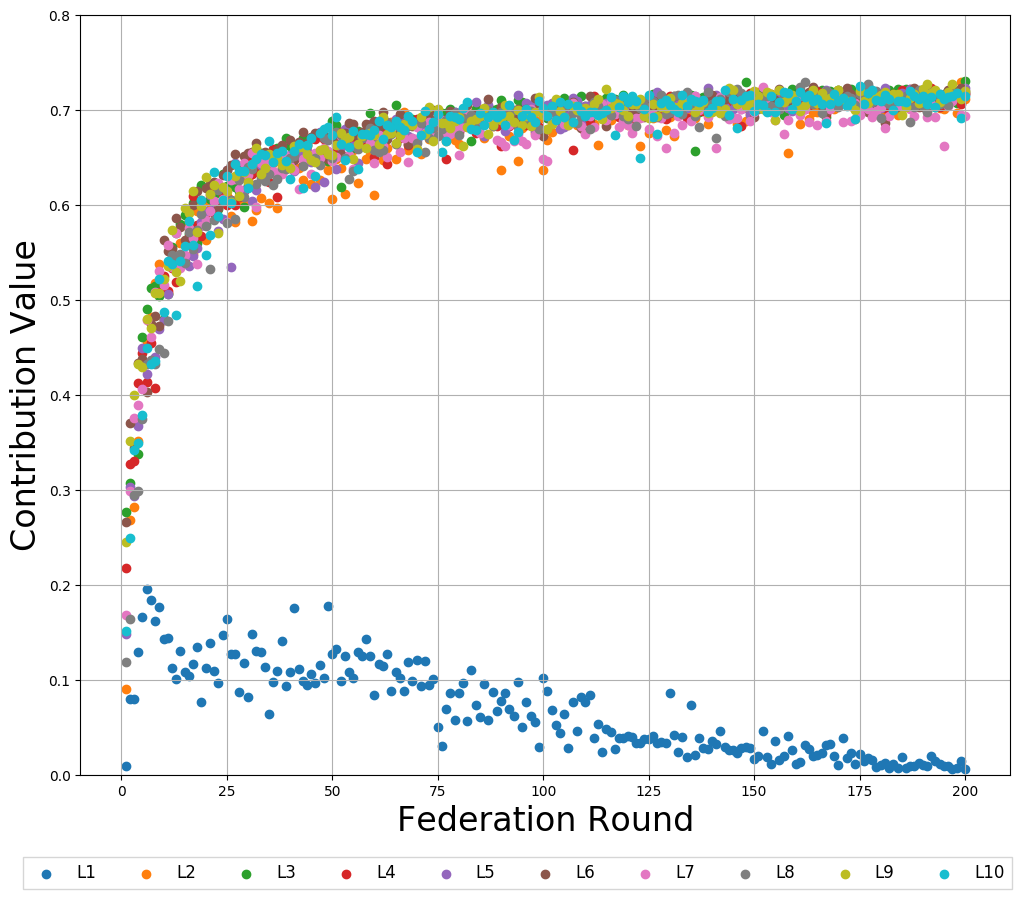}
    \label{subfig:cifar10_uniformiid_uniformlabelshuffling_gmean_contributionvalue_1learner}
  }
  \caption{Learners contribution value for the \textbf{Uniform Label Shuffling} attack in the \textbf{Uniform \& IID} learning environment, with \textbf{1 corrupted learner.}}
  \label{fig:cifar10_uniformiid_uniformlabelshuffling_contributionvalue_1learner}
\end{figure}
\begin{figure}[htpb]  
  \subfloat[DVW-MicroAccuracy]{
    \includegraphics[width=0.33\linewidth]{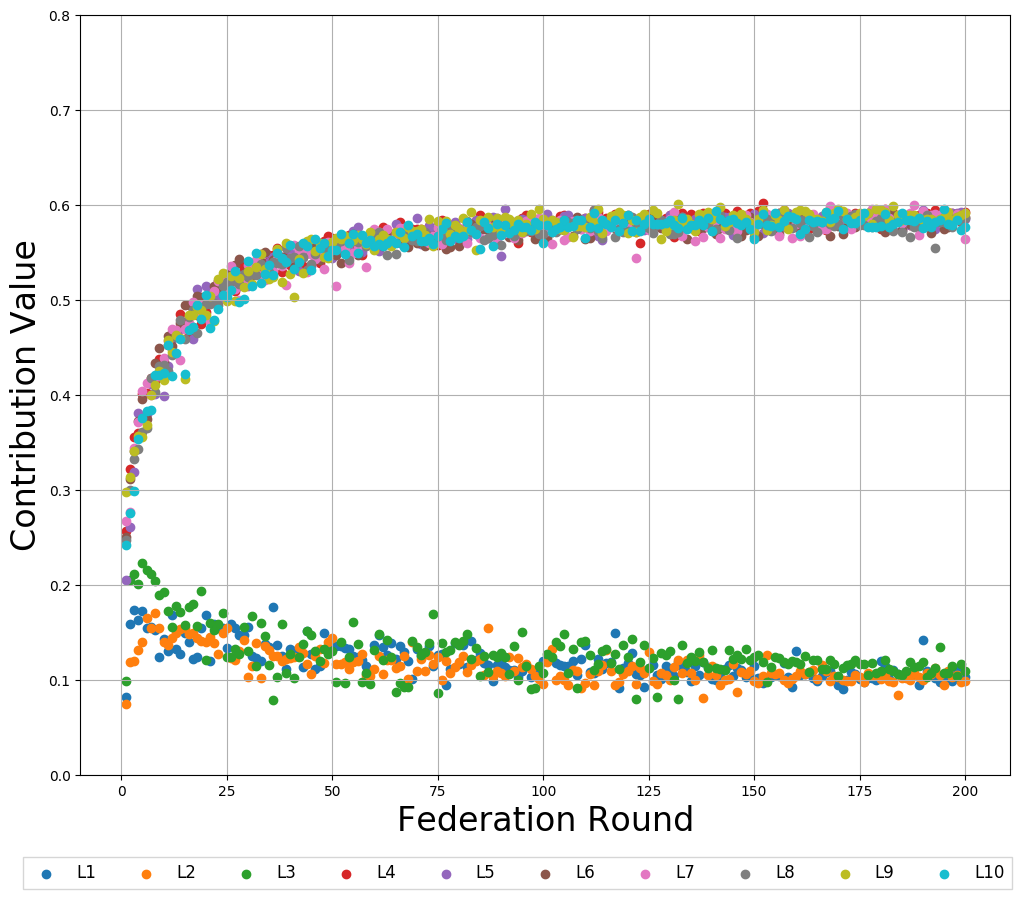}
    \label{subfig:cifar10_uniformiid_uniformlabelshuffling_microaccuracy_contributionvalue_3learners}
  }
  \subfloat[DVW-MacroAccuracy]{
  \centering
    \includegraphics[width=0.33\linewidth]{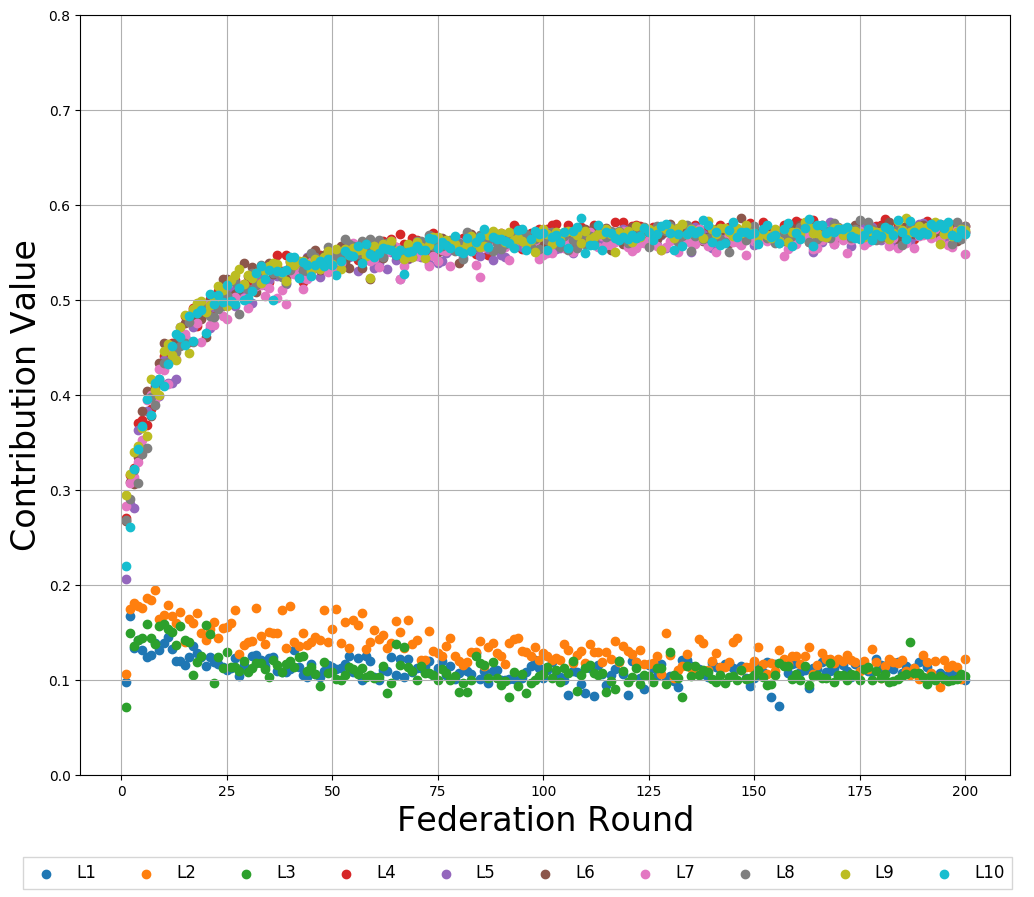}
    \label{subfig:cifar10_uniformiid_uniformlabelshuffling_macroaccuracy_contributionvalue_3learners}
  }
  \subfloat[DVW-GMean]{
  \centering
    \includegraphics[width=0.33\linewidth]{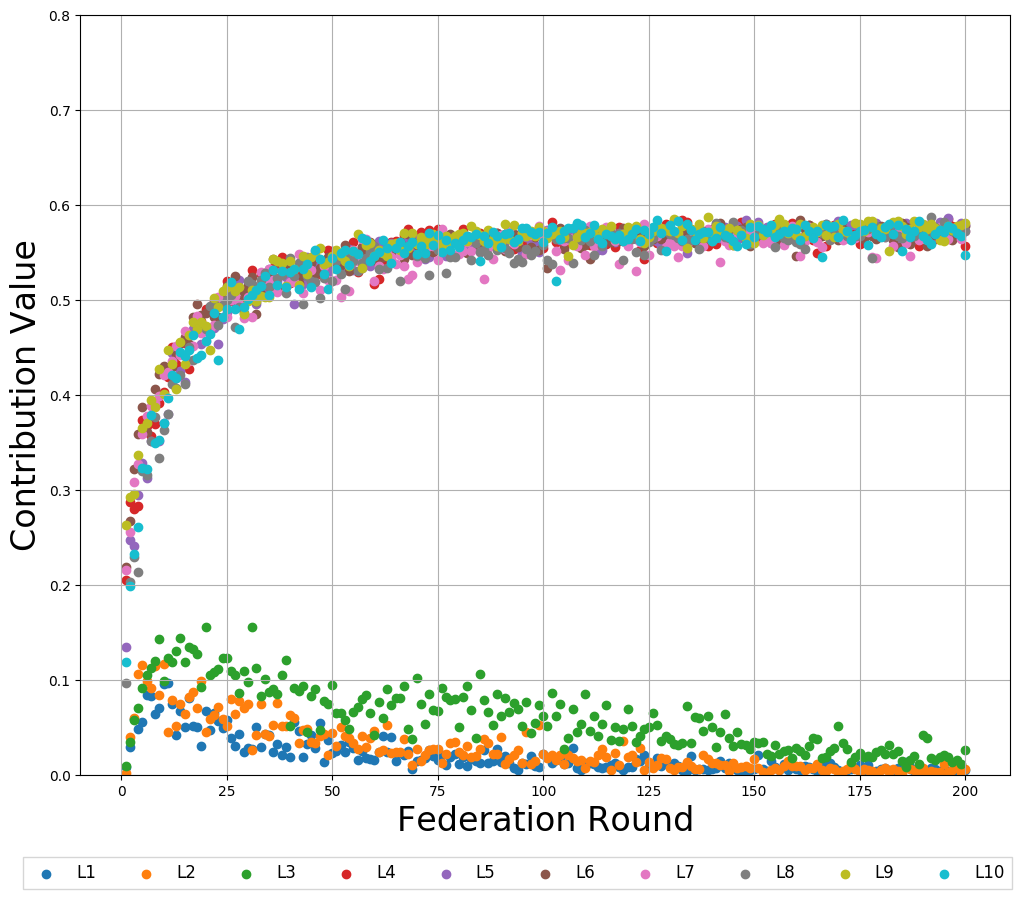}
    \label{subfig:cifar10_uniformiid_uniformlabelshuffling_gmean_contributionvalue_3learners}
  }
  \caption{Learners contribution value for the \textbf{Uniform Label Shuffling} attack in the \textbf{Uniform \& IID} learning environment, with \textbf{3 corrupted learners.}}
  \label{fig:cifar10_uniformiid_uniformlabelshuffling_contributionvalue_3learners}
\end{figure}

\begin{figure}[htpb]
  \subfloat[DVW-MicroAccuracy]{
  \centering
    \includegraphics[width=0.33\linewidth]{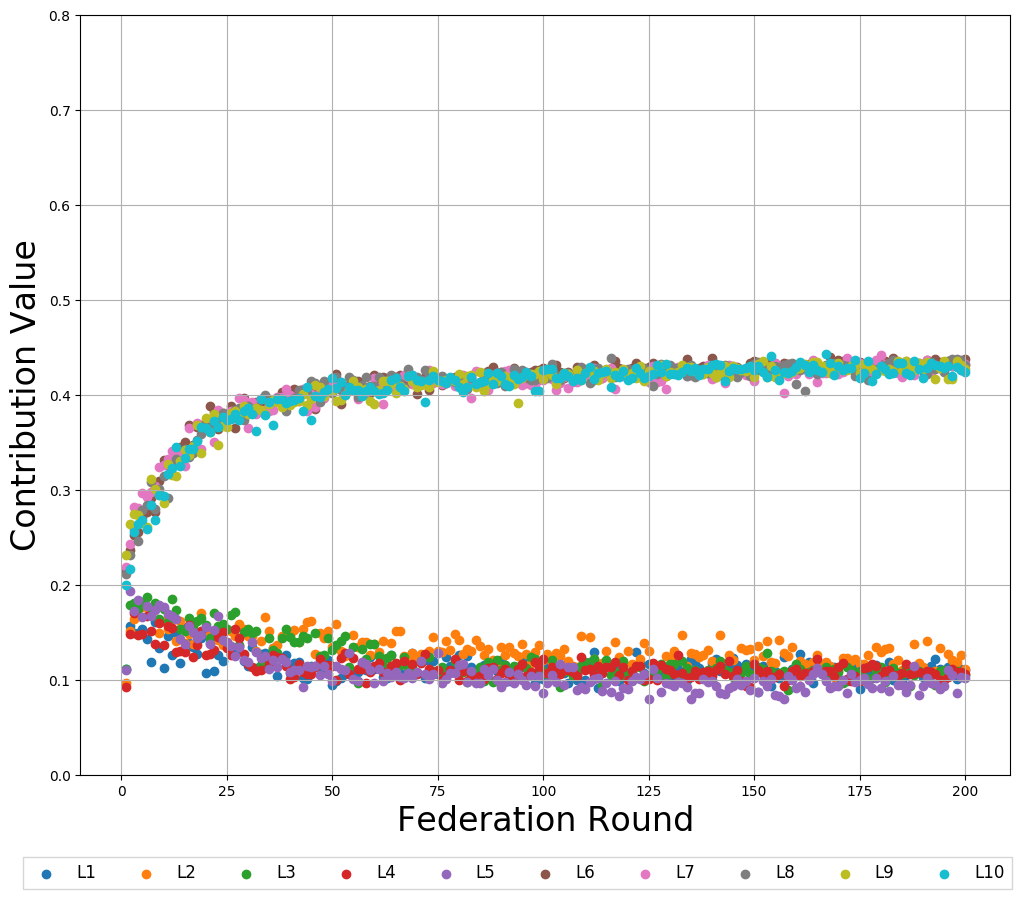}
    \label{subfig:cifar10_uniformiid_uniformlabelshuffling_microaccuracy_contributionvalue_5learners}
  }
  \subfloat[DVW-MacroAccuracy]{
  \centering
    \includegraphics[width=0.33\linewidth]{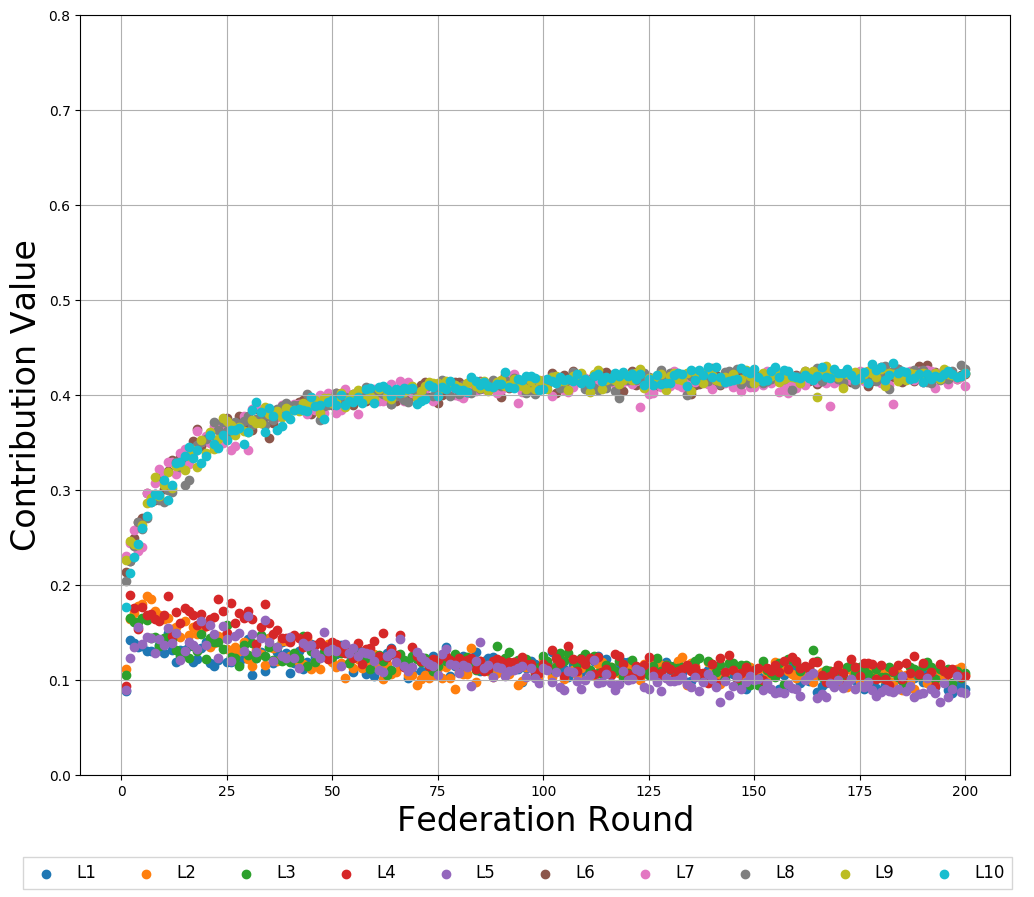}
    \label{subfig:cifar10_uniformiid_uniformlabelshuffling_macroaccuracy_contributionvalue_5learners}
  }
  \subfloat[DVW-GMean]{
  \centering
    \includegraphics[width=0.33\linewidth]{plots/UniformLabelShuffling/PoliciesContributionValueByLearner/Cifar10_UniformIID_UniformLabelShuffling_5LearnersL1L2L3L4L5_LearnersContributionValue_DVWGMean0001.png}
    \label{subfig:cifar10_uniformiid_uniformlabelshuffling_gmean_contributionvalue_5learners}
  }
  \caption{Learners contribution value for the \textbf{Uniform Label Shuffling} attack in the \textbf{Uniform \& IID} learning environment, with \textbf{5 corrupted learners.}}
  \label{fig:cifar10_uniformiid_uniformlabelshuffling_contributionvalue_5learners}
\end{figure} 

\begin{figure}[htpb]  
  \subfloat[DVW-MicroAccuracy]{
  \centering
    \includegraphics[width=0.33\linewidth]{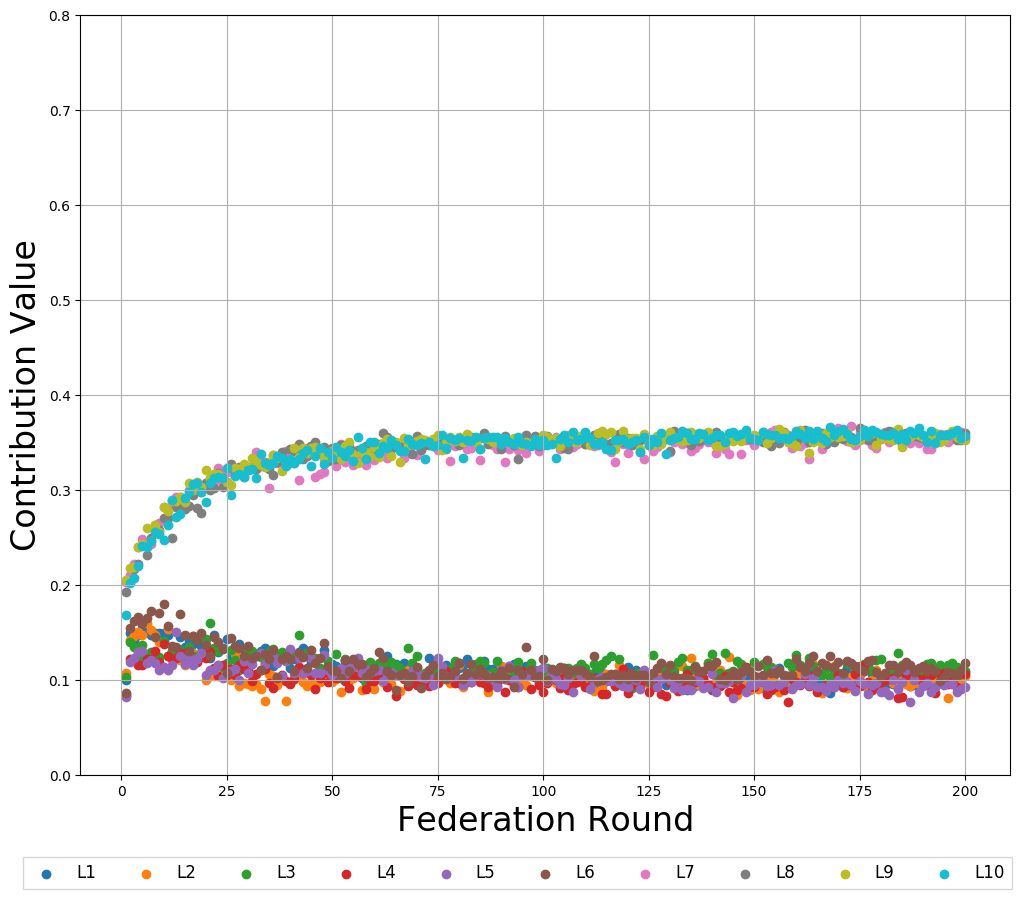}
    \label{subfig:cifar10_uniformiid_uniformlabelshuffling_microaccuracy_contributionvalue_6learners}
  }
  \subfloat[DVW-MacroAccuracy]{
  \centering
    \includegraphics[width=0.33\linewidth]{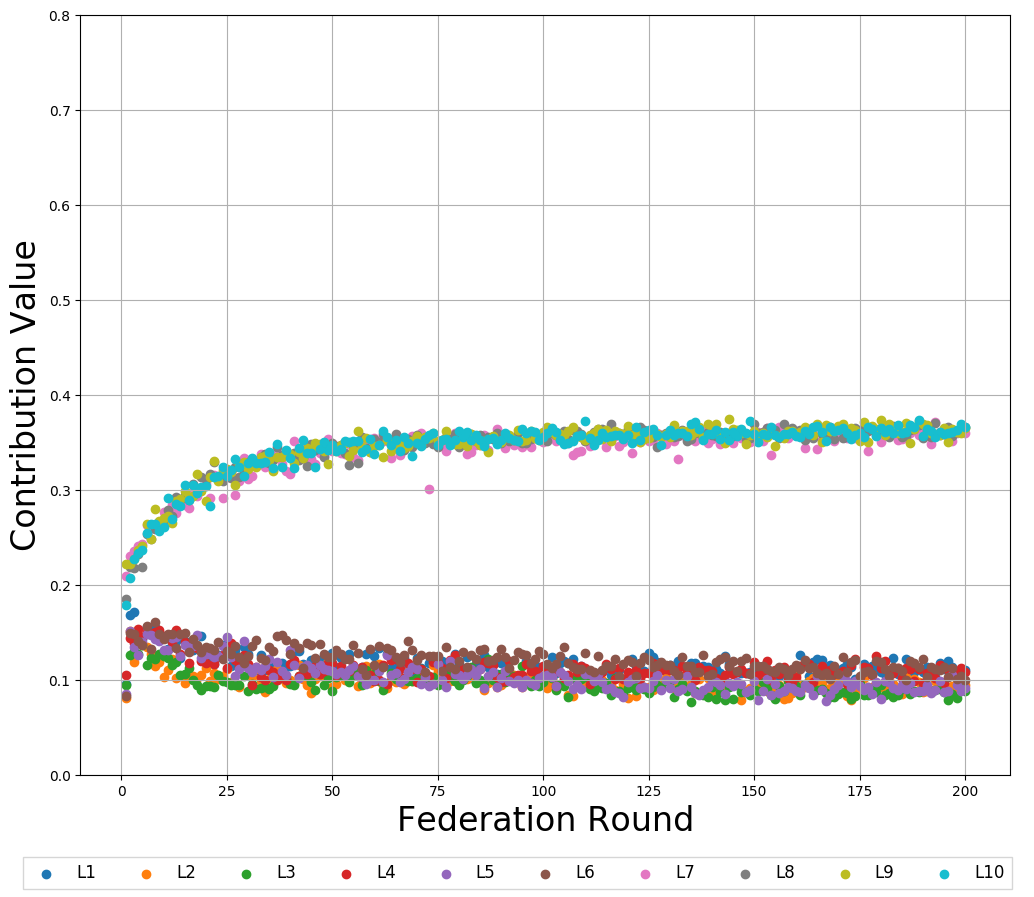}
    \label{subfig:cifar10_uniformiid_uniformlabelshuffling_macroaccuracy_contributionvalue_6learners}
  }
  \subfloat[DVW-GMean]{
  \centering
    \includegraphics[width=0.33\linewidth]{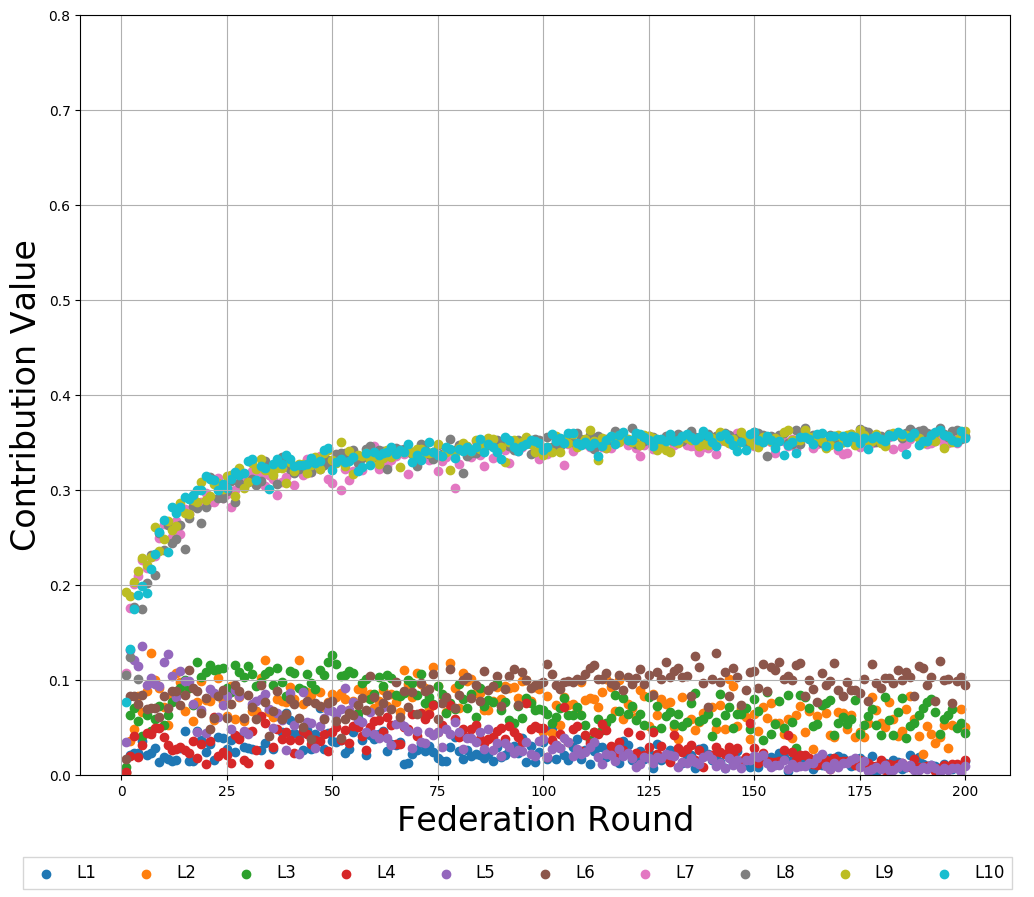}
    \label{subfig:cifar10_uniformiid_uniformlabelshuffling_gmean_contributionvalue_6learners}
  }
  \caption{Learners contribution value for the \textbf{Uniform Label Shuffling} attack in the \textbf{Uniform \& IID} learning environment, with \textbf{6 corrupted learners.}}
  \label{fig:cifar10_uniformiid_uniformlabelshuffling_contributionvalue_6learners}
\end{figure}

\begin{figure}[htpb]
  \subfloat[DVW-MicroAccuracy]{
  \centering
    \includegraphics[width=0.33\linewidth]{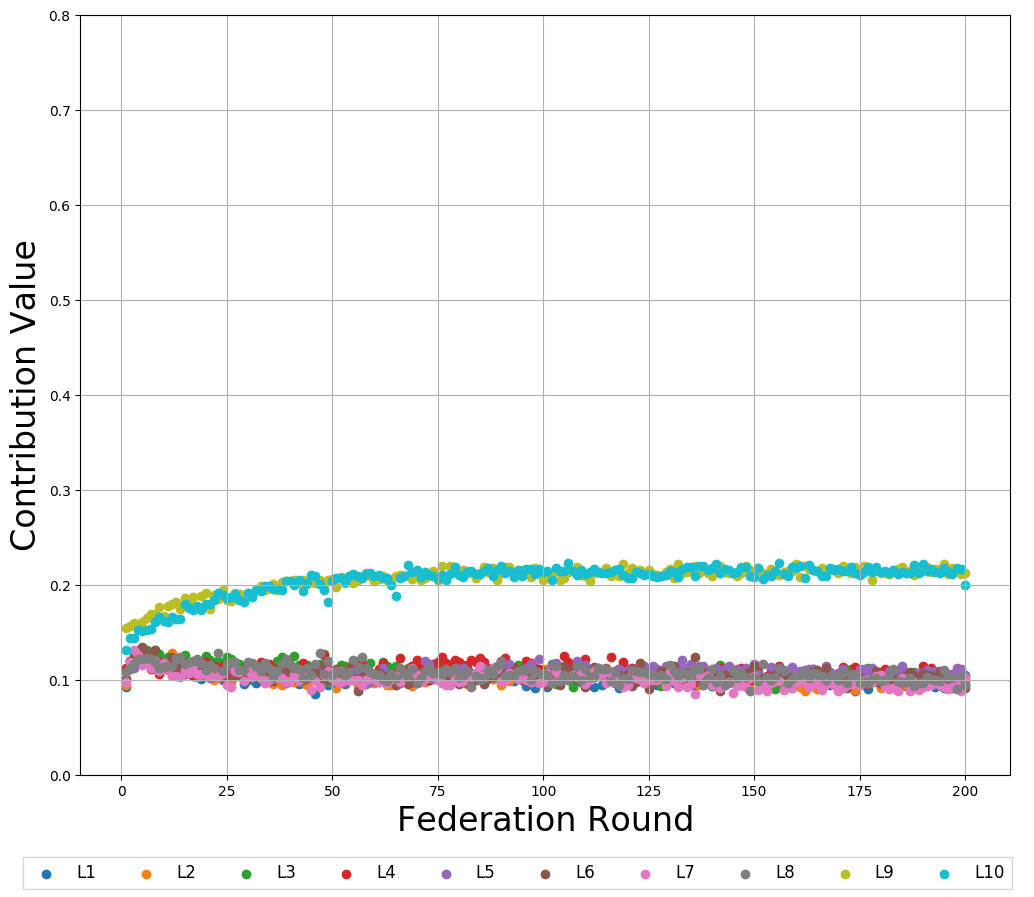}
    \label{subfig:cifar10_uniformiid_uniformlabelshuffling_microaccuracy_contributionvalue_8learners}
  }
  \subfloat[DVW-MacroAccuracy]{
  \centering
    \includegraphics[width=0.33\linewidth]{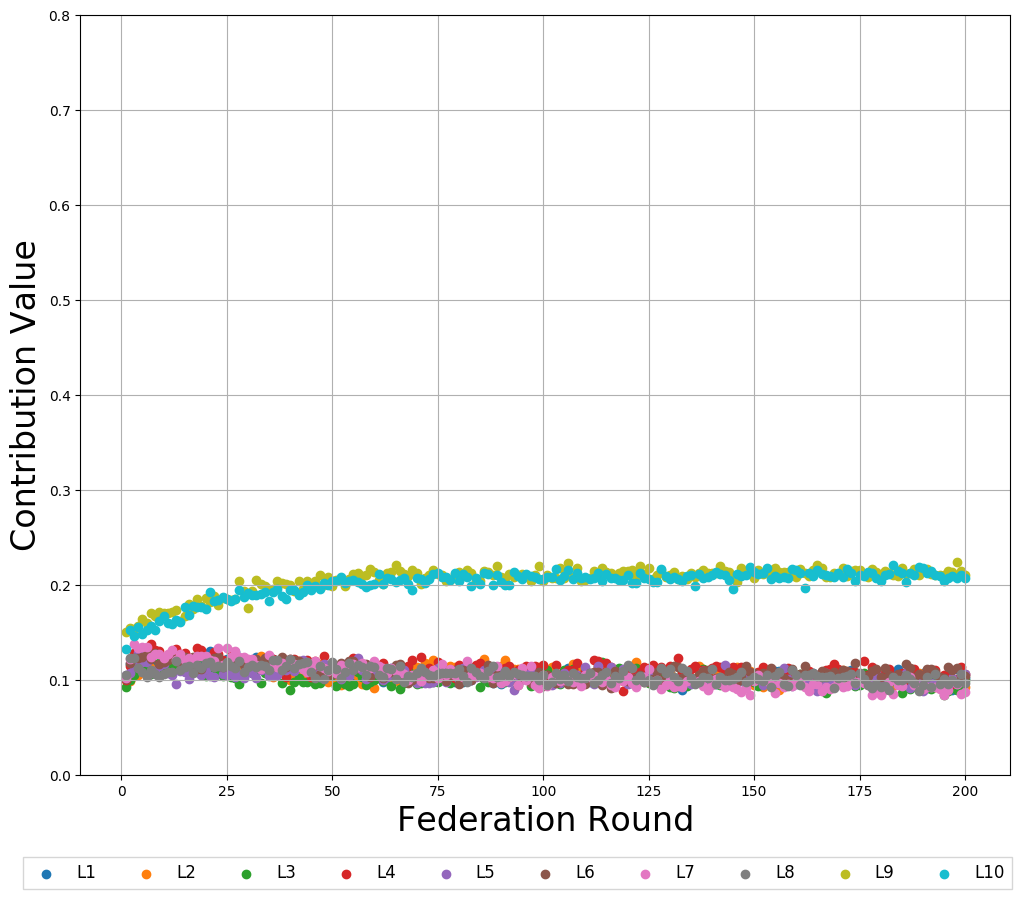}
    \label{subfig:cifar10_uniformiid_uniformlabelshuffling_macroaccuracy_contributionvalue_8learners}
  }
  \subfloat[DVW-GMean]{
  \centering
    \includegraphics[width=0.33\linewidth]{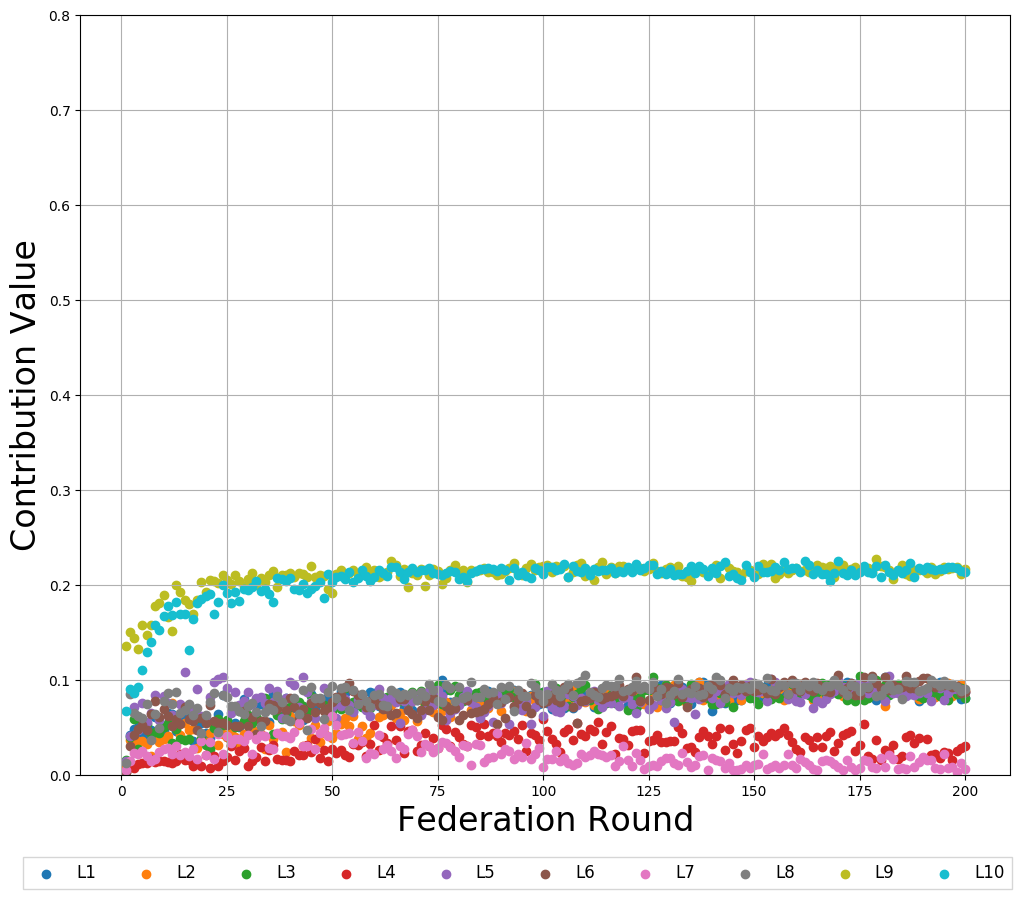}
    \label{subfig:cifar10_uniformiid_uniformlabelshuffling_gmean_contributionvalue_8learners}
  }
  \caption{Learners contribution value for the \textbf{Uniform Label Shuffling} attack in the \textbf{Uniform \& IID} learning environment, with \textbf{8 corrupted learners.}} 
  \label{fig:cifar10_uniformiid_uniformlabelshuffling_contributionvalue_8learners}
\end{figure}

\newpage %

\subsubsection{Learners Performance Per Class} \label{appendix:uniform_label_shuffling_uniform_iid_learners_performance_perclass}
In \cref{fig:cifar10_uniformiid_uniformlabelshuffling_performanceperclass_1learner,fig:cifar10_uniformiid_uniformlabelshuffling_performanceperclass_3learners,fig:cifar10_uniformiid_uniformlabelshuffling_performanceperclass_5learners,fig:cifar10_uniformiid_uniformlabelshuffling_contributionvalue_6learners,fig:cifar10_uniformiid_uniformlabelshuffling_contributionvalue_8learners} we demonstrate the per-community (global) and per-learner model accuracy for every class of the distributed validation dataset at the final federation round. The scaling of the colors is different across the weighting schemes' matrices and the distinction between corrupted and non-corrupted learners should be made by inspecting each matrix independently (intra-matrix).

\begin{figure}[htpb]

  \subfloat[DVW-MicroAccuracy]{
    \includegraphics[width=0.33\linewidth]{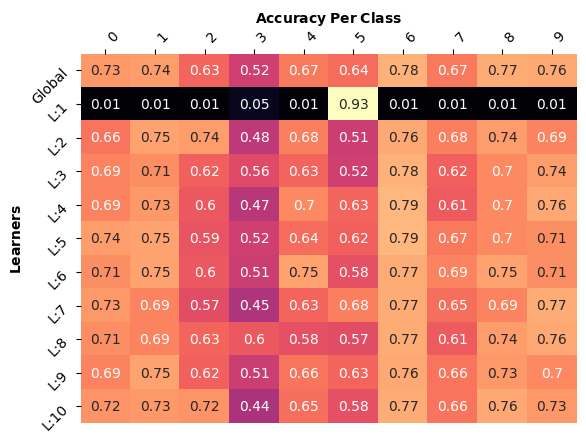}
    \label{subfig:cifar10_uniformiid_uniformlabelshuffling_microaccuracy_performanceperclass_1learner}
  }
  \subfloat[DVW-MacroAccuracy]{
    \includegraphics[width=0.33\linewidth]{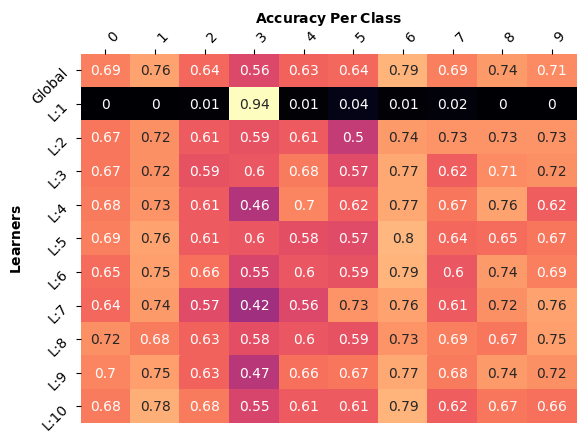}
    \label{subfig:cifar10_uniformiid_uniformlabelshuffling_macroaccuracy_performanceperclass_1learner}
  }
  \subfloat[DVW-GMean]{
    \includegraphics[width=0.33\linewidth]{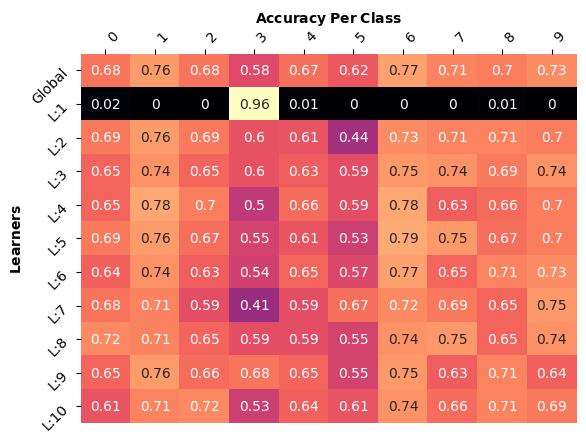}
    \label{subfig:cifar10_uniformiid_uniformlabelshuffling_gmean_performanceperclass_1learner}
  }
  \caption{Accuracy per class in the last federation round for the community (global) model and each learner for the \textbf{Uniform Label Shuffling} attack in the \textbf{Uniform \& IID} learning environment, with \textbf{1 corrupted learner.}}
  \label{fig:cifar10_uniformiid_uniformlabelshuffling_performanceperclass_1learner}
\end{figure}

\begin{figure}[htpb]  
  \subfloat[DVW-MicroAccuracy]{
    \includegraphics[width=0.33\linewidth]{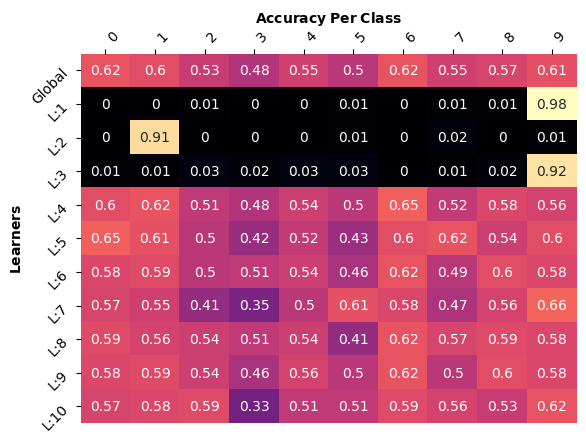}
    \label{subfig:cifar10_uniformiid_uniformlabelshuffling_microaccuracy_performanceperclass_3learners}
  }
  \subfloat[DVW-MacroAccuracy]{
  \centering
    \includegraphics[width=0.33\linewidth]{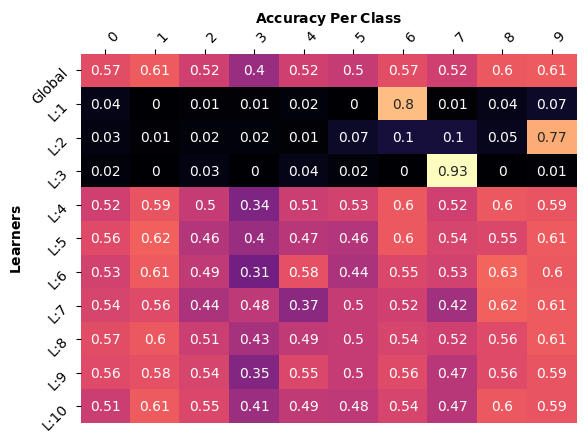}
    \label{subfig:cifar10_uniformiid_uniformlabelshuffling_macroaccuracy_performanceperclass_3learners}
  }
  \subfloat[DVW-GMean]{
  \centering
    \includegraphics[width=0.33\linewidth]{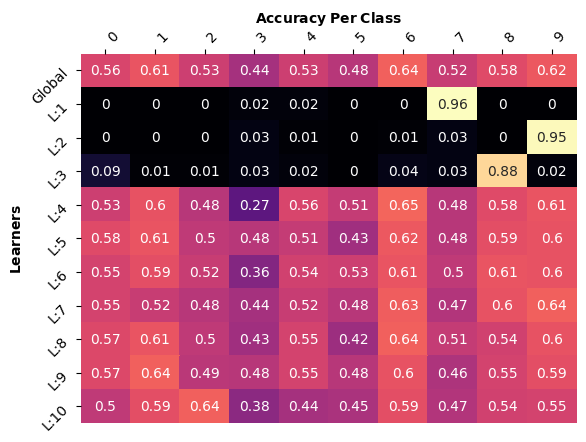}
    \label{subfig:cifar10_uniformiid_uniformlabelshuffling_gmean_performanceperclass_3learners}
  }
  \caption{Accuracy per class in the last federation round for the community (global) model and each learner for the \textbf{Uniform Label Shuffling} attack in the \textbf{Uniform \& IID} learning environment, with \textbf{3 corrupted learners.}}
  \label{fig:cifar10_uniformiid_uniformlabelshuffling_performanceperclass_3learners}
\end{figure}

\begin{figure}[htpb]
  \subfloat[DVW-MicroAccuracy]{
  \centering
    \includegraphics[width=0.33\linewidth]{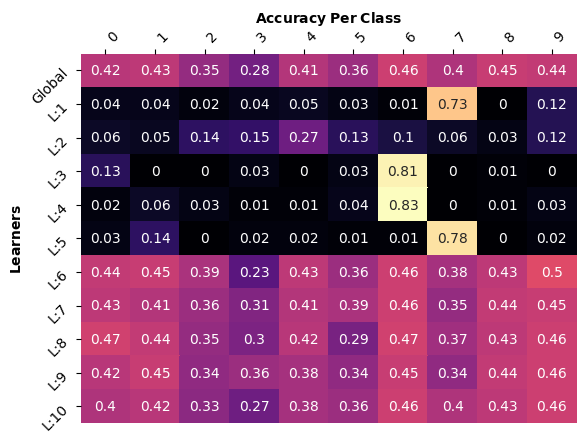}
    \label{subfig:cifar10_uniformiid_uniformlabelshuffling_microaccuracy_performanceperclass_5learners}
  }
  \subfloat[DVW-MacroAccuracy]{
  \centering
    \includegraphics[width=0.33\linewidth]{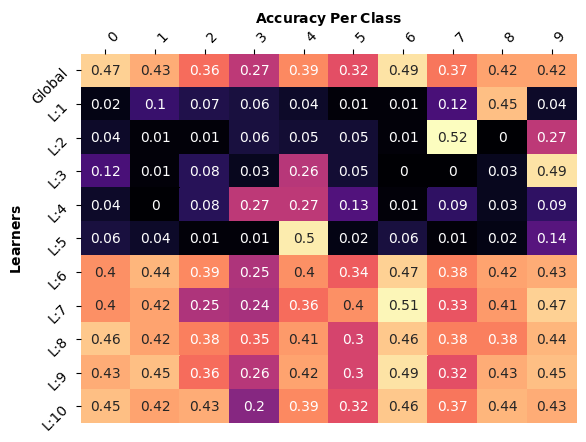}
    \label{subfig:cifar10_uniformiid_uniformlabelshuffling_macroaccuracy_performanceperclass_5learners}
  }
  \subfloat[DVW-GMean]{
  \centering
    \includegraphics[width=0.33\linewidth]{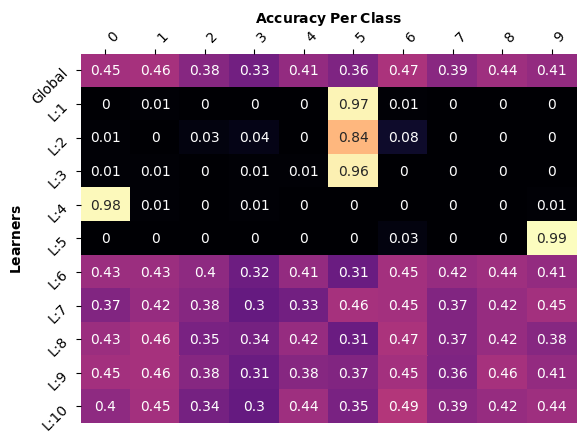}
    \label{subfig:cifar10_uniformiid_uniformlabelshuffling_gmean_performanceperclass_5learners}
  }
  \caption{Accuracy per class in the last federation round for the community (global) model and each learner for the \textbf{Uniform Label Shuffling} attack in the \textbf{Uniform \& IID} learning environment, with \textbf{5 corrupted learners.}}
  \label{fig:cifar10_uniformiid_uniformlabelshuffling_performanceperclass_5learners}
\end{figure} 

\begin{figure}[htpb]  
  \subfloat[DVW-MicroAccuracy]{
  \centering
    \includegraphics[width=0.33\linewidth]{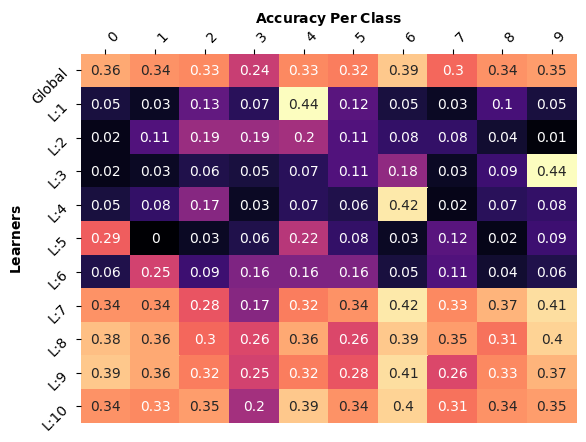}
    \label{subfig:cifar10_uniformiid_uniformlabelshuffling_microaccuracy_performanceperclass_6learners}
  }
  \subfloat[DVW-MacroAccuracy]{
  \centering
    \includegraphics[width=0.33\linewidth]{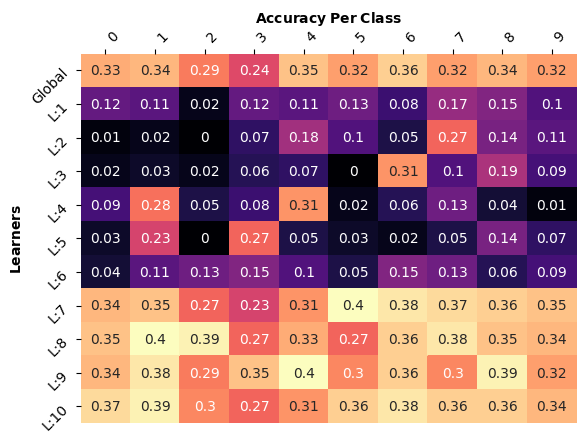}
    \label{subfig:cifar10_uniformiid_uniformlabelshuffling_macroaccuracy_performanceperclass_6learners}
  }
  \subfloat[DVW-GMean]{
  \centering
    \includegraphics[width=0.33\linewidth]{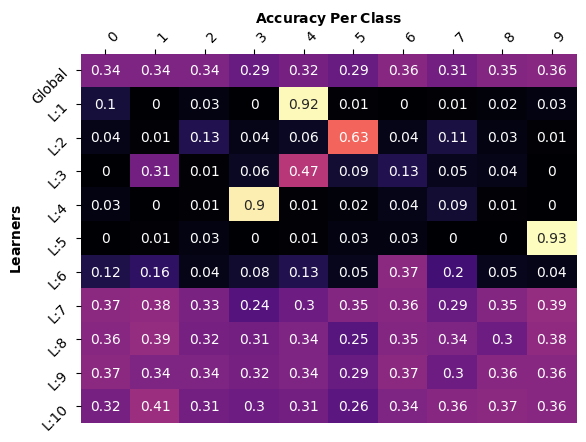}
    \label{subfig:cifar10_uniformiid_uniformlabelshuffling_gmean_performanceperclass_6learners}
  }
  \caption{Accuracy per class in the last federation round for the community (global) model and each learner for the \textbf{Uniform Label Shuffling} attack in the \textbf{Uniform \& IID} learning environment, with \textbf{6 corrupted learners.}}
  \label{fig:cifar10_uniformiid_uniformlabelshuffling_performanceperclass_6learners}
\end{figure}

\begin{figure}[htpb]
  \subfloat[DVW-MicroAccuracy]{
  \centering
    \includegraphics[width=0.33\linewidth]{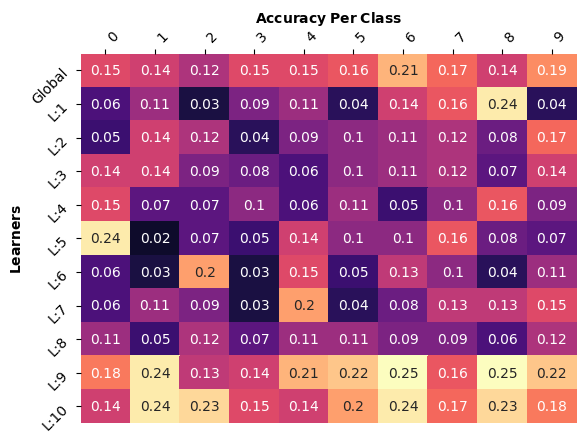}
    \label{subfig:cifar10_uniformiid_uniformlabelshuffling_microaccuracy_performanceperclass_8learners}
  }
  \subfloat[DVW-MacroAccuracy]{
  \centering
    \includegraphics[width=0.33\linewidth]{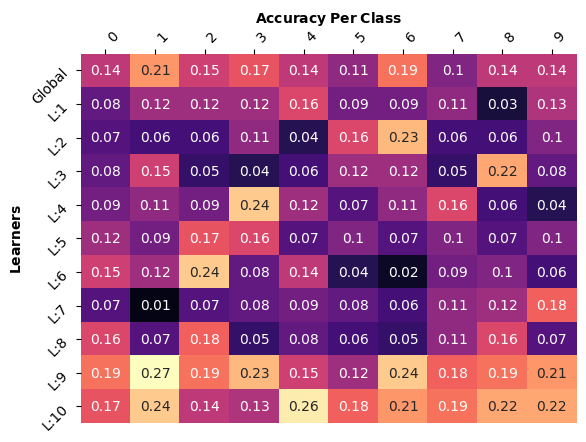}
    \label{subfig:cifar10_uniformiid_uniformlabelshuffling_macroaccuracy_performanceperclass_8learners}
  }
  \subfloat[DVW-GMean]{
  \centering
    \includegraphics[width=0.33\linewidth]{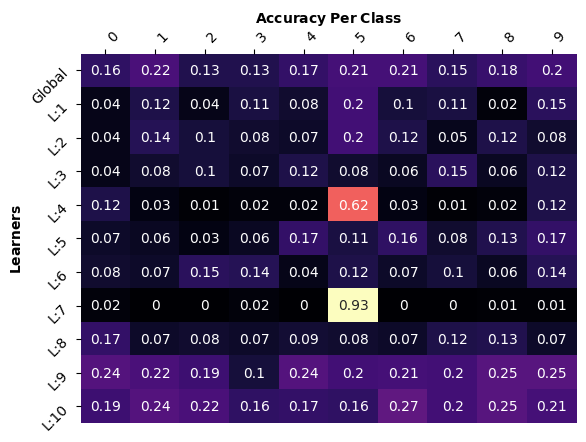}
    \label{subfig:cifar10_uniformiid_uniformlabelshuffling_gmean_performanceperclass_8learners}
  }
  \caption{Accuracy per class in the last federation round for the community (global) model and each learner for the \textbf{Uniform Label Shuffling} attack in the \textbf{Uniform \& IID} learning environment, with \textbf{8 corrupted learners.}} 
  \label{fig:cifar10_uniformiid_uniformlabelshuffling_performanceperclass_8learners}
\end{figure}

\newpage %

\subsection{PowerLaw \& IID} \label{appendix:uniform_label_shuffling_powerlaw_iid}

\subsubsection{Federation Convergence} \label{appendix:uniform_label_shuffling_powerlaw_iid_federation_convergence}
\begin{figure}[htpb]
\centering
  \subfloat[1 corrupted learner]{
    \includegraphics[width=0.45\linewidth]{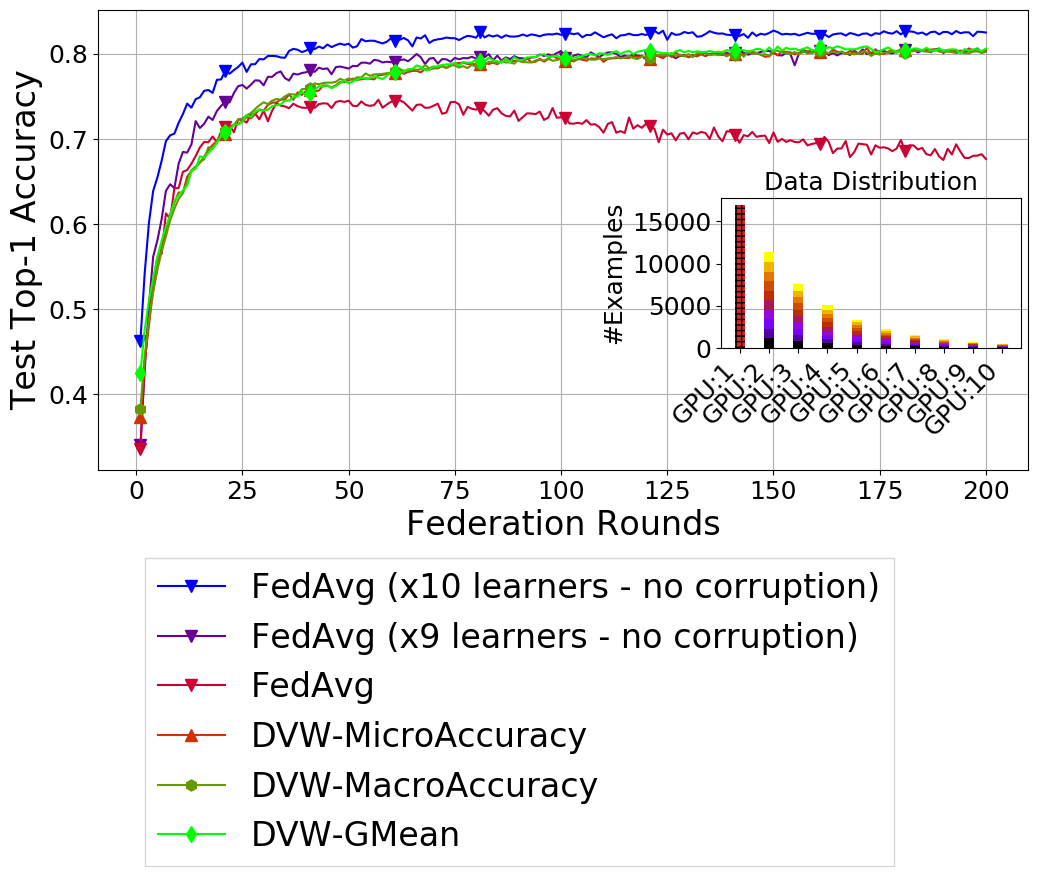}
    \label{subfig:cifar10_powerlawiid_uniformlabelshuffling_convergence_1learner}
  }
  \subfloat[3 corrupted learners]{
    \includegraphics[width=0.45\linewidth]{plots/UniformLabelShuffling/PoliciesConvergence/Cifar10_PowerLawIID_UniformLabelShuffling_3Learners_L1L2L3_VanillaSGD_PoliciesConvergence.png}
    \label{subfig:cifar10_powerlawiid_uniformlabelshuffling_convergence_3learners}
  }
  
  \subfloat[5 corrupted learners]{
    \includegraphics[width=0.45\linewidth]{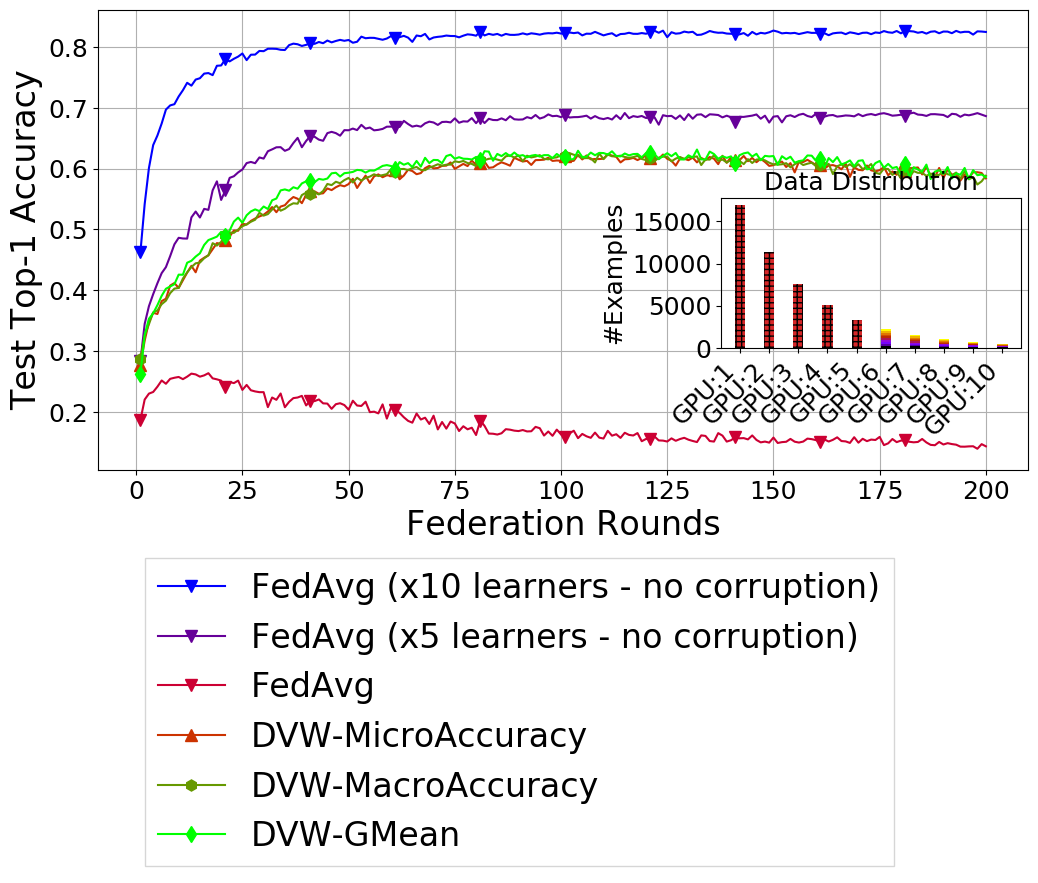}
    \label{subfig:cifar10_powerlawiid_uniformlabelshuffling_convergence_5learner5}
  }
  
  \caption{\textbf{Federation convergence for the Uniform Label Shuffling data poisoning attack in the PowerLaw \& IID learning environment.} Federation performance is measured on the test top-1 accuracy over an increasing number of corrupted learners. Corrupted learners are marked with the red hatch within the data distribution inset. For every learning environment, we include the convergence of the federation with no corruption (10 honest learners) and with exclusion of the corrupted learners (x honest learners). We also present the convergence of the federation for FedAvg (baseline) and different performance weighting aggregation schemes, Micro-Accuracy, Macro-Accuracy and Geometric-Mean.}
  \label{fig:cifar10_powerlawiid_uniformlabelshuffling_federation_convergence}
\end{figure}

Figure \ref{fig:cifar10_powerlawiid_uniformlabelshuffling_federation_convergence} demonstrates the federation convergence rate for every learning environment in the powerlaw and iid settings. Similar to the uniform case, FedAvg is not able to learn as the number of corrupted learners increases and fails to generalize (see 3 and 5 corrupted learners). In antithesis, our Performance Weighting scheme can improve the resiliency of the federation by suppressing the contribution of the corrupted learners and in turn provide more learning robustness.

\newpage %

\subsubsection{Learners Contribution Value} \label{appendix:uniform_label_shuffling_powerlaw_iid_learners_contribution_value}

\begin{figure}[htpb]
\centering
  \subfloat[DVW-MicroAccuracy]{
    \includegraphics[width=0.33\linewidth]{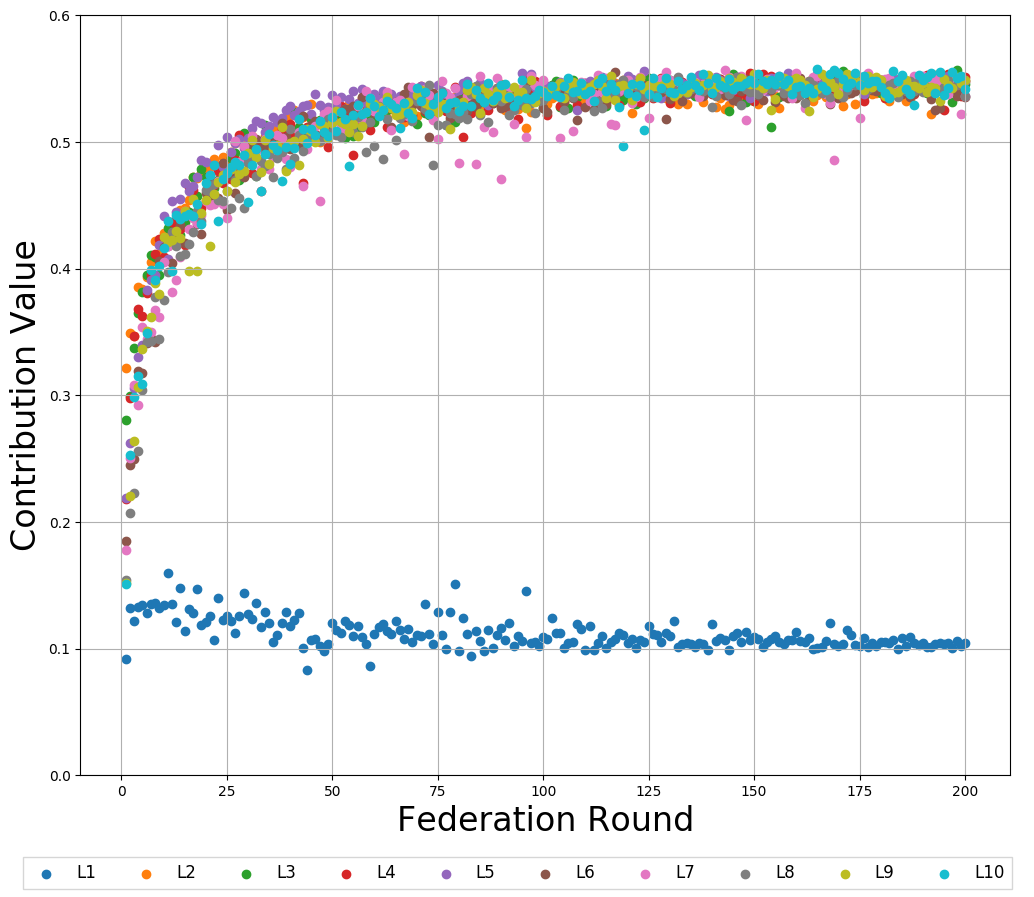}
    \label{subfig:cifar10_powerlawiid_uniformlabelshuffling_microaccuracy_contributionvalue_1learner}
  }
  \subfloat[DVW-MacroAccuracy]{
    \includegraphics[width=0.33\linewidth]{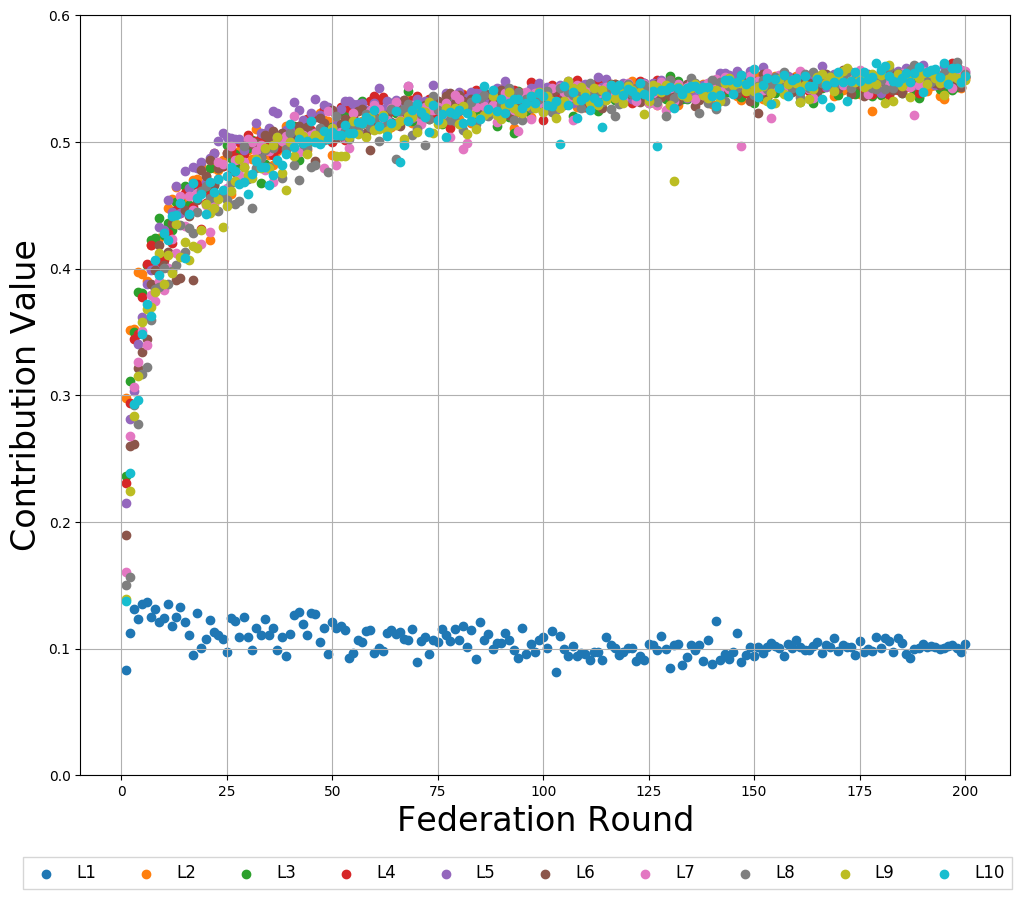}
    \label{subfig:cifar10_powerlawiid_uniformlabelshuffling_macroaccuracy_contributionvalue_1learner}
  }
  \subfloat[DVW-GMean]{
    \includegraphics[width=0.33\linewidth]{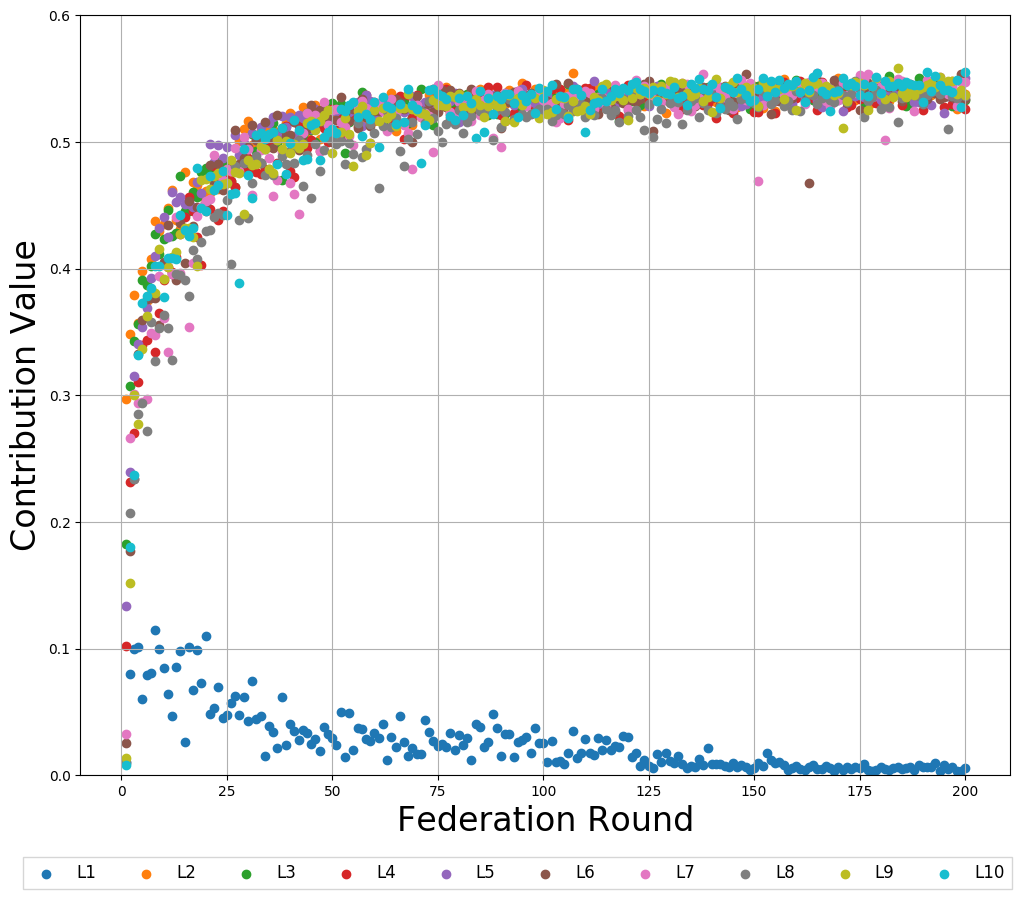}
    \label{subfig:cifar10_powerlawiid_uniformlabelshuffling_gmean_contributionvalue_1learner}
  }
  \caption{Learners contribution value for the \textbf{Uniform Label Shuffling} attack in the \textbf{PowerLaw \& IID} learning environment, \textbf{with 1 corrupted learner.}}
  \label{fig:cifar10_powerlawiid_uniformlabelshuffling_contributionvalue_1learner}
\end{figure}

\begin{figure}[htpb]  
  \subfloat[DVW-MicroAccuracy]{
    \includegraphics[width=0.33\linewidth]{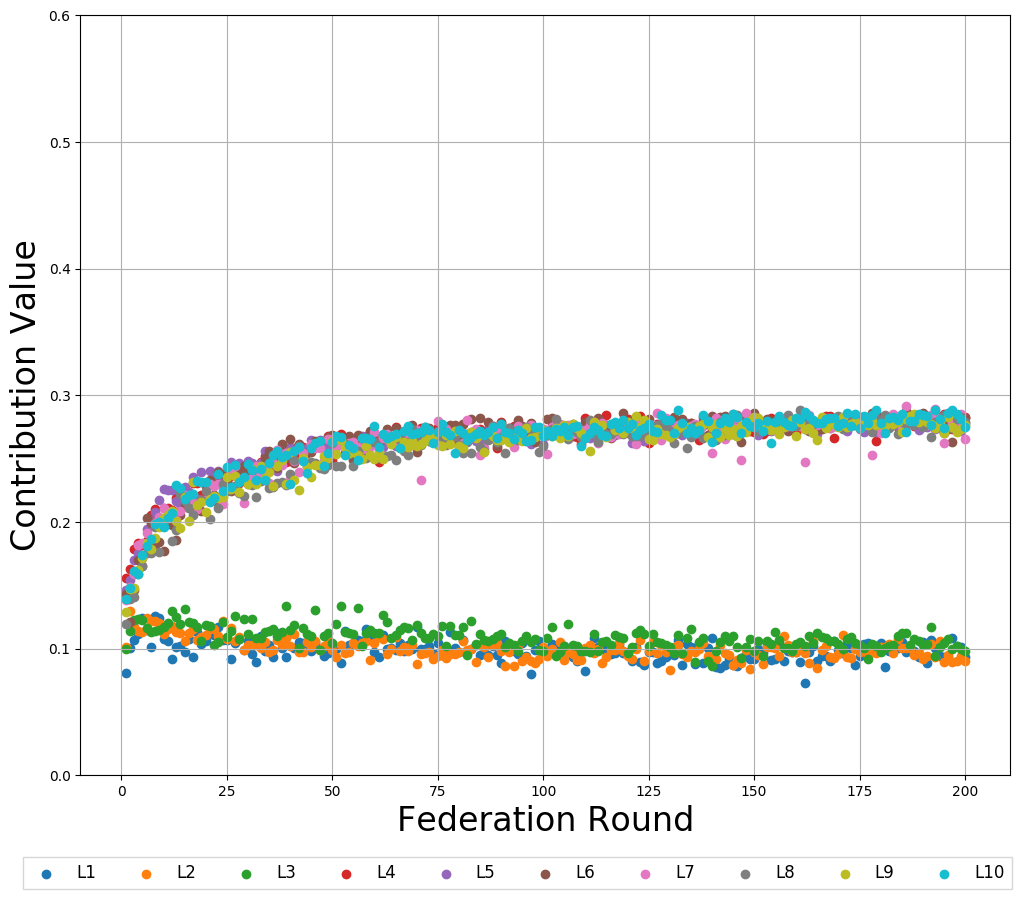}
    \label{subfig:cifar10_powerlawiid_uniformlabelshuffling_microaccuracy_contributionvalue_3learners}
  }
  \subfloat[DVW-MacroAccuracy]{
  \centering
    \includegraphics[width=0.33\linewidth]{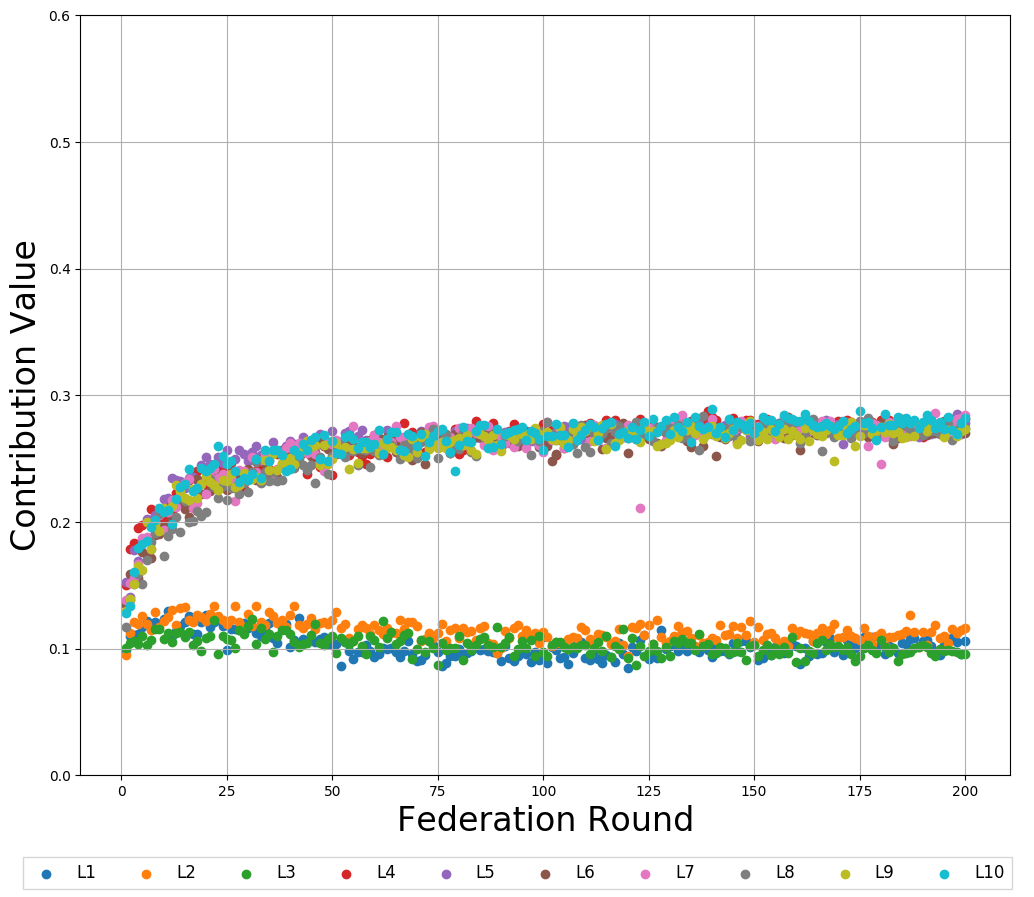}
    \label{subfig:cifar10_powerlawiid_uniformlabelshuffling_macroaccuracy_contributionvalue_3learners}
  }
  \subfloat[DVW-GMean]{
  \centering
    \includegraphics[width=0.33\linewidth]{plots/UniformLabelShuffling/PoliciesContributionValueByLearner/Cifar10_PowerLawIID_UniformLabelShuffling_3LearnersL1L2L3_LearnersContributionValue_DVWGMean0001.png}
    \label{subfig:cifar10_powerlawiid_uniformlabelshuffling_gmean_contributionvalue_3learners}
  }
  \caption{Learners contribution value for the \textbf{Uniform Label Shuffling} attack in the \textbf{PowerLaw \& IID} learning environment, \textbf{with 3 corrupted learners.}}
  \label{fig:cifar10_powerlawiid_uniformlabelshuffling_contributionvalue_3learners}
\end{figure}

\begin{figure}[htpb]  
  \subfloat[DVW-MicroAccuracy]{
    \includegraphics[width=0.33\linewidth]{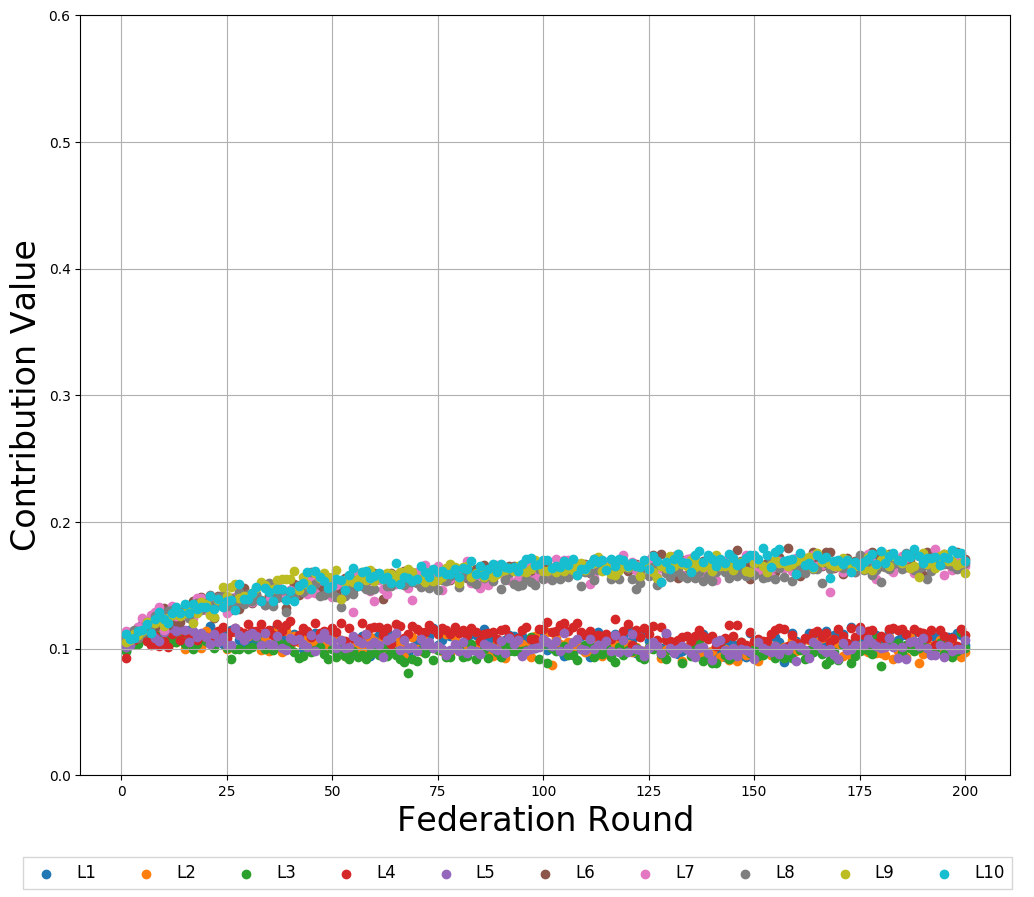}
    \label{subfig:cifar10_powerlawiid_uniformlabelshuffling_microaccuracy_contributionvalue_5learners}
  }
  \subfloat[DVW-MacroAccuracy]{
  \centering
    \includegraphics[width=0.33\linewidth]{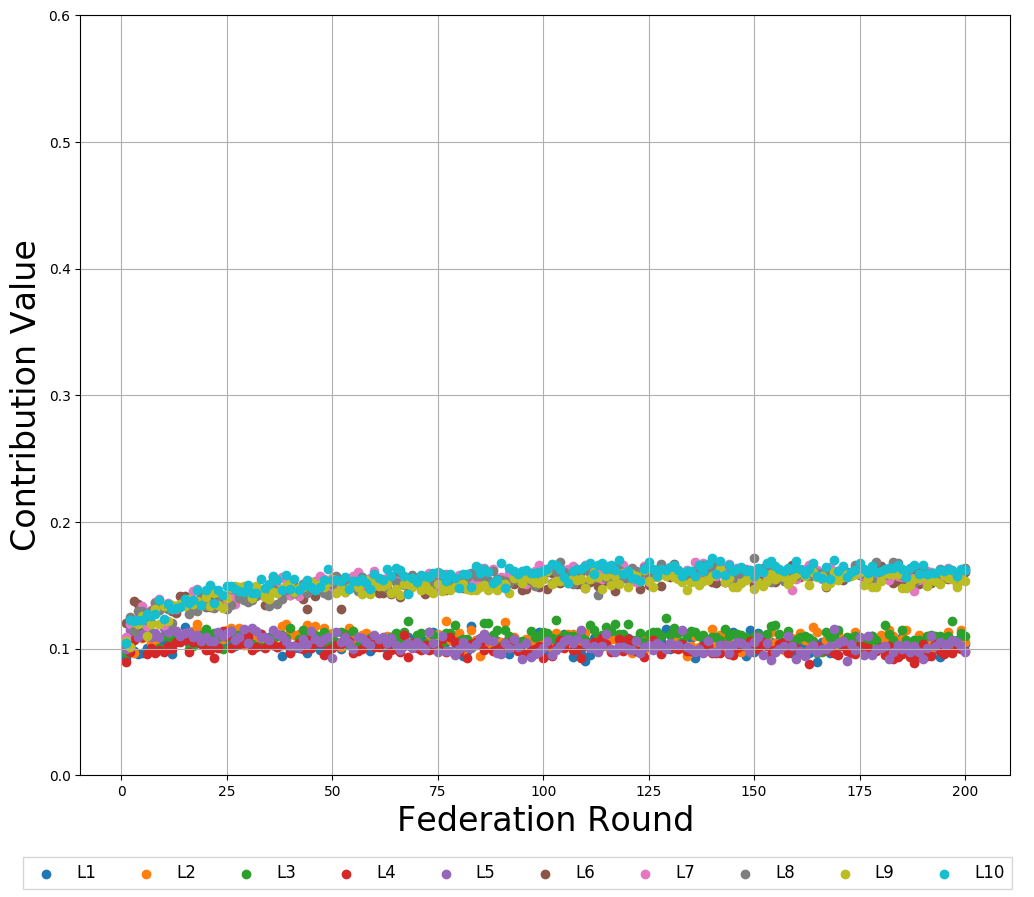}
    \label{subfig:cifar10_powerlawiid_uniformlabelshuffling_macroaccuracy_contributionvalue_5learners}
  }
  \subfloat[DVW-GMean]{
  \centering
    \includegraphics[width=0.33\linewidth]{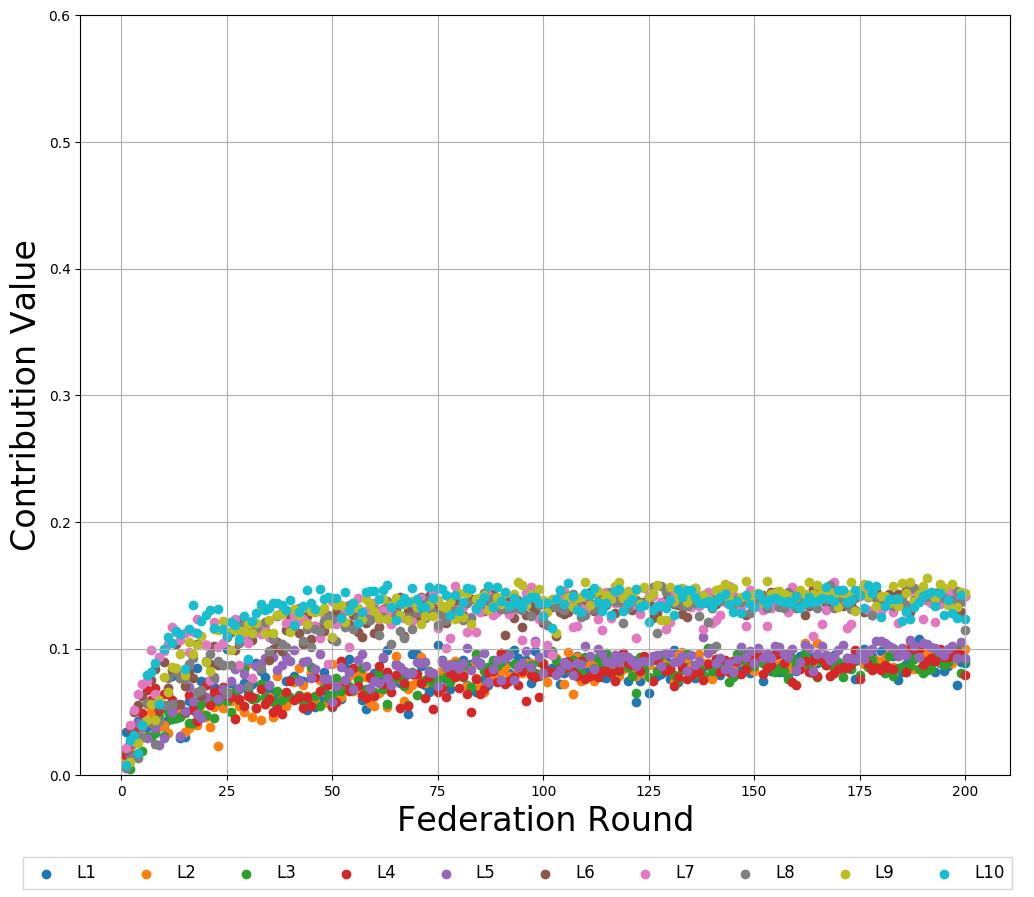}
    \label{subfig:cifar10_powerlawiid_uniformlabelshuffling_gmean_contributionvalue_5learners}
  } 
  \caption{Learners contribution value for the \textbf{Uniform Label Shuffling} attack in the \textbf{PowerLaw \& IID} learning environment, \textbf{with 5 corrupted learners.}}
  \label{fig:cifar10_powerlawiid_uniformlabelshuffling_contributionvalue_5learners}
\end{figure}

In \cref{fig:cifar10_powerlawiid_uniformlabelshuffling_contributionvalue_1learner,fig:cifar10_powerlawiid_uniformlabelshuffling_contributionvalue_3learners,fig:cifar10_powerlawiid_uniformlabelshuffling_contributionvalue_5learners} we show the contribution/weighting value of each learner in the federation for every learning environment for the three different performance metrics. Similar to the uniform and iid environments, while all metrics are able to distinguish corrupted from non-corrupted learners from very early stages of the federated training, the Geometric Mean progressively also downgrades the performance of the corrupted learners.

\newpage %

\subsubsection{Learners Performance Per Class} \label{appendix:uniform_label_shuffling_powerlaw_iid_learners_performance_perclass}

\begin{figure}[htpb]

  \subfloat[DVW-MicroAccuracy]{
    \includegraphics[width=0.33\linewidth]{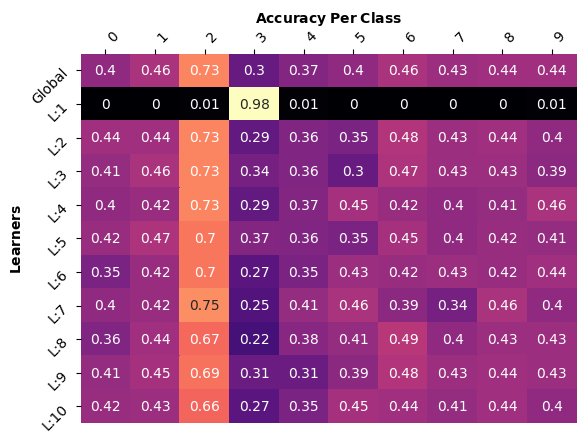}
    \label{subfig:cifar10_powerlawiid_uniformlabelshuffling_microaccuracy_performanceperclass_1learner}
  }
  \subfloat[DVW-MacroAccuracy]{
    \includegraphics[width=0.33\linewidth]{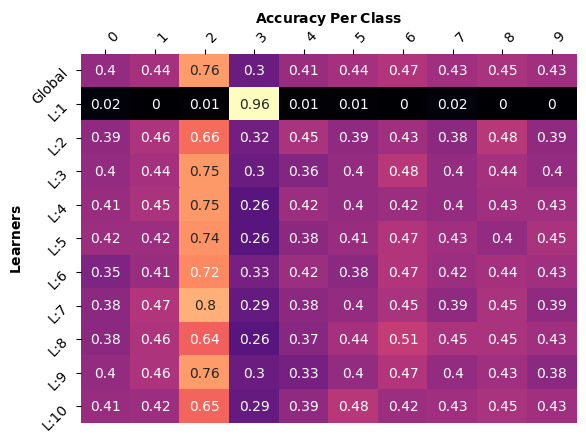}
    \label{subfig:cifar10_powerlawiid_uniformlabelshuffling_macroaccuracy_performanceperclass_1learner}
  }
  \subfloat[DVW-GMean]{
    \includegraphics[width=0.33\linewidth]{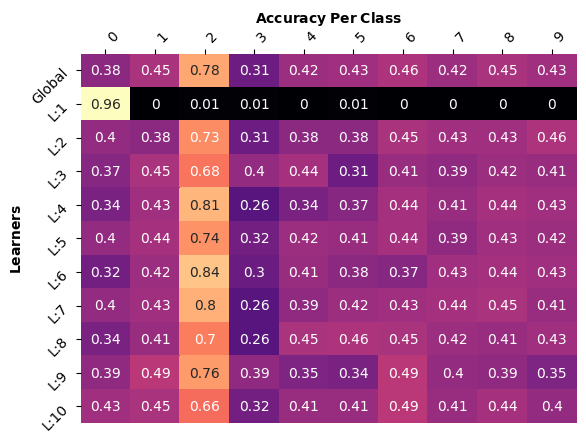}
    \label{subfig:cifar10_powerlawiid_uniformlabelshuffling_gmean_performanceperclass_1learner}
  }
  \caption{Accuracy per class in the last federation round for the community (global) model and each learner for the \textbf{Uniform Label Shuffling} attack in the \textbf{PowerLaw \& IID} learning environment, with \textbf{1 corrupted learner.}}
  \label{fig:cifar10_powerlawiid_uniformlabelshuffling_performanceperclass_1learner}
\end{figure}

\begin{figure}[htpb]  
  \subfloat[DVW-MicroAccuracy]{
    \includegraphics[width=0.33\linewidth]{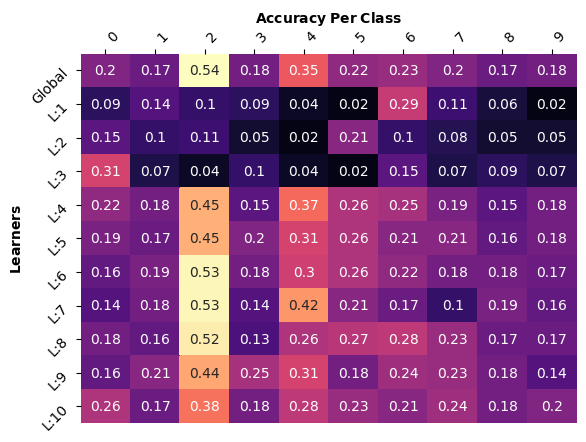}
    \label{subfig:cifar10_powerlawiid_uniformlabelshuffling_microaccuracy_performanceperclass_3learners}
  }
  \subfloat[DVW-MacroAccuracy]{
  \centering
    \includegraphics[width=0.33\linewidth]{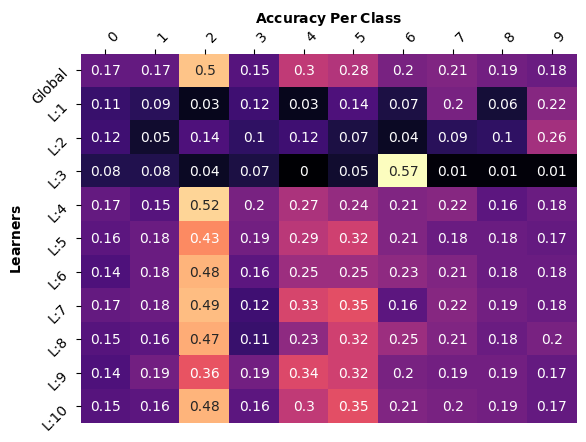}
    \label{subfig:cifar10_powerlawiid_uniformlabelshuffling_macroaccuracy_performanceperclass_3learners}
  }
  \subfloat[DVW-GMean]{
  \centering
    \includegraphics[width=0.33\linewidth]{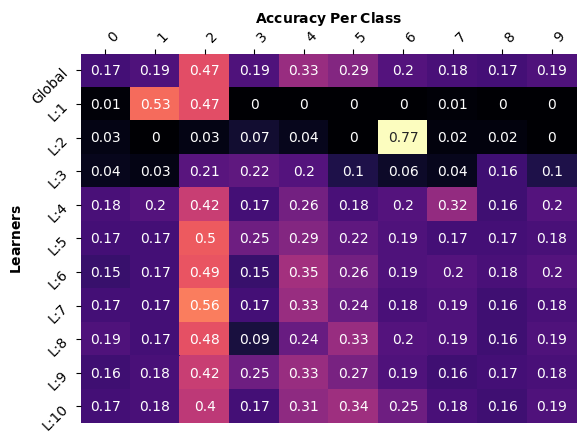}
    \label{subfig:cifar10_powerlawiid_uniformlabelshuffling_gmean_performanceperclass_3learners}
  }
  \caption{Accuracy per class in the last federation round for the community (global) model and each learner for the \textbf{Uniform Label Shuffling} attack in the \textbf{PowerLaw \& IID} learning environment, with \textbf{3 corrupted learners.}}
  \label{fig:cifar10_powerlawiid_uniformlabelshuffling_performanceperclass_3learners}
\end{figure}

\begin{figure}[htpb]
  \subfloat[DVW-MicroAccuracy]{
  \centering
    \includegraphics[width=0.33\linewidth]{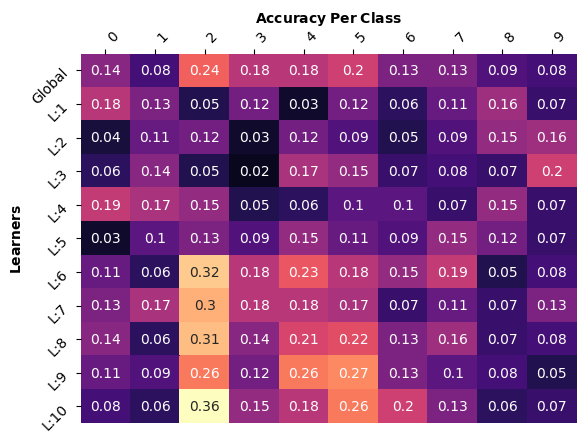}
    \label{subfig:cifar10_powerlawiid_uniformlabelshuffling_microaccuracy_performanceperclass_5learners}
  }
  \subfloat[DVW-MacroAccuracy]{
  \centering
    \includegraphics[width=0.33\linewidth]{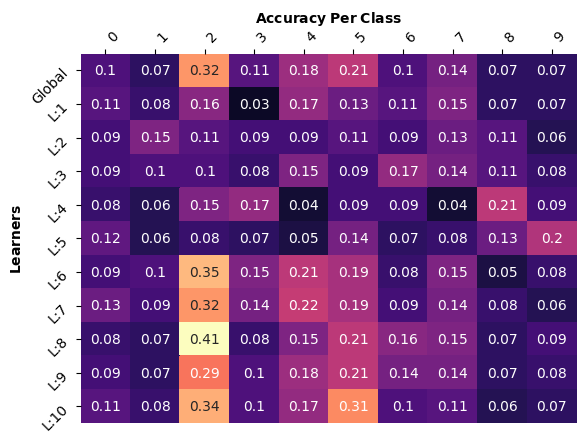}
    \label{subfig:cifar10_powerlawiid_uniformlabelshuffling_macroaccuracy_performanceperclass_5learners}
  }
  \subfloat[DVW-GMean]{
  \centering
    \includegraphics[width=0.33\linewidth]{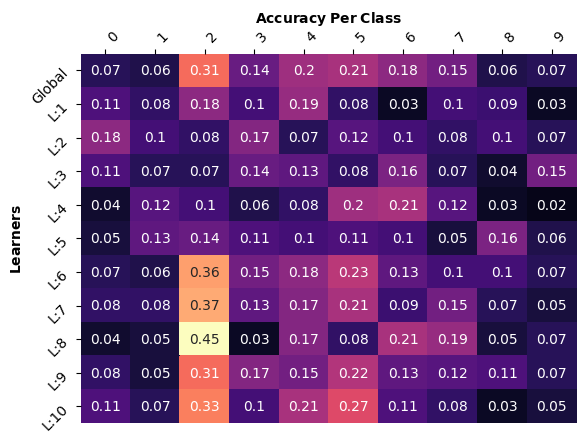}
    \label{subfig:cifar10_powerlawiid_uniformlabelshuffling_gmean_performanceperclass_5learners}
  }
  \caption{Accuracy per class in the last federation round for the community (global) model and each learner for the \textbf{Uniform Label Shuffling} attack in the \textbf{PowerLaw \& IID} learning environment, with \textbf{5 corrupted learners.}}
  \label{fig:cifar10_powerlawiid_uniformlabelshuffling_performanceperclass_5learners}
\end{figure}

In \cref{fig:cifar10_powerlawiid_uniformlabelshuffling_performanceperclass_1learner,fig:cifar10_powerlawiid_uniformlabelshuffling_performanceperclass_3learners,fig:cifar10_powerlawiid_uniformlabelshuffling_performanceperclass_5learners} we present the per-community (global) and per-learner model accuracy for every class of the distributed validation dataset at the final federation round. Similar to the performance per class in the uniform case, the scaling of the colors is different across the weighting schemes' matrices and the distinction between corrupted and non-corrupted learners should be made by inspecting each matrix independently (intra-matrix not inter-matrix).

\newpage %

\section{Targeted Label Flipping} \label{appendix:targeted_label_flipping}
In this corruption mode, the labels of training examples of a specific class are flipped to a target class. Here, we flip airplanes (class:0) to birds (class:2). The experiments we conducted evaluate the performance of the federation over an increasing number of corrupted learners. This learner-based corruption can also be seen as total data and class-level corruption. Similar to the previous corruption mode, for the CIFAR-10 domain we have in total 50000 training examples. In the Uniform \& IID case, every learner accounts for 5000 training examples and holds 500 examples per class (there exist 5000 examples per class). Every time the label of training examples for a particular class is flipped, then the ratio of corruption in terms of class-level corruption is increased by 10\% (500/5000) and the total amount of data corruption is similarly increased by 1\% (500/5000). For instance, in the case of 3 corrupted learners the class-level corruption is equal to 30\% (1500/5000), while the total amount of data corruption is 3\% (1500/50000). The same percentage increase does not hold though for the PowerLaw \& IID case. For this learning environment, the distribution of training examples for class 0 across learners is: $\{L1:1700, L2:1132, L3:754, L4:503, L5:335, L6:224, L7:149, L8:100, L9:67, L10:36\}$. Therefore, for 1 corrupted learner, the class-level and total amount of data corruption is equal to 34\% (1700/5000) and 3.4\% (1700/50000), respectively, while for 3 corrupted learners, the class-level corruption is equal to 72\% ((1700+1132+754)/5000) and 7.2\% ((1700+1132+754)/50000).

\subsection{Uniform \& IID} \label{appendix:targeted_label_flipping_uniform_iid}

\subsubsection{Federation Convergence} \label{appendix:targeted_label_flipping_uniform_iid_federation_convergence}
\begin{figure}[htpb]
\centering
  \subfloat[1 corrupted learner]{
    \includegraphics[width=0.45\linewidth]{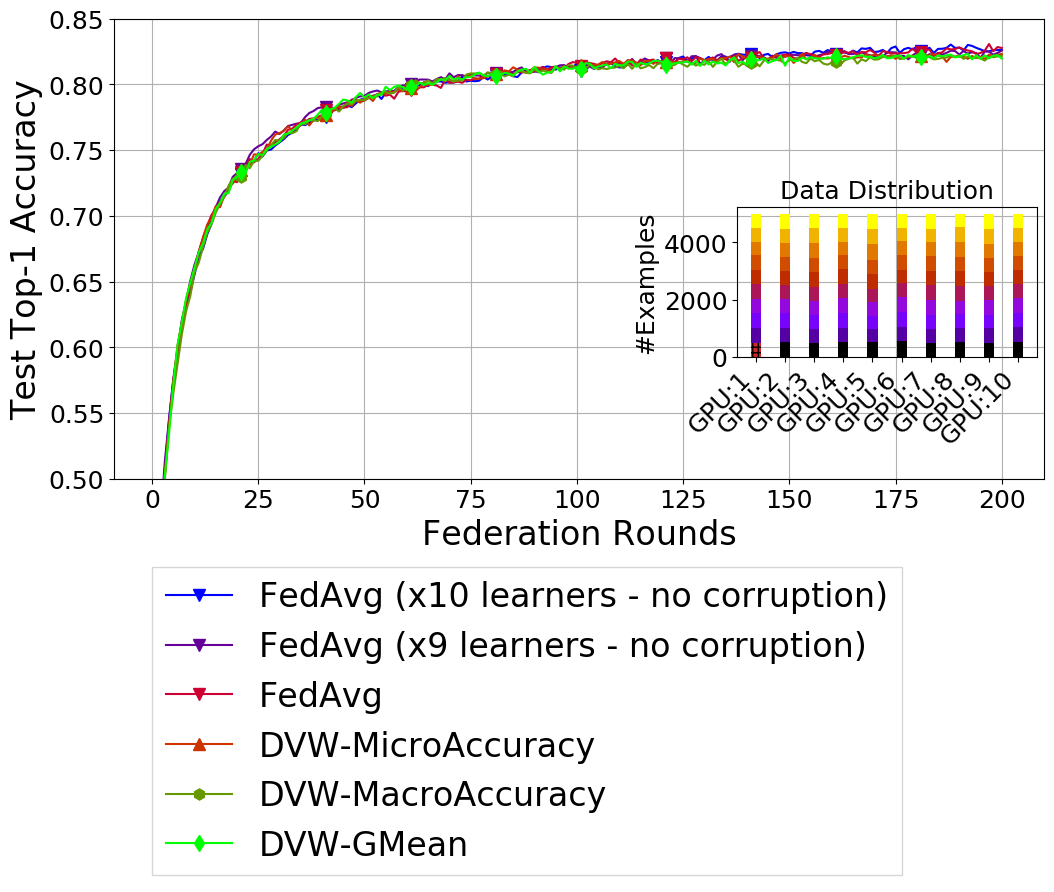}
    \label{subfig:cifar10_uniformiid_targetedlabelflipping0to2_convergence_1learner}
  }
  \subfloat[3 corrupted learners]{
    \includegraphics[width=0.45\linewidth]{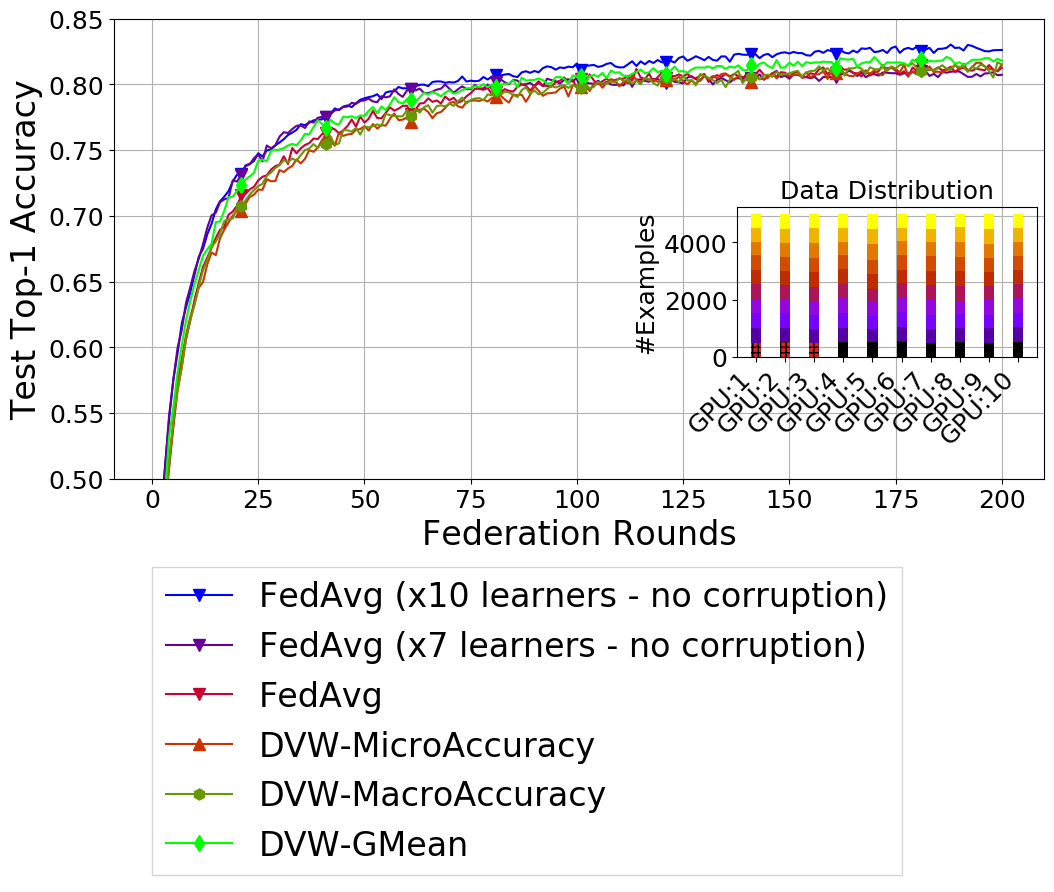}
    \label{subfig:cifar10_uniformiid_targetedlabelflipping0to2_convergence_3learners}
  }
  
  \subfloat[5 corrupted learners]{
    \includegraphics[width=0.45\linewidth]{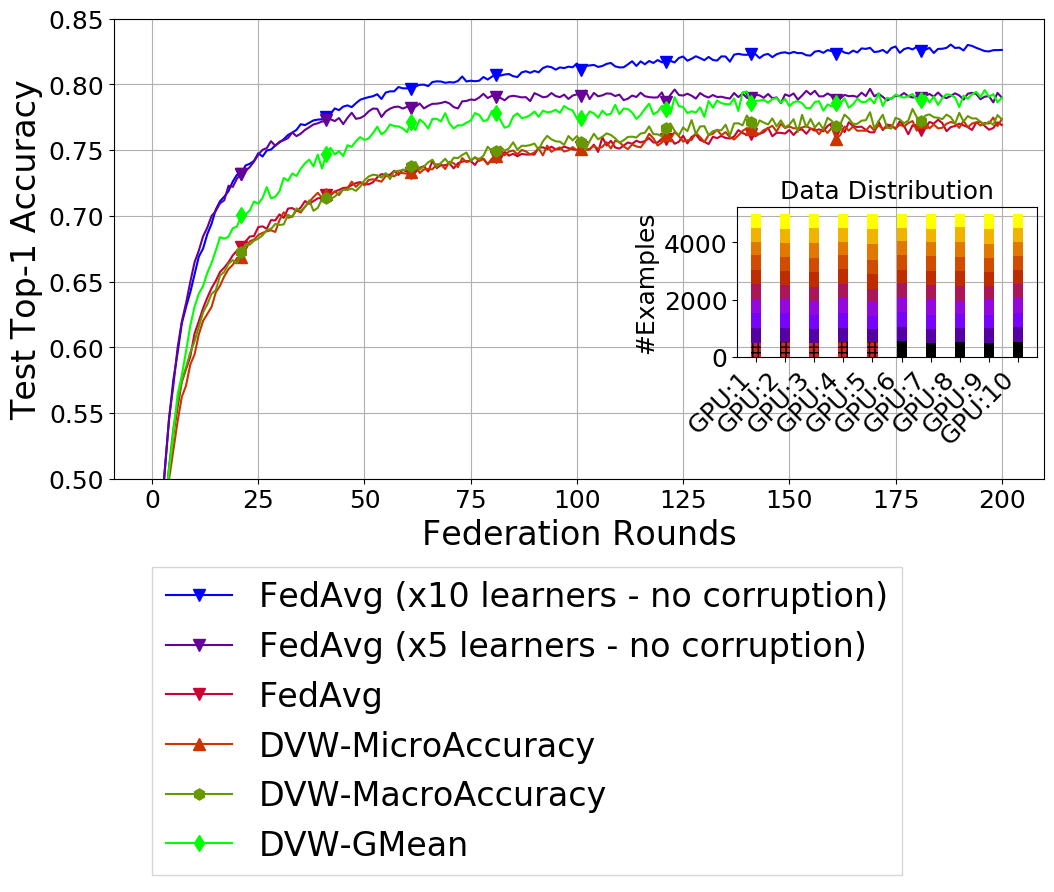}
    \label{subfig:cifar10_uniformiid_targetedlabelflipping0to2_convergence_5learners}
  }
  \subfloat[6 Corrupted learners]{
    \includegraphics[width=0.45\linewidth]{plots/TargetedlabelFlipping/PoliciesConvergence/Cifar10_UniformIID_TargetedLabelFlipping0to2_6Learners_L1L2L3L4L5L6_VanillaSGD_PoliciesConvergence.png}
    \label{subfig:cifar10_uniformiid_targetedlabelflipping0to2_convergence_6learners}
  }
  
  \subfloat[8 corrupted learners]{
  \centering
    \includegraphics[width=0.45\linewidth]{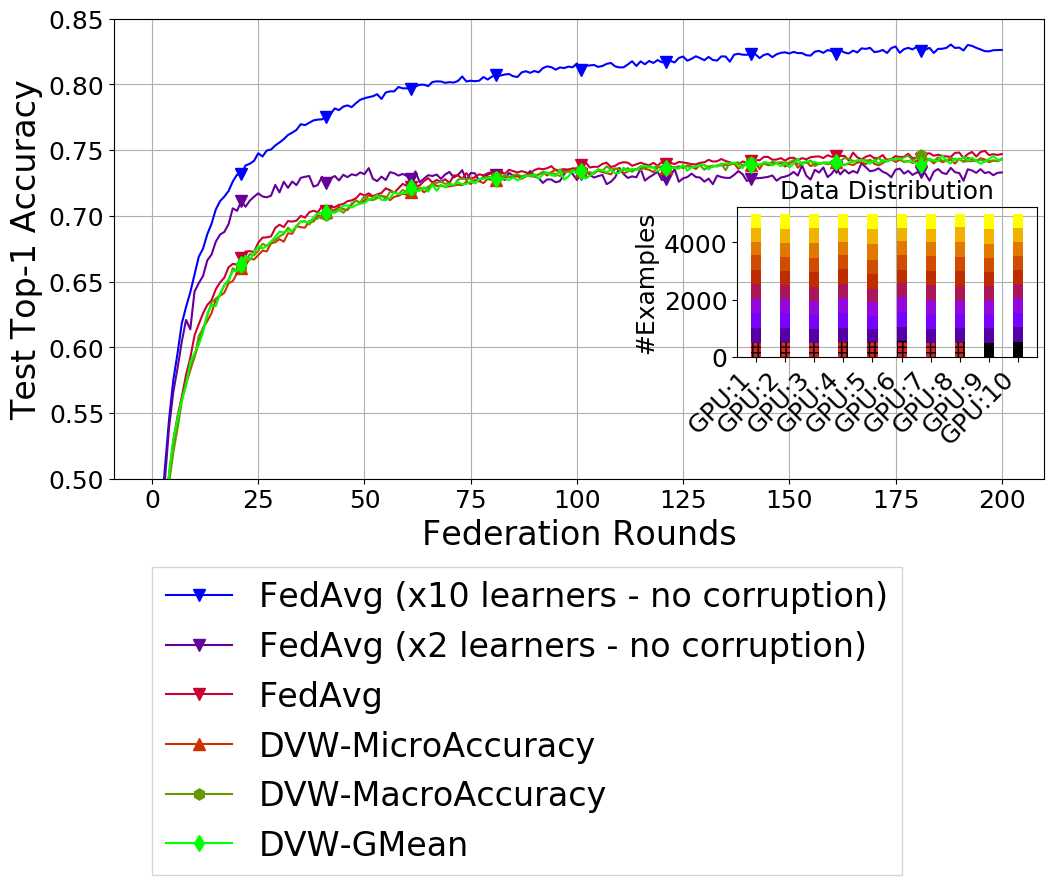}
    \label{subfig:cifar10_uniformiid_targetedlabelflipping0to2_convergence_8learners}
  }
  
  \caption{\textbf{Federation convergence for the Targeted Label Flipping  data poisoning attack in the Uniform \& IID learning environment.} Source class is airplane and target class is bird. Federation performance is measured on the test top-1 accuracy over an increasing number of corrupted learners. Corrupted learners' flipped examples are marked with the red hatch within the data distribution inset. For every learning environment, we include the convergence of the federation with no corruption (10 honest learners) and with exclusion of the corrupted learners (x honest learners). We also present the convergence of the federation for FedAvg (baseline) and different performance weighting aggregation schemes, Micro-Accuracy, Macro-Accuracy and Geometric-Mean. Geometric Mean demonstrates the highest resiliency against label shuffling poisoning.}
  \label{fig:cifar10_uniformiid_targetedlabelflipping0to2_federation_convergence}
\end{figure}

Figure \ref{fig:cifar10_uniformiid_targetedlabelflipping0to2_federation_convergence} demonstrates the federation convergence rate for the targeted label flipping in the uniform and iid learning environments. FedAvg underperforms compared to its performance weighting counterparts especially in the case of 50\% and 60\% corruption. Interestingly in these two corrupted environments, compared to Micro- and Macro-Accuracy, the Geometric Mean can reach higher generalization levels, converge faster and result in a similar performance with the federation environment that has excluded the corrupted learners.

\newpage %

\subsubsection{Learners Contribution Value} \label{appendix:targeted_label_flipping_uniform_iid_learners_contribution_value}

\begin{figure}[htpb]
  \subfloat[DVW-MicroAccuracy]{
  \centering
    \includegraphics[width=0.33\linewidth]{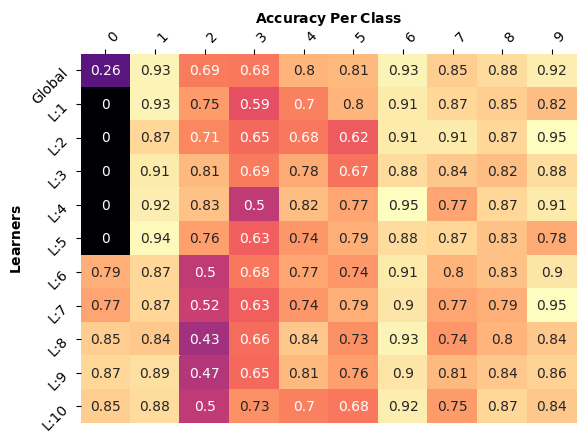}
    \label{subfig:cifar10_uniformiid_targetedlabelflipping_microaccuracy_performanceperclass_5learners}
  }
  \subfloat[DVW-MacroAccuracy]{
  \centering
    \includegraphics[width=0.33\linewidth]{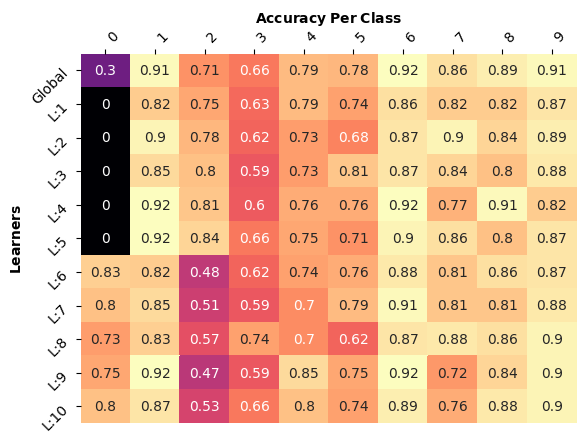}
    \label{subfig:cifar10_uniformiid_targetedlabelflipping_macroaccuracy_performanceperclass_5learners}
  }
  \subfloat[DVW-GMean]{
  \centering
    \includegraphics[width=0.33\linewidth]{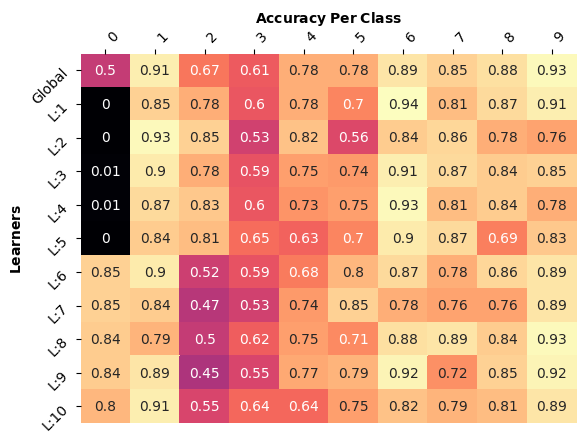}
    \label{subfig:cifar10_uniformiid_targetedlabelflipping_gmean_performanceperclass_5learners}
  }
  \caption{Accuracy per class in the last federation round for the community (global) model and each learner for the \textbf{Targeted Label Flipping} attack in the \textbf{Uniform \& IID} learning environment, with \textbf{5 corrupted learners.}}
  \label{fig:cifar10_uniformiid_targetedlabelflipping_performanceperclass_5learners}
\end{figure}

In \cref{fig:cifar10_uniformiid_targetedlabelflipping_contributionvalue_1learner,fig:cifar10_uniformiid_targetedlabelflipping_contributionvalue_3learners,fig:cifar10_uniformiid_targetedlabelflipping_contributionvalue_5learners,fig:cifar10_uniformiid_targetedlabelflipping_contributionvalue_6learners,fig:cifar10_uniformiid_targetedlabelflipping_contributionvalue_8learners} we demonstrate the contribution/weighting value of each learner in the federation for every learning environment for the three different performance metrics. Interestingly, not all performance weighting schemes are able to distinguish corrupted vs non-corrupted learners in these learning environments. Specifically, compared to Micro-Accuracy, the Macro-Accuracy scheme is able to detect corrupted learners and downgrade them, whereas Micro-Accuracy cannot detect the corruption and equally weights corrupted and non-corrupted learners (except for the environment with 8 corrupted learners). Oppositely, Geometric Mean detects the corrupted learners from the very beginning and preserves the downgraded performance score for these learners throughout the learning process. The higher contribution values for the corrupted learners assigned by Geometric Mean (Figures \cref{subfig:cifar10_uniformiid_targetedlabelflipping_gmean_contributionvalue_1learner,subfig:cifar10_uniformiid_targetedlabelflipping_gmean_contributionvalue_3learners,subfig:cifar10_uniformiid_targetedlabelflipping_gmean_contributionvalue_5learners,subfig:cifar10_uniformiid_targetedlabelflipping_gmean_contributionvalue_6learners}) can be attributed to the case that some of the corrupted learners can provide randomly correct predictions for class 0 and therefore the Geometric Mean does not assign the $0.001$ error value (see the performance per class for learners L3 and L4 in plot \ref{subfig:cifar10_uniformiid_targetedlabelflipping_gmean_performanceperclass_5learners}).

\begin{figure}[htpb]
  \subfloat[DVW-MicroAccuracy]{
    \includegraphics[width=0.33\linewidth]{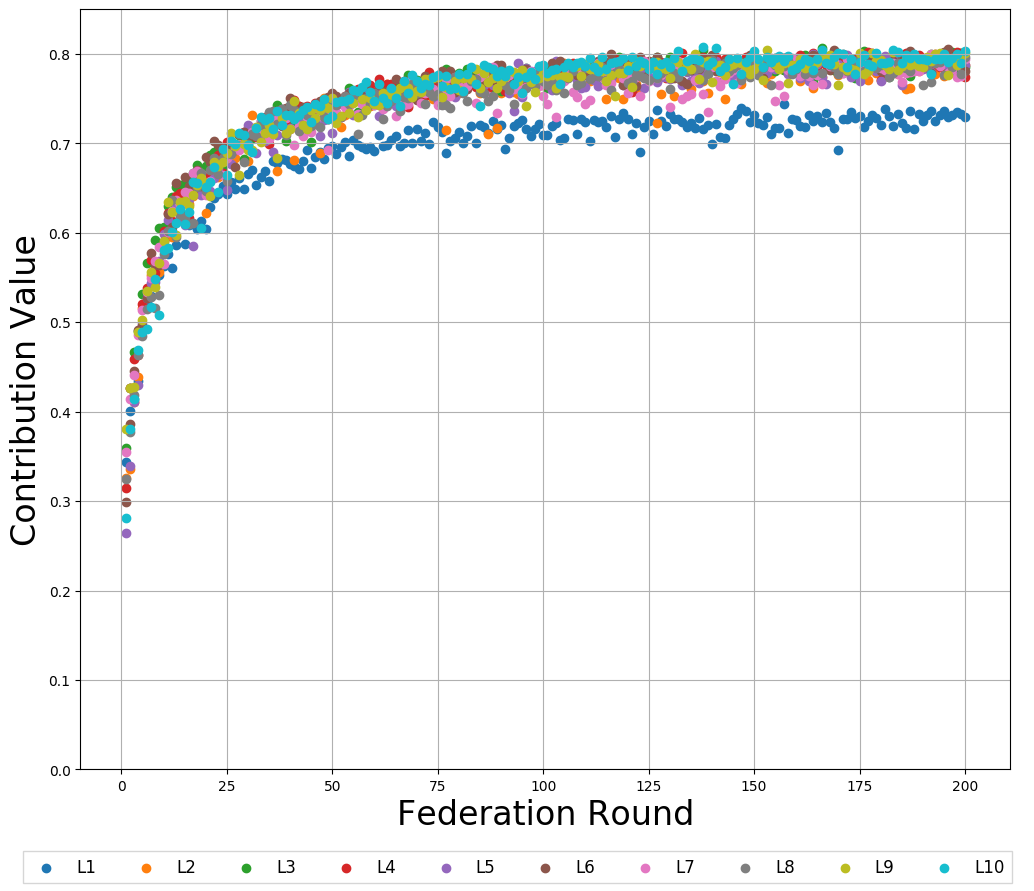}
    \label{subfig:cifar10_uniformiid_targetedlabelflipping_microaccuracy_contributionvalue_1learner}
  }
  \subfloat[DVW-MacroAccuracy]{
    \includegraphics[width=0.33\linewidth]{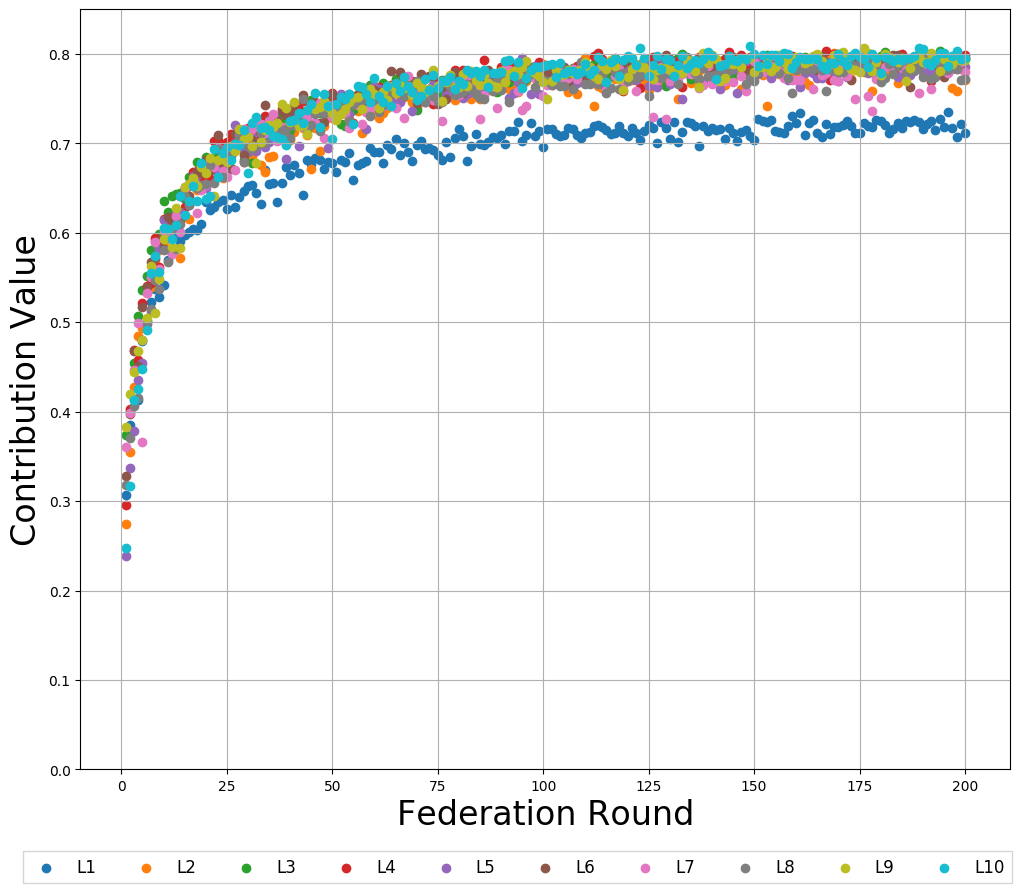}
    \label{subfig:cifar10_uniformiid_targetedlabelflipping_macroaccuracy_contributionvalue_1learner}
  }
  \subfloat[DVW-GMean]{
    \includegraphics[width=0.33\linewidth]{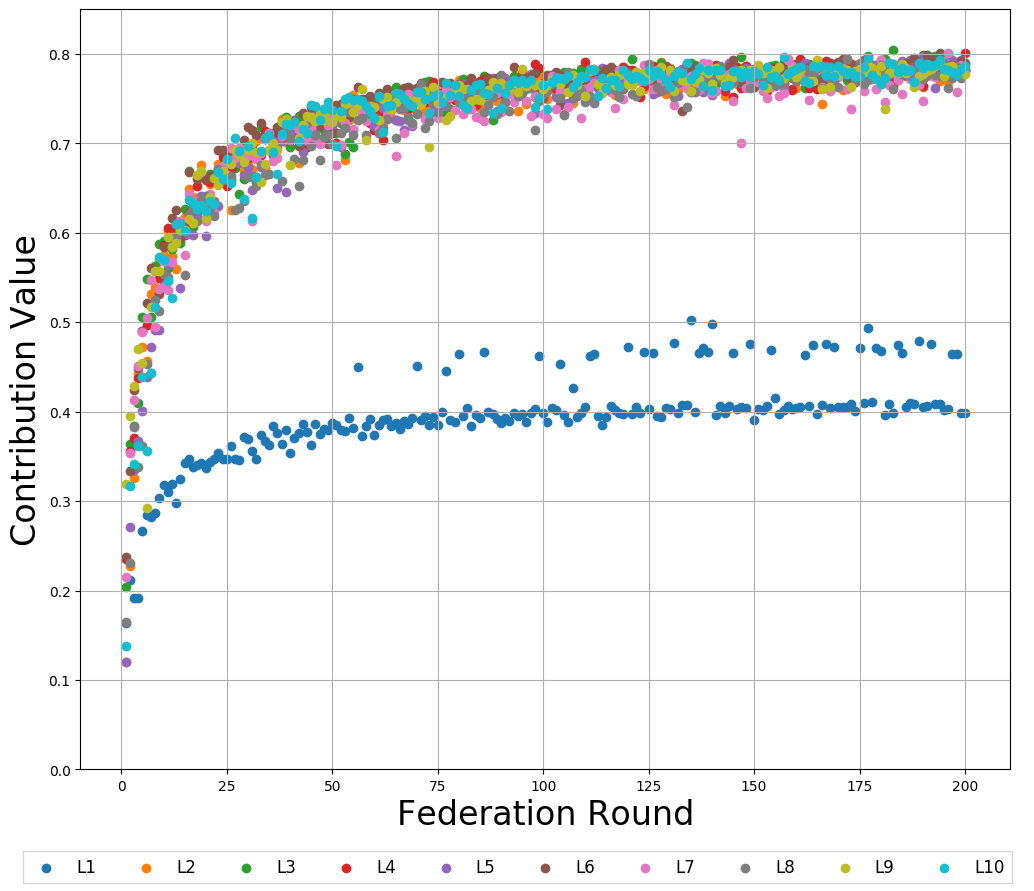}
    \label{subfig:cifar10_uniformiid_targetedlabelflipping_gmean_contributionvalue_1learner}
  }
  \caption{Learners contribution value for the \textbf{Targeted Label Flipping} attack in the \textbf{Uniform \& IID} learning environment, \textbf{with 1 corrupted learner.}}
  \label{fig:cifar10_uniformiid_targetedlabelflipping_contributionvalue_1learner}
\end{figure}

\begin{figure}[htpb]  
  \subfloat[DVW-MicroAccuracy]{
    \includegraphics[width=0.33\linewidth]{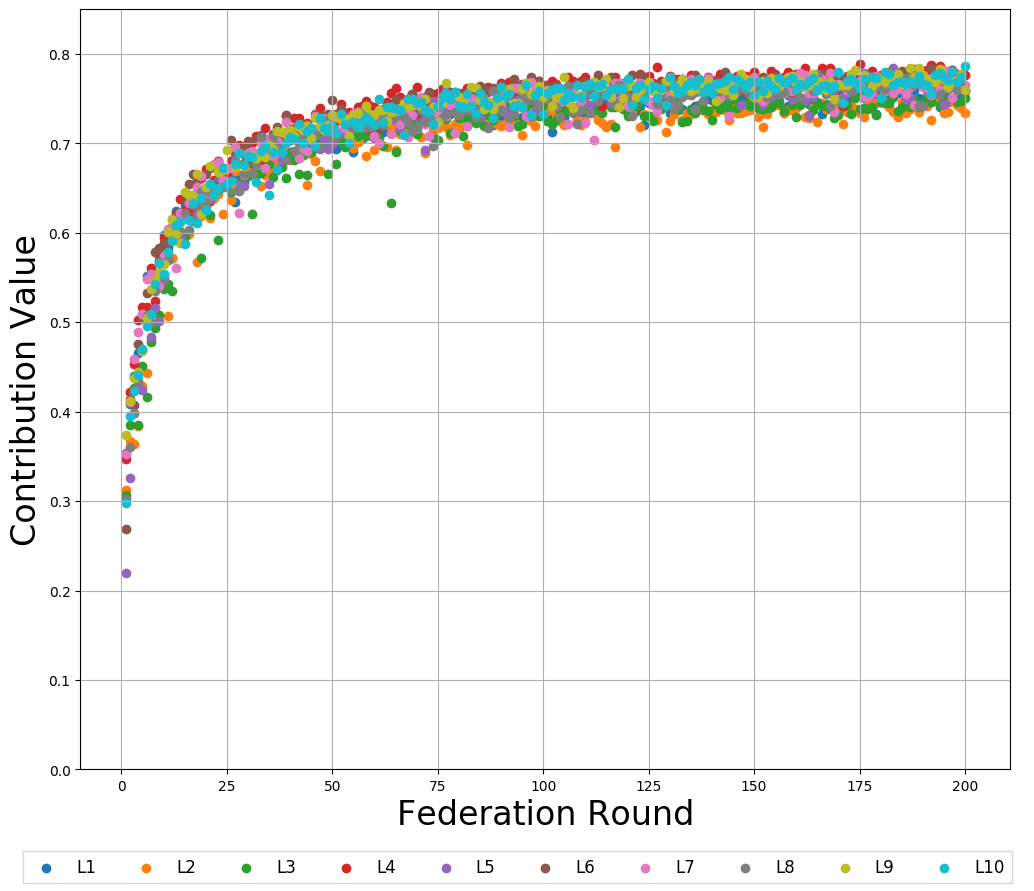}
    \label{subfig:cifar10_uniformiid_targetedlabelflipping_microaccuracy_contributionvalue_3learners}
  }
  \subfloat[DVW-MacroAccuracy]{
  \centering
    \includegraphics[width=0.33\linewidth]{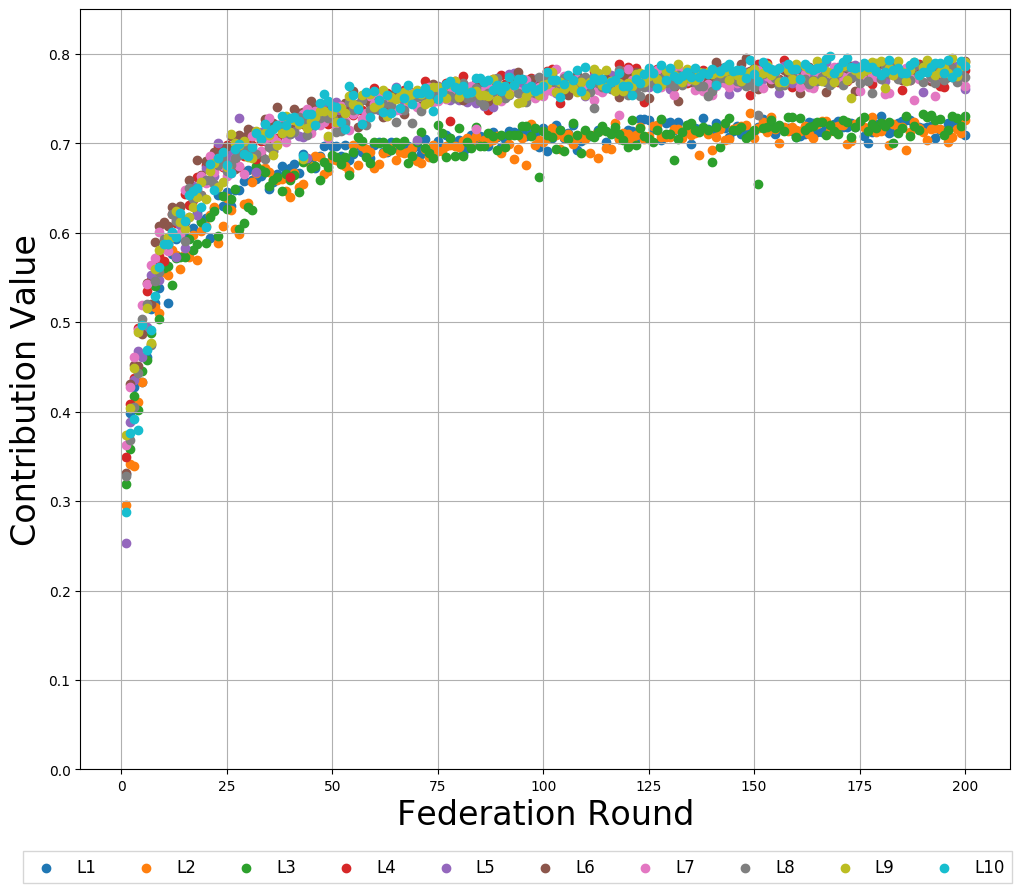}
    \label{subfig:cifar10_uniformiid_targetedlabelflipping_macroaccuracy_contributionvalue_3learners}
  }
  \subfloat[DVW-GMean]{
  \centering
    \includegraphics[width=0.33\linewidth]{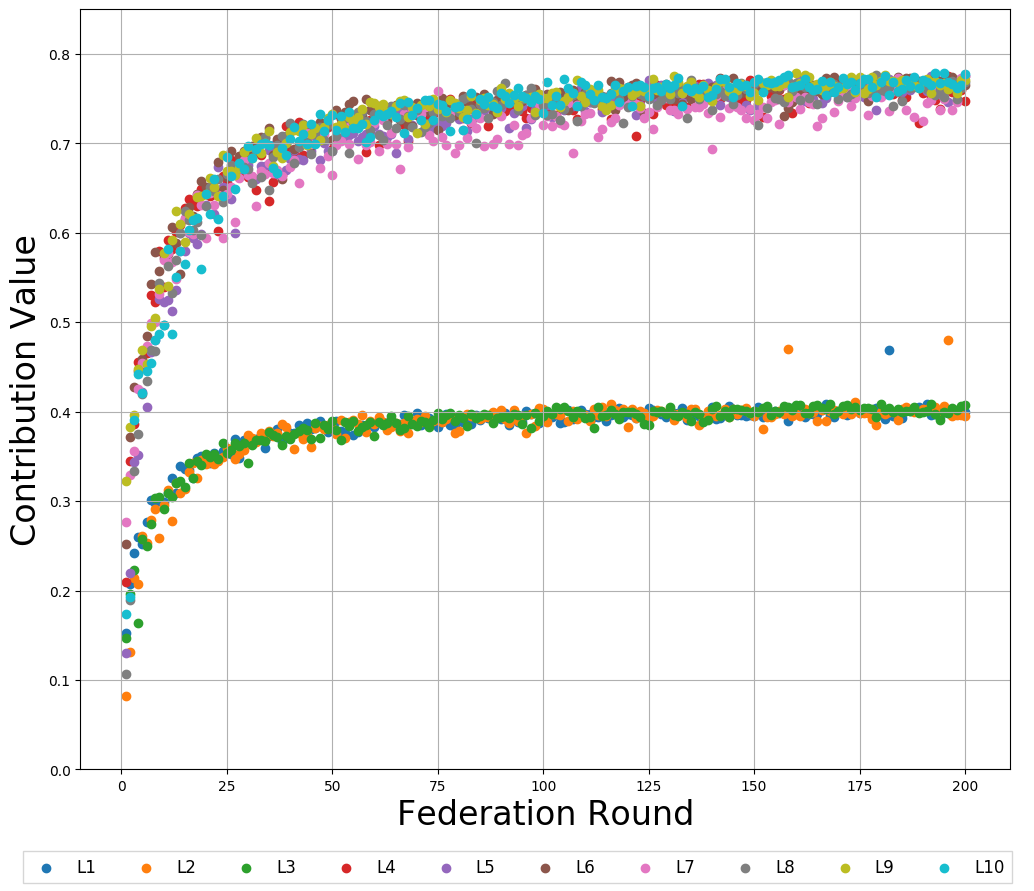}
    \label{subfig:cifar10_uniformiid_targetedlabelflipping_gmean_contributionvalue_3learners}
  }
  \caption{Learners contribution value for the \textbf{Targeted Label Flipping} attack in the \textbf{Uniform \& IID} learning environment, \textbf{with 3 corrupted learners.}}
  \label{fig:cifar10_uniformiid_targetedlabelflipping_contributionvalue_3learners}
\end{figure}

\begin{figure}[htpb]  
  \subfloat[DVW-MicroAccuracy]{
    \includegraphics[width=0.33\linewidth]{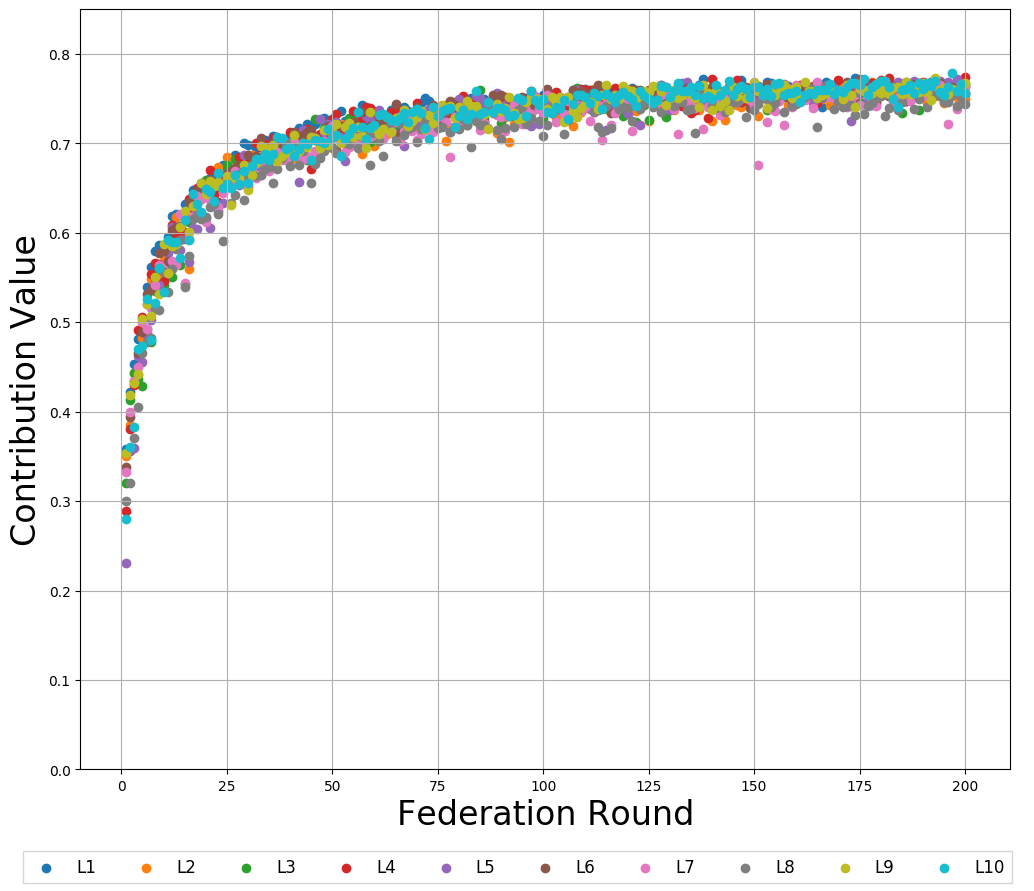}
    \label{subfig:cifar10_uniformiid_targetedlabelflipping_microaccuracy_contributionvalue_5learners}
  }
  \subfloat[DVW-MacroAccuracy]{
  \centering
    \includegraphics[width=0.33\linewidth]{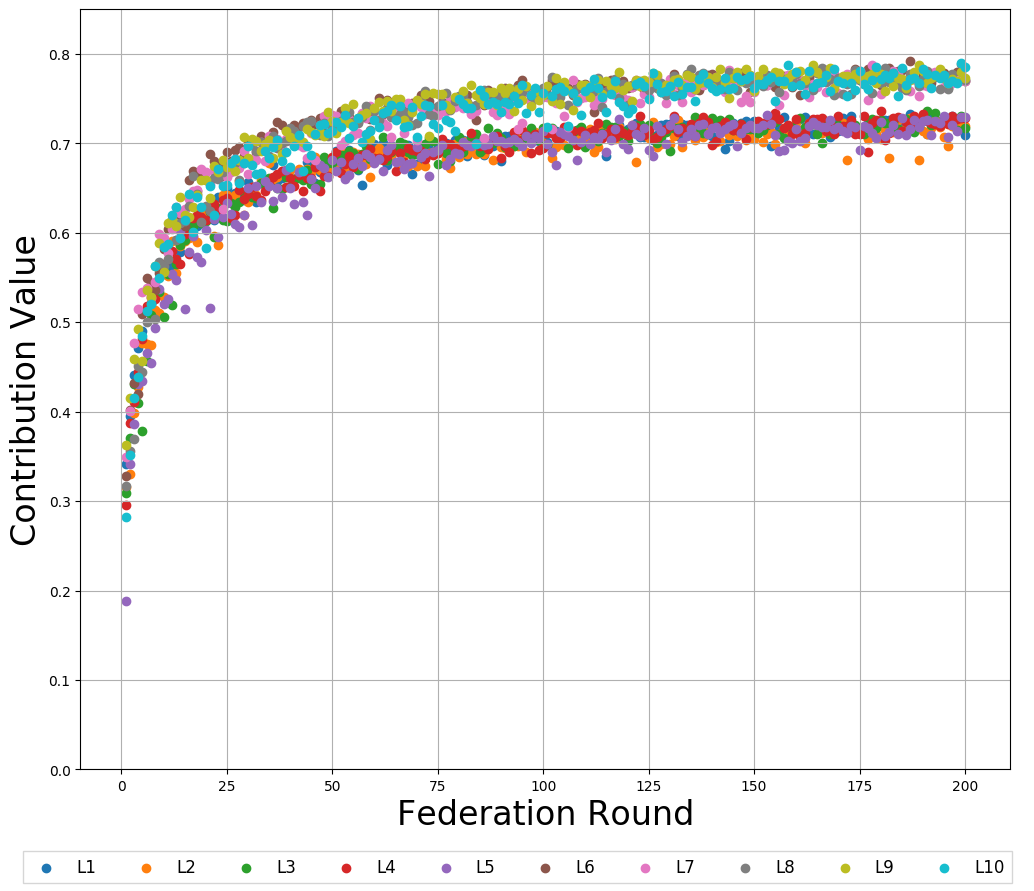}
    \label{subfig:cifar10_uniformiid_targetedlabelflipping_macroaccuracy_contributionvalue_5learners}
  }
  \subfloat[DVW-GMean]{
  \centering
    \includegraphics[width=0.33\linewidth]{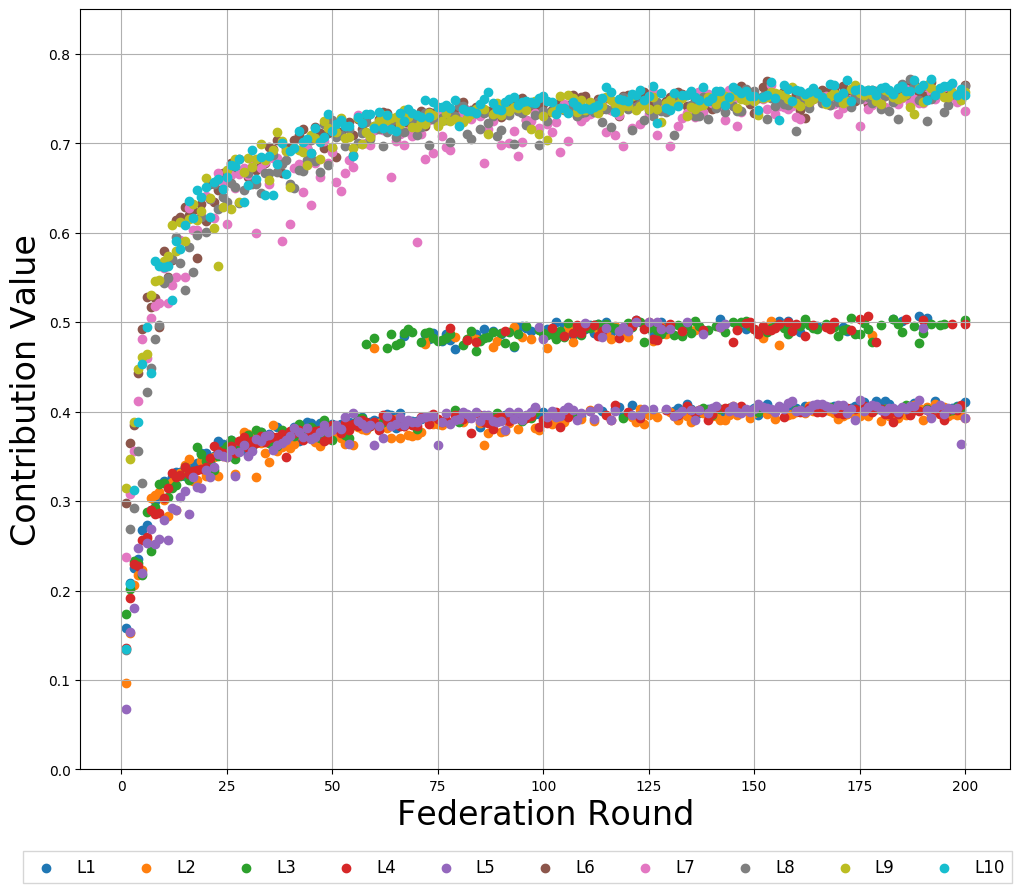}
    \label{subfig:cifar10_uniformiid_targetedlabelflipping_gmean_contributionvalue_5learners}
  }
  \caption{Learners contribution value for the \textbf{Targeted Label Flipping} attack in the \textbf{Uniform \& IID} learning environment, \textbf{with 5 corrupted learners.}}
  \label{fig:cifar10_uniformiid_targetedlabelflipping_contributionvalue_5learners}
\end{figure}

\begin{figure}[htpb]  
  \subfloat[DVW-MicroAccuracy]{
    \includegraphics[width=0.33\linewidth]{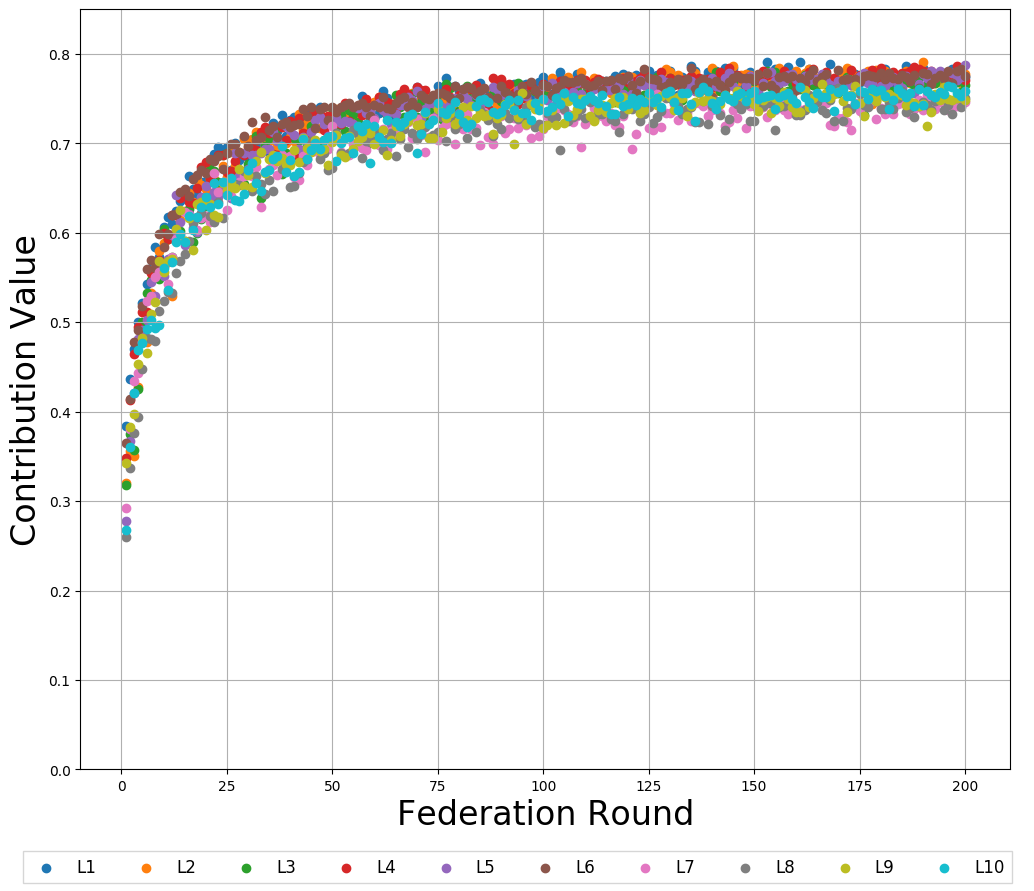}
    \label{subfig:cifar10_uniformiid_targetedlabelflipping_microaccuracy_contributionvalue_6learners}
  }
  \subfloat[DVW-MacroAccuracy]{
  \centering
    \includegraphics[width=0.33\linewidth]{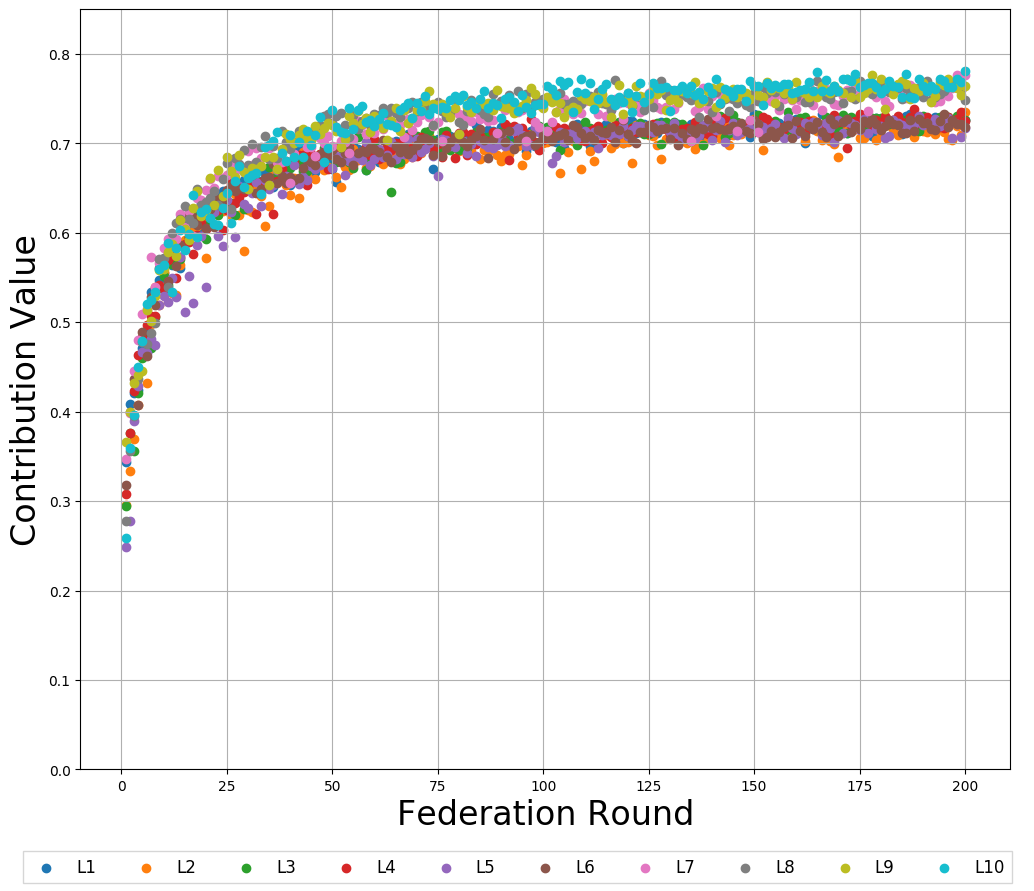}
    \label{subfig:cifar10_uniformiid_targetedlabelflipping_macroaccuracy_contributionvalue_6learners}
  }
  \subfloat[DVW-GMean]{
  \centering
    \includegraphics[width=0.33\linewidth]{plots/TargetedlabelFlipping/PoliciesContributionValueByLearner/Cifar10_UniformIID_TargetedLabelFlipping0to2_6LearnersL1L2L3L4L5L6_LearnersContributionValue_DVWGMean0001.png}
    \label{subfig:cifar10_uniformiid_targetedlabelflipping_gmean_contributionvalue_6learners}
  }
  \caption{Learners contribution value for the \textbf{Targeted Label Flipping} attack in the \textbf{Uniform \& IID} learning environment, \textbf{with 6 corrupted learners.}}
  \label{fig:cifar10_uniformiid_targetedlabelflipping_contributionvalue_6learners}
\end{figure}

\begin{figure}[htpb]  
  \subfloat[DVW-MicroAccuracy]{
    \includegraphics[width=0.33\linewidth]{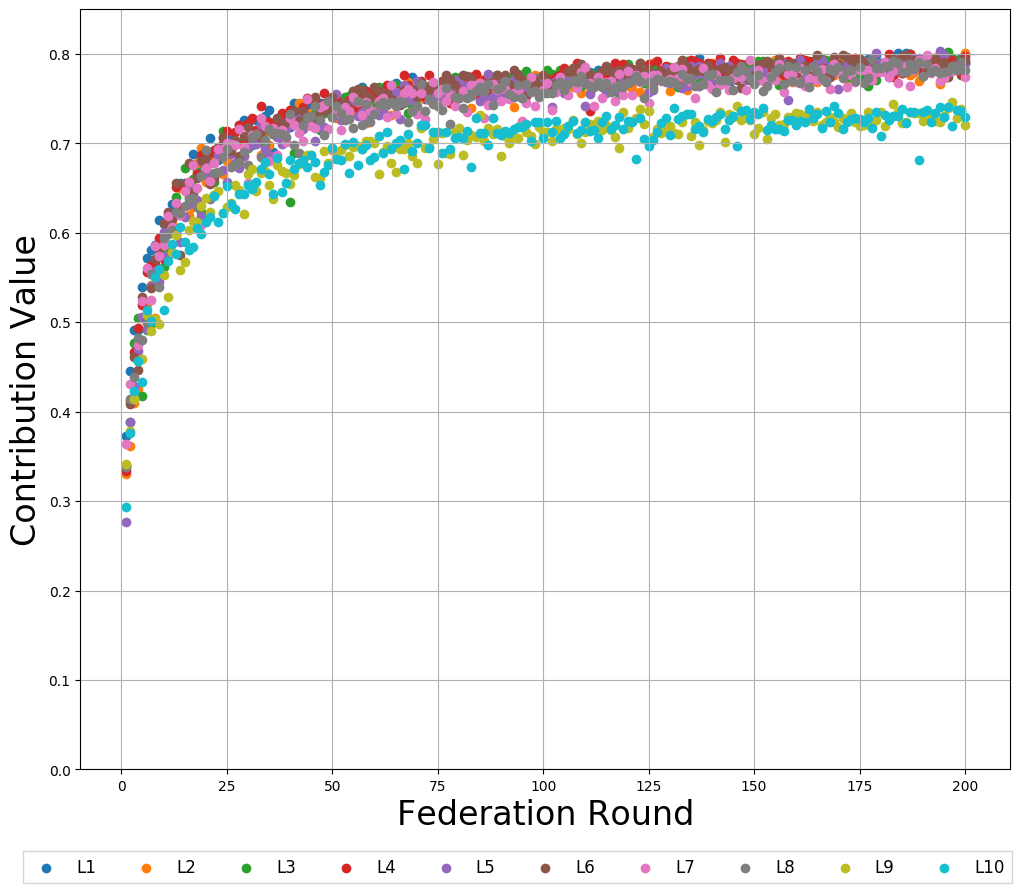}
    \label{subfig:cifar10_uniformiid_targetedlabelflipping_microaccuracy_contributionvalue_8learners}
  }
  \subfloat[DVW-MacroAccuracy]{
  \centering
    \includegraphics[width=0.33\linewidth]{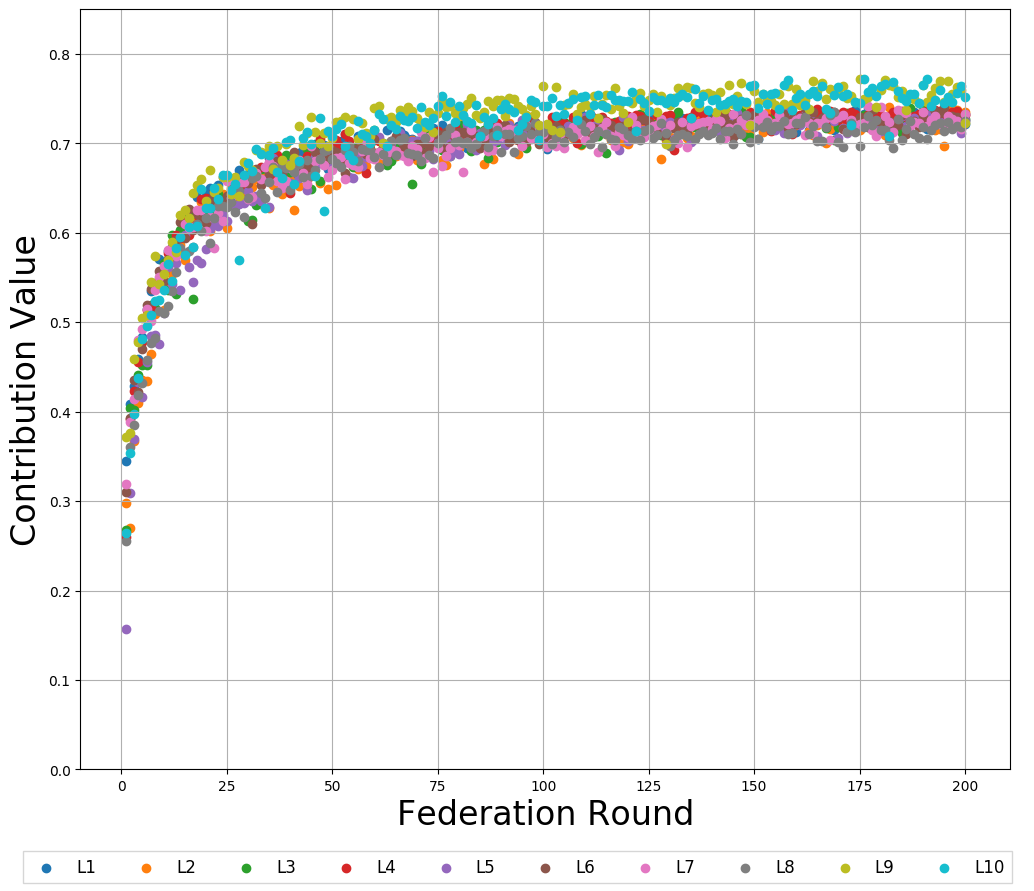}
    \label{subfig:cifar10_uniformiid_targetedlabelflipping_macroaccuracy_contributionvalue_8learners}
  }
  \subfloat[DVW-GMean]{
  \centering
    \includegraphics[width=0.33\linewidth]{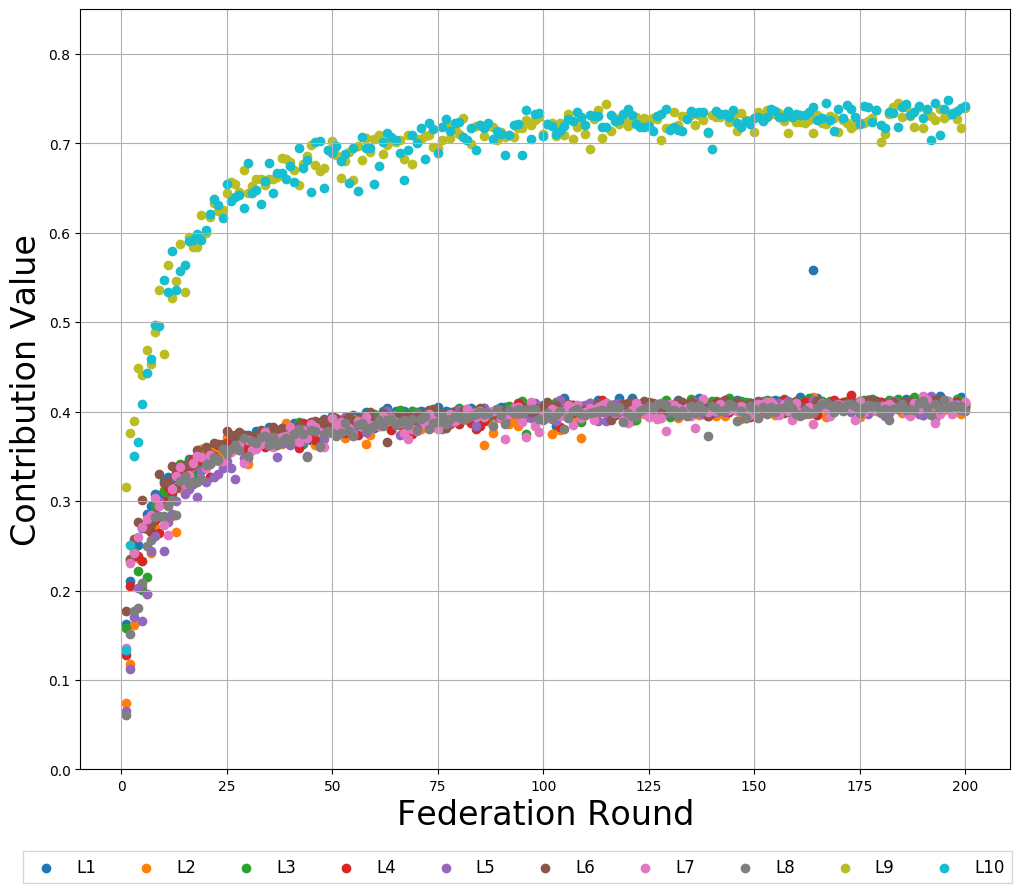}
    \label{subfig:cifar10_uniformiid_targetedlabelflipping_gmean_contributionvalue_8learners}
  }
  \caption{Learners contribution value for the \textbf{Targeted Label Flipping} attack in the \textbf{Uniform \& IID} learning environment, \textbf{with 8 corrupted learners.}}
  \label{fig:cifar10_uniformiid_targetedlabelflipping_contributionvalue_8learners}
\end{figure}

\newpage %

\subsubsection{Learners Performance Per Class} \label{appendix:uniform_targeted_label_flipping_uniformiid_learners_performance_perclass}

In \cref{fig:cifar10_uniformiid_targetedlabelflipping_performanceperclass_1learner,fig:cifar10_uniformiid_targetedlabelflipping_performanceperclass_3learners,fig:cifar10_uniformiid_targetedlabelflipping_performanceperclass_5learners,fig:cifar10_uniformiid_targetedlabelflipping_performanceperclass_6learners,fig:cifar10_uniformiid_targetedlabelflipping_performanceperclass_8learners} we demonstrate the per-community (global) and per-learner model accuracy for every class of the distributed validation dataset at the final federation round. The higher the corruption is the harder it is for the community model to learn class 0. In particular, after 50\% of corruption Micro- and Macro-Accuracy cannot help the federation to accurately learn examples from class 0, while with Geometric Mean the community model can reach 50\% and 30\% accuracy for class 0 when 50\% and 60\% corruption exists, respectively.

\begin{figure}[htpb]

  \subfloat[DVW-MicroAccuracy]{
    \includegraphics[width=0.33\linewidth]{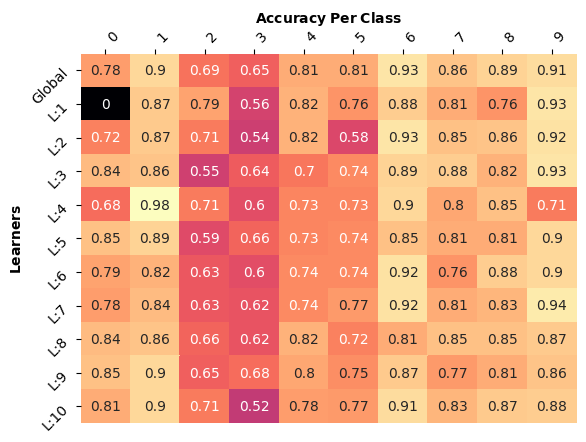}
    \label{subfig:cifar10_uniformiid_targetedlabelflipping_microaccuracy_performanceperclass_1learner}
  }
  \subfloat[DVW-MacroAccuracy]{
    \includegraphics[width=0.33\linewidth]{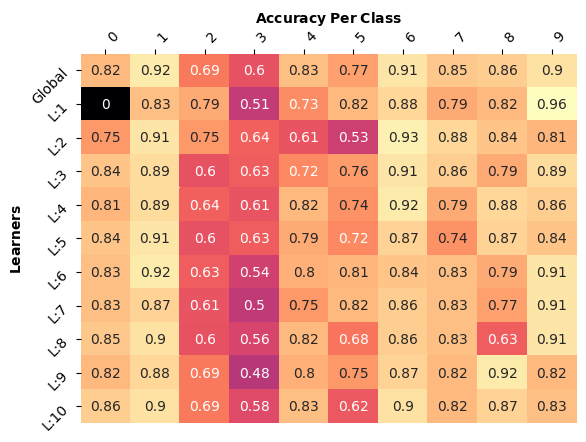}
    \label{subfig:cifar10_uniformiid_targetedlabelflipping_macroaccuracy_performanceperclass_1learner}
  }
  \subfloat[DVW-GMean]{
    \includegraphics[width=0.33\linewidth]{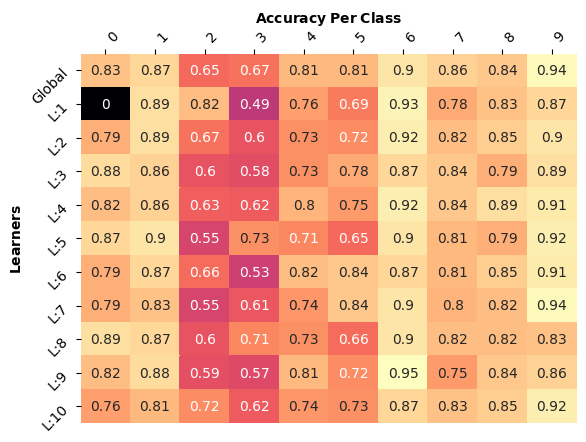}
    \label{subfig:cifar10_uniformiid_targetedlabelflipping_gmean_performanceperclass_1learner}
  }
  \caption{Accuracy per class in the last federation round for the community (global) model and each learner for the \textbf{Targeted Label Flipping} attack in the \textbf{Uniform \& IID} learning environment, with \textbf{1 corrupted learner.}}
  \label{fig:cifar10_uniformiid_targetedlabelflipping_performanceperclass_1learner}
\end{figure}

\begin{figure}[htpb]  
  \subfloat[DVW-MicroAccuracy]{
    \includegraphics[width=0.33\linewidth]{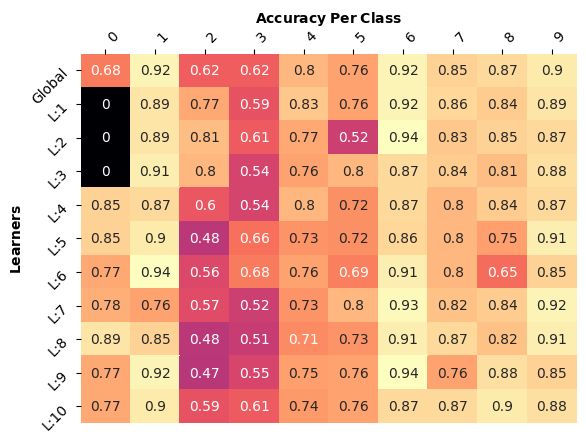}
    \label{subfig:cifar10_uniformiid_targetedlabelflipping_microaccuracy_performanceperclass_3learners}
  }
  \subfloat[DVW-MacroAccuracy]{
  \centering
    \includegraphics[width=0.33\linewidth]{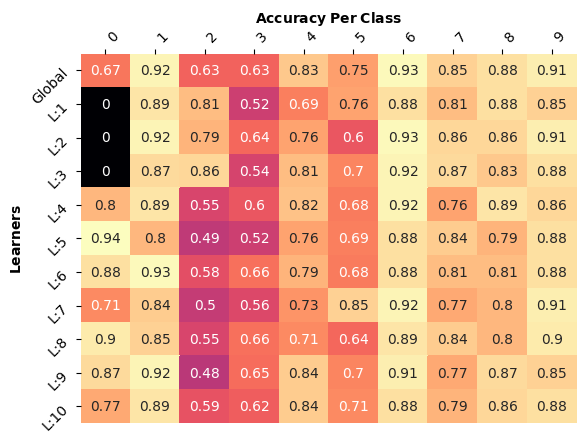}
    \label{subfig:cifar10_uniformiid_targetedlabelflipping_macroaccuracy_performanceperclass_3learners}
  }
  \subfloat[DVW-GMean]{
  \centering
    \includegraphics[width=0.33\linewidth]{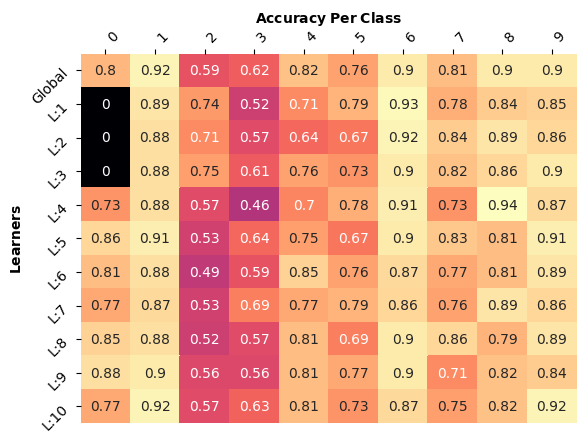}
    \label{subfig:cifar10_uniformiid_targetedlabelflipping_gmean_performanceperclass_3learners}
  }
  \caption{Accuracy per class in the last federation round for the community (global) model and each learner for the \textbf{Targeted Label Flipping} attack in the \textbf{Uniform \& IID} learning environment, with \textbf{3 corrupted learners.}}
  \label{fig:cifar10_uniformiid_targetedlabelflipping_performanceperclass_3learners}
\end{figure}

\begin{figure}[htpb]
  \subfloat[DVW-MicroAccuracy]{
  \centering
    \includegraphics[width=0.33\linewidth]{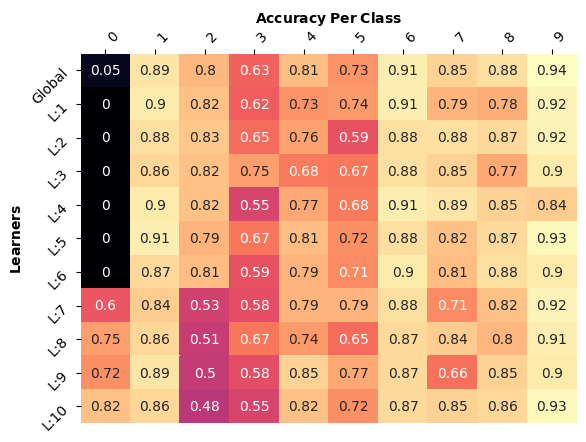}
    \label{subfig:cifar10_uniformiid_targetedlabelflipping_microaccuracy_performanceperclass_6learners}
  }
  \subfloat[DVW-MacroAccuracy]{
  \centering
    \includegraphics[width=0.33\linewidth]{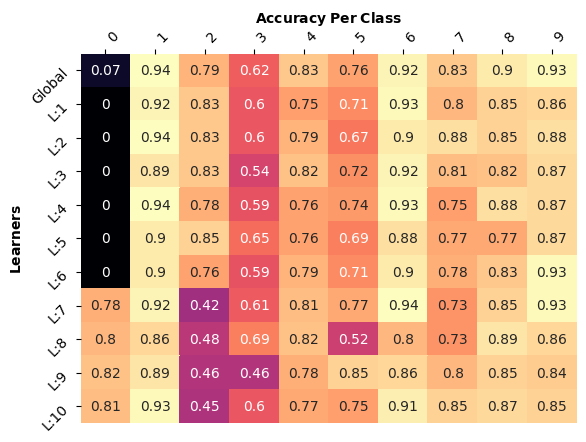}
    \label{subfig:cifar10_uniformiid_targetedlabelflipping_macroaccuracy_performanceperclass_6learners}
  }
  \subfloat[DVW-GMean]{
  \centering
    \includegraphics[width=0.33\linewidth]{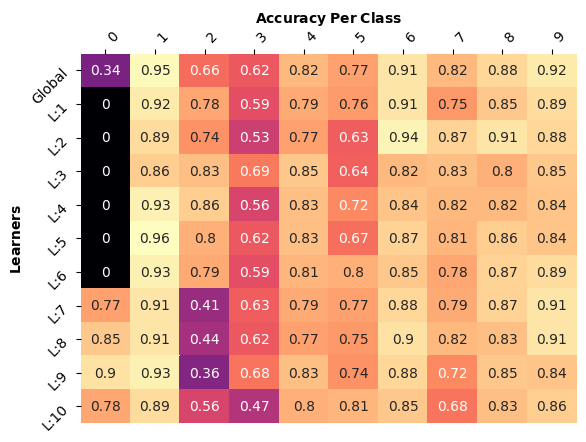}
    \label{subfig:cifar10_uniformiid_targetedlabelflipping_gmean_performanceperclass_6learners}
  }
  \caption{Accuracy per class in the last federation round for the community (global) model and each learner for the \textbf{Targeted Label Flipping} attack in the \textbf{Uniform \& IID} learning environment, with \textbf{6 corrupted learners.}}
  \label{fig:cifar10_uniformiid_targetedlabelflipping_performanceperclass_6learners}
\end{figure}

\begin{figure}[htpb]
  \subfloat[DVW-MicroAccuracy]{
  \centering
    \includegraphics[width=0.33\linewidth]{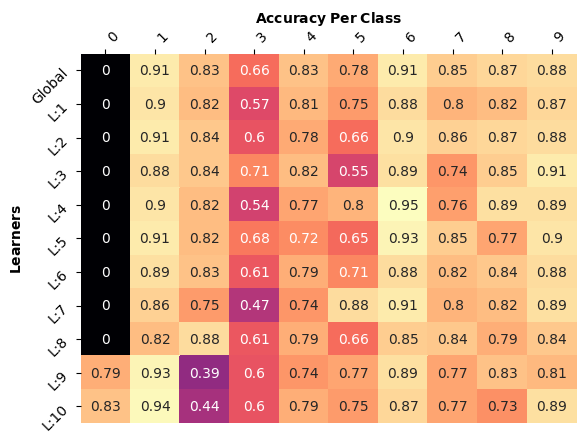}
    \label{subfig:cifar10_uniformiid_targetedlabelflipping_microaccuracy_performanceperclass_8learners}
  }
  \subfloat[DVW-MacroAccuracy]{
  \centering
    \includegraphics[width=0.33\linewidth]{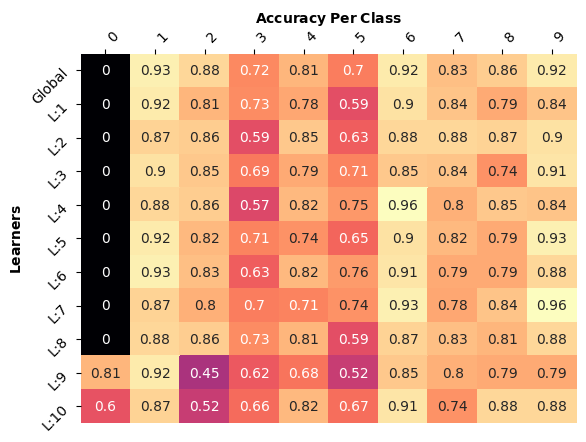}
    \label{subfig:cifar10_uniformiid_targetedlabelflipping_macroaccuracy_performanceperclass_8learners}
  }
  \subfloat[DVW-GMean]{
  \centering
    \includegraphics[width=0.33\linewidth]{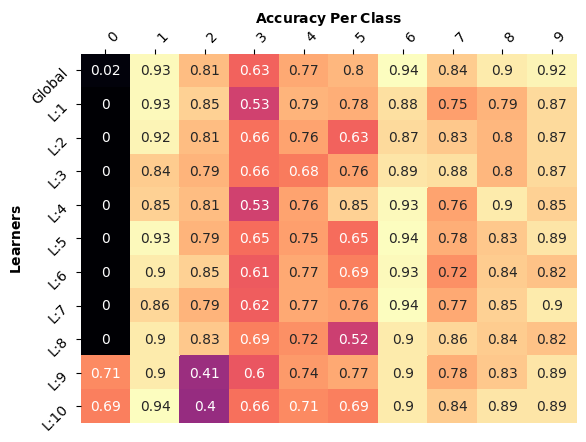}
    \label{subfig:cifar10_uniformiid_targetedlabelflipping_gmean_performanceperclass_8learners}
  }
  \caption{Accuracy per class in the last federation round for the community (global) model and each learner for the \textbf{Targeted Label Flipping} attack in the \textbf{Uniform \& IID} learning environment, with \textbf{8 corrupted learners.}}
  \label{fig:cifar10_uniformiid_targetedlabelflipping_performanceperclass_8learners}
\end{figure}

\newpage %

\subsection{PowerLaw \& IID} \label{appendix:targeted_label_flipping_powerlaw_iid}

\subsubsection{Federation Convergence} \label{appendix:targeted_label_flipping_powerlaw_iid_federation_convergence}
\begin{figure}[htpb]
\centering
  \subfloat[1 corrupted learner]{
    \includegraphics[width=0.45\linewidth]{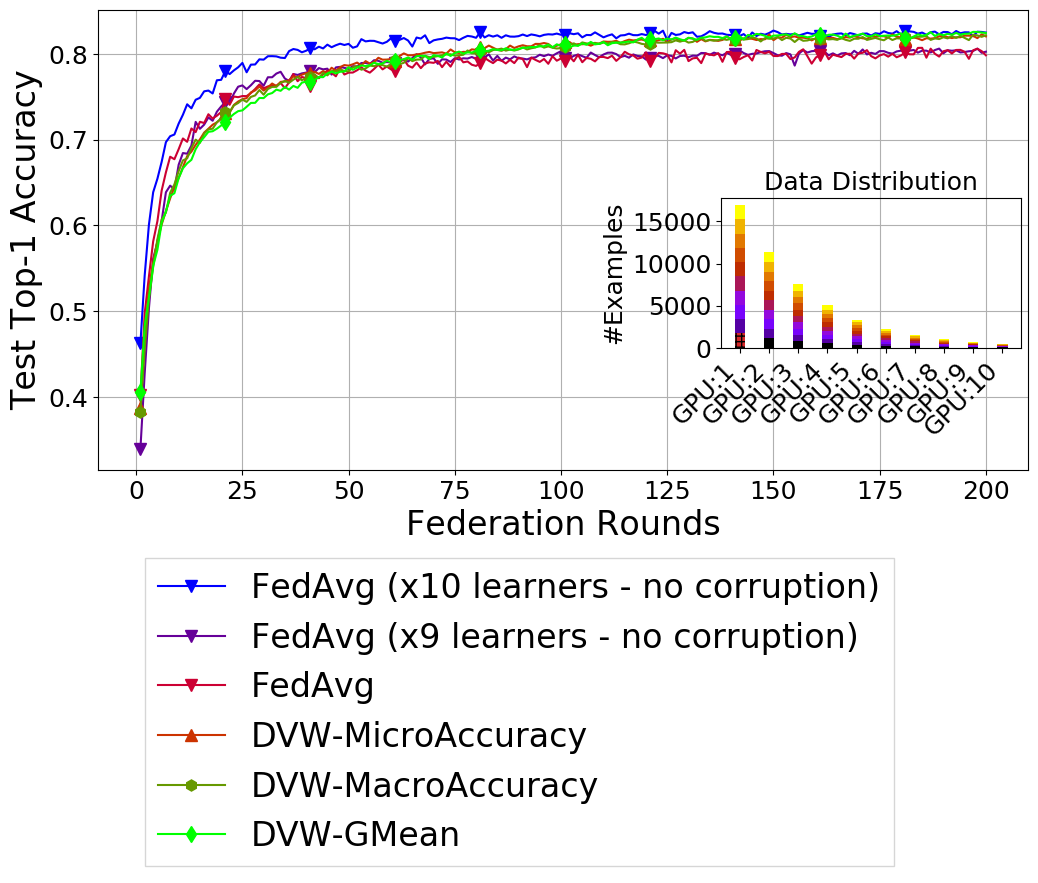}
    \label{subfig:cifar10_powerlawiid_targetedlabelflipping0to2_convergence_1learner}
  }
  \subfloat[3 corrupted learners]{
    \includegraphics[width=0.45\linewidth]{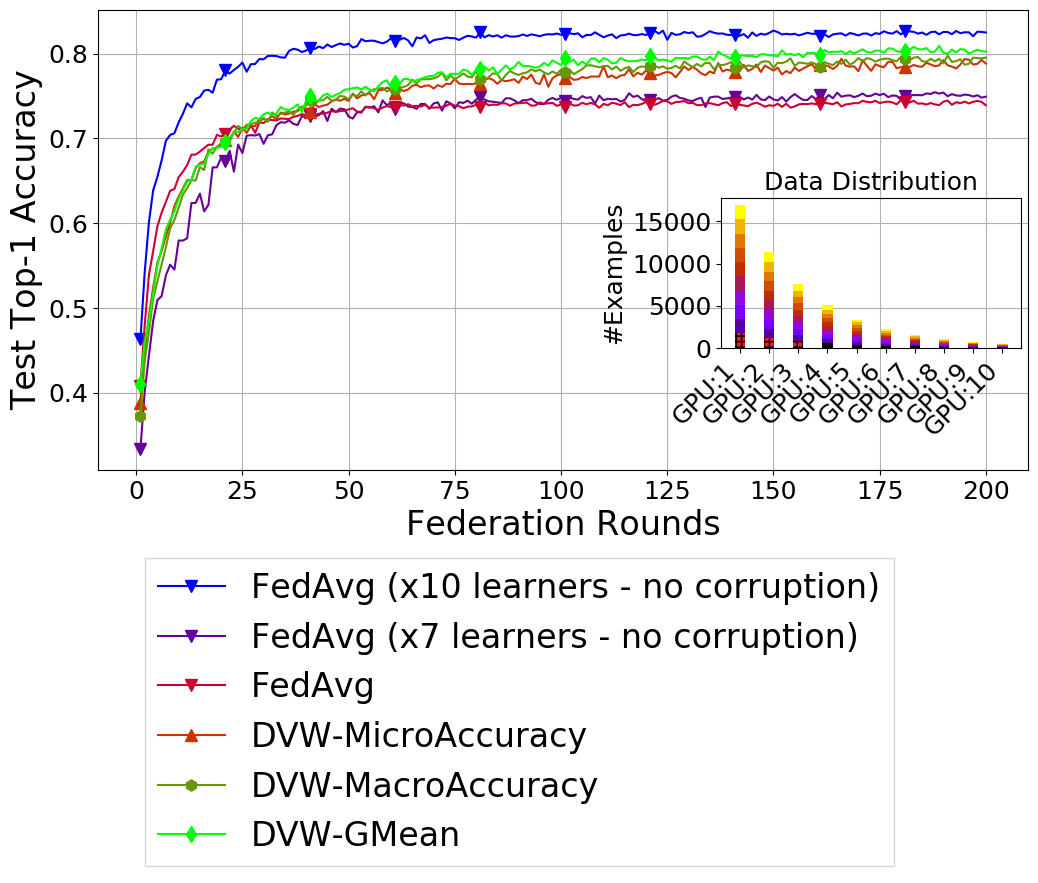}
    \label{subfig:cifar10_powerlawiid_targetedlabelflipping0to2_convergence_3learners}
  }
  
  \subfloat[5 corrupted learners]{
    \includegraphics[width=0.45\linewidth]{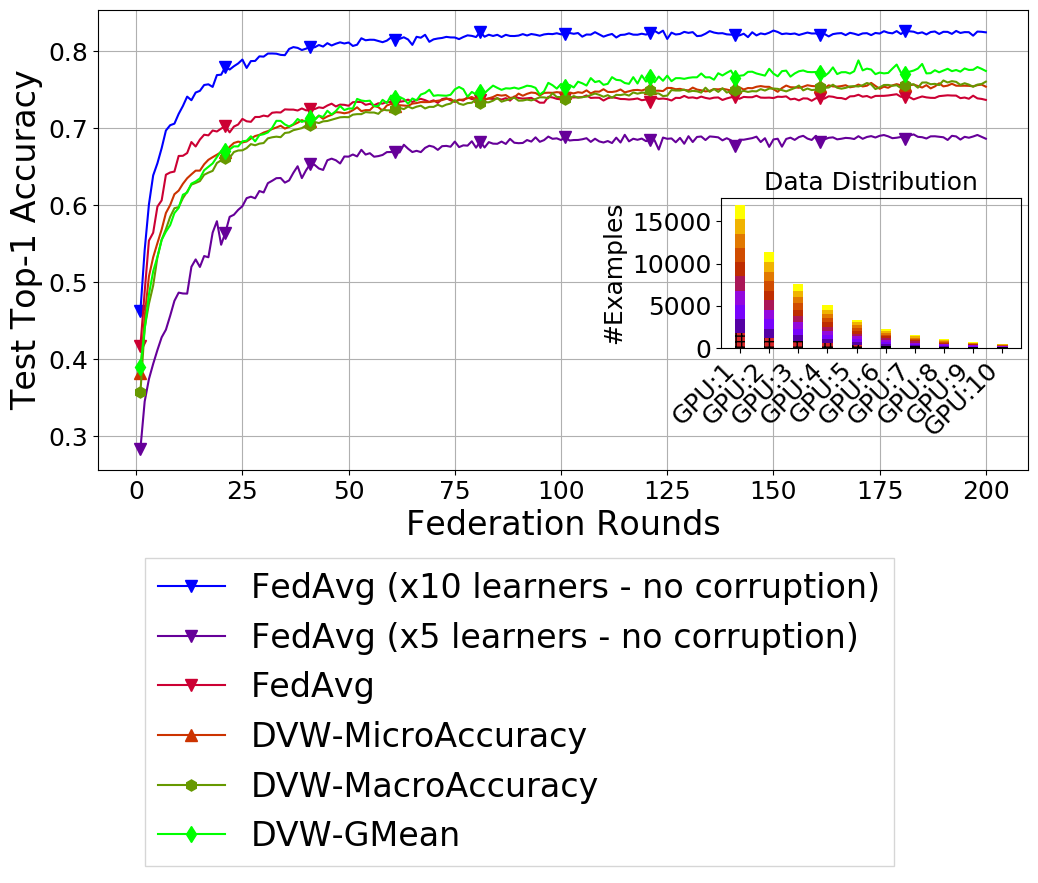}
    \label{subfig:cifar10_powerlawiid_targetedlabelflipping0to2_convergence_5learners}
  }
  
  \caption{\textbf{Federation convergence for the Targeted Label Flipping  data poisoning attack in the PowerLaw \& IID learning environment.} Source class is airplane and target class is bird. Federation performance is measured on the test top-1 accuracy over an increasing number of corrupted learners. Corrupted learners' flipped examples are marked with the red hatch within the data distribution inset. For every learning environment, we include the convergence of the federation with no corruption (10 honest learners) and with exclusion of the corrupted learners (x honest learners). We also present the convergence of the federation for FedAvg (baseline) and different performance weighting aggregation schemes, Micro-Accuracy, Macro-Accuracy and Geometric-Mean.}
  \label{fig:cifar10_powerlawiid_targetedlabelflipping0to2_federation_convergence}
\end{figure}

Figure \ref{fig:cifar10_powerlawiid_targetedlabelflipping0to2_federation_convergence} demonstrates the federation convergence rate for the targeted label flipping in the powerlaw and iid learning environments. FedAvg with and without corruption does not perform equally well compared to the performance aggregation schemes. Due to the diminishing amounts of data assigned per learner in this particular learning environment, the federation without any corrupted learners cannot match the generalization power of the rest of the schemes. Interestingly, all performance weighting schemes outperform FedAvg with and without corruption and Geometric Mean is shown to provide higher generalization power to the federation compared to Micro- and Macro-Accuracy.

\newpage %

\subsubsection{Learners Contribution Value} \label{appendix:targeted_label_flipping_powerlaw_iid_learners_contribution_value}

\begin{figure}[htpb]
  \subfloat[DVW-MicroAccuracy]{
    \includegraphics[width=0.33\linewidth]{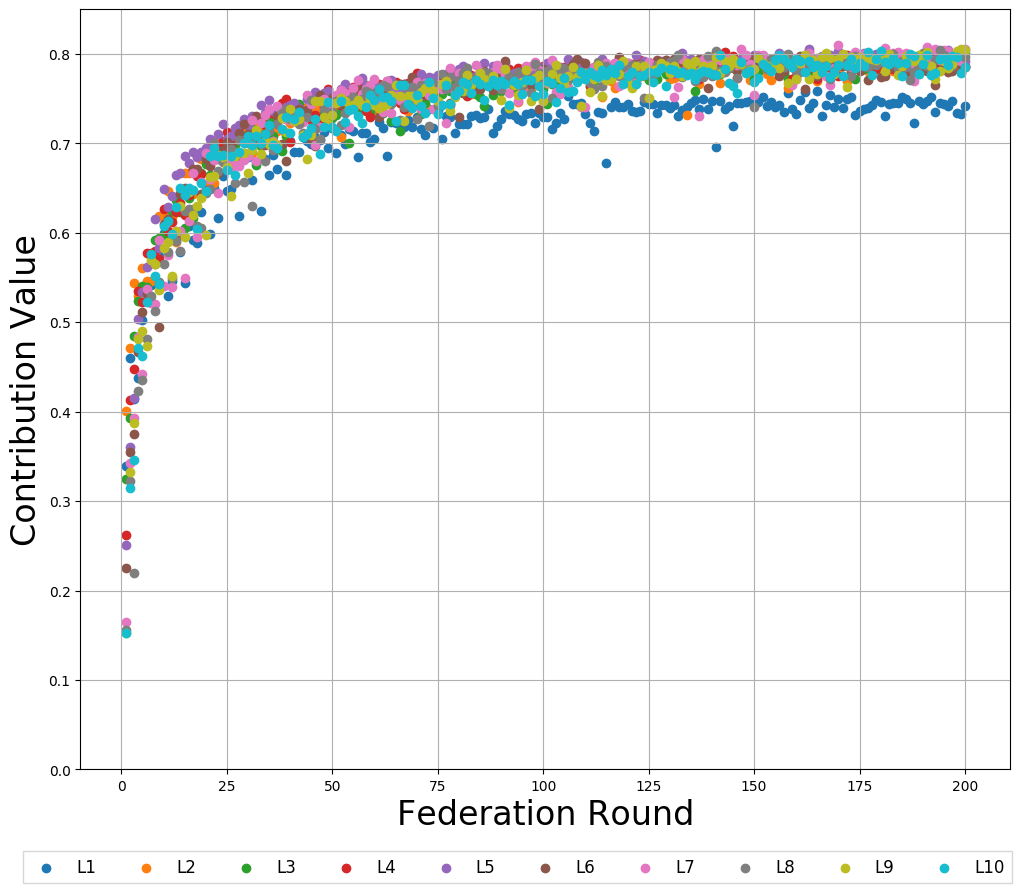}
    \label{subfig:cifar10_powerlawiid_targetedlabelflipping_microaccuracy_contributionvalue_1learner}
  }
  \subfloat[DVW-MacroAccuracy]{
    \includegraphics[width=0.33\linewidth]{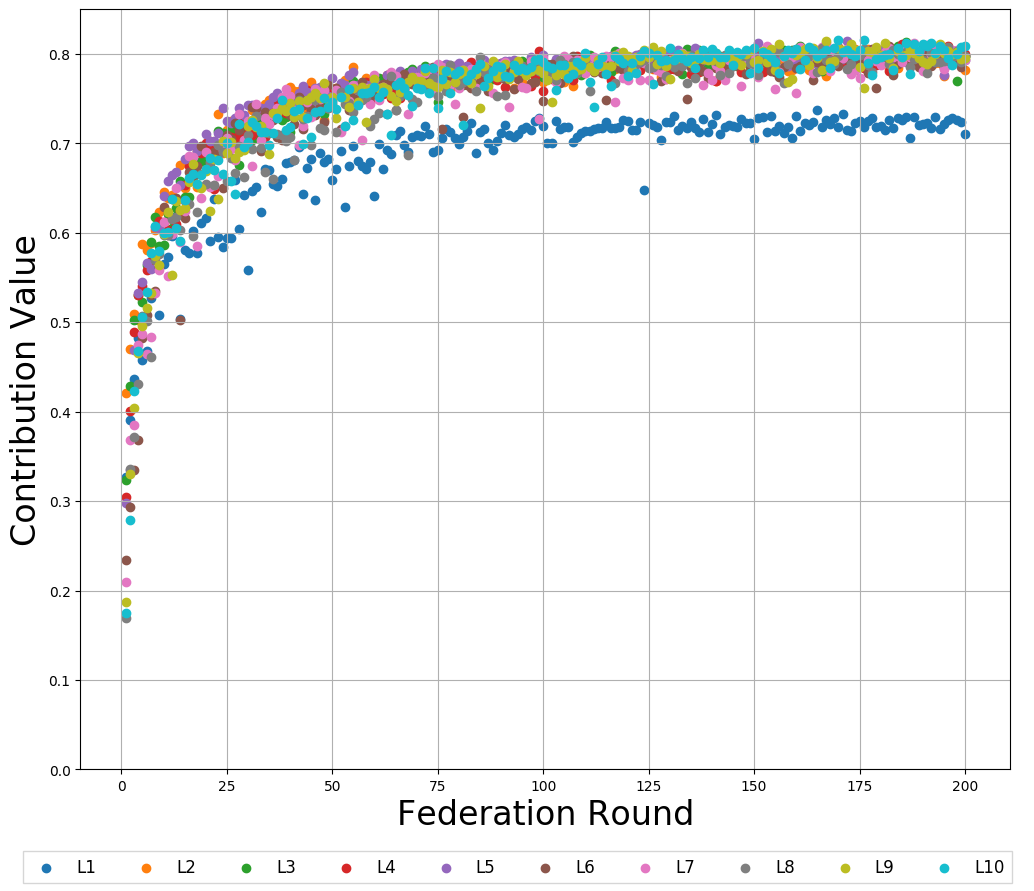}
    \label{subfig:cifar10_powerlawiid_targetedlabelflipping_macroaccuracy_contributionvalue_1learner}
  }
  \subfloat[DVW-GMean]{
    \includegraphics[width=0.33\linewidth]{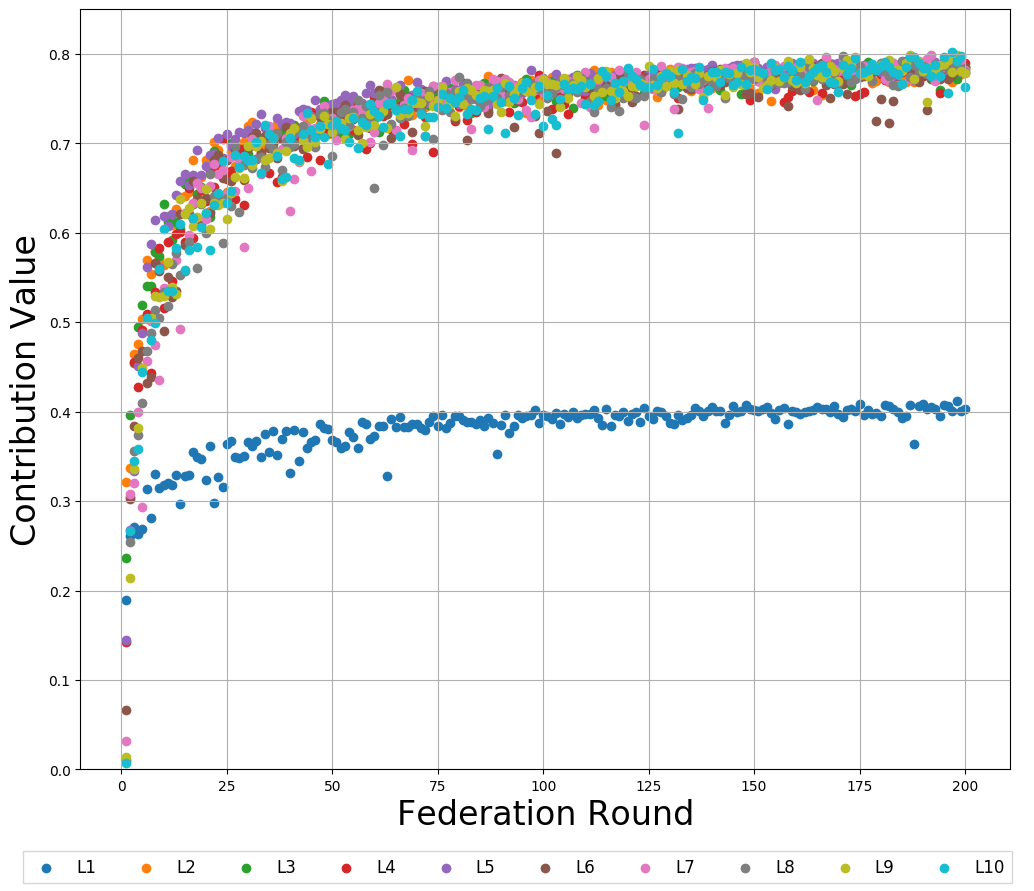}
    \label{subfig:cifar10_powerlawiid_targetedlabelflipping_gmean_contributionvalue_1learner}
  }
  \caption{Learners contribution value for the \textbf{Targeted Label Flipping} attack in the \textbf{PowerLaw \& IID} learning environment, \textbf{with 1 corrupted learner.}}
  \label{fig:cifar10_powerlawiid_targetedlabelflipping_contributionvalue_1learner}
\end{figure}

\begin{figure}[htpb]
  \subfloat[DVW-MicroAccuracy]{
    \includegraphics[width=0.33\linewidth]{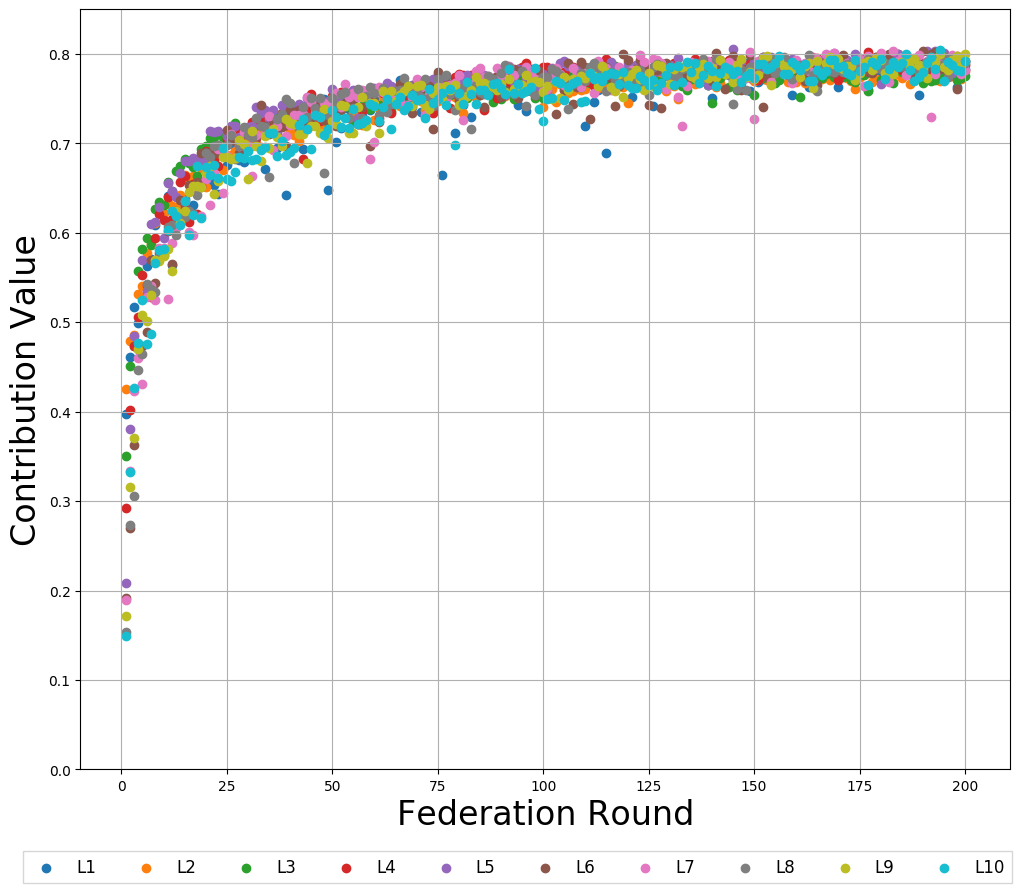}
    \label{subfig:cifar10_powerlawiid_targetedlabelflipping_microaccuracy_contributionvalue_3learners}
  }
  \subfloat[DVW-MacroAccuracy]{
    \includegraphics[width=0.33\linewidth]{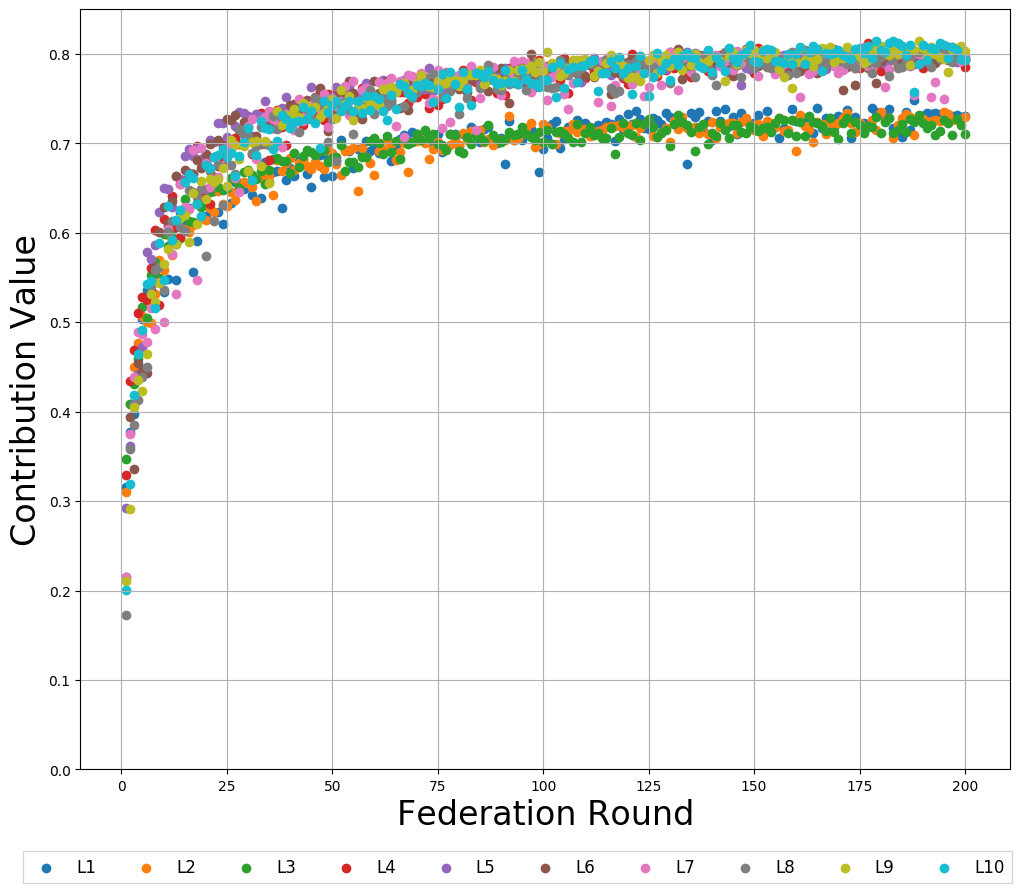}
    \label{subfig:cifar10_powerlawiid_targetedlabelflipping_macroaccuracy_contributionvalue_3learners}
  }
  \subfloat[DVW-GMean]{
    \includegraphics[width=0.33\linewidth]{plots/TargetedlabelFlipping/PoliciesContributionValueByLearner/Cifar10_PowerLawIID_TargetedLabelFlipping0to2_3LearnersL1L2L3_LearnersContributionValue_DVWGMean0001.png}
    \label{subfig:cifar10_powerlawiid_targetedlabelflipping_gmean_contributionvalue_3learners}
  }
  \caption{Learners contribution value for the \textbf{Targeted Label Flipping} attack in the \textbf{PowerLaw \& IID} learning environment, \textbf{with 3 corrupted learners.}}
  \label{fig:cifar10_powerlawiid_targetedlabelflipping_contributionvalue_3learners}
\end{figure}

\begin{figure}[htpb]
  \subfloat[DVW-MicroAccuracy]{
    \includegraphics[width=0.33\linewidth]{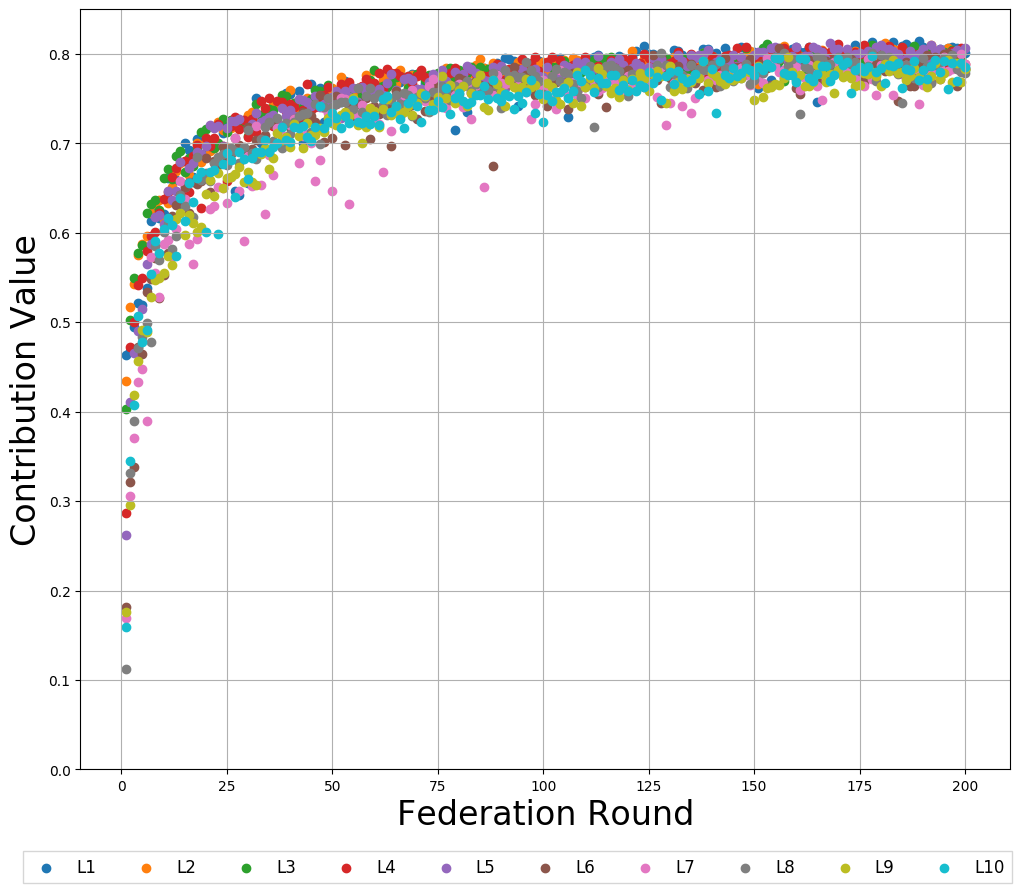}
    \label{subfig:cifar10_powerlawiid_targetedlabelflipping_microaccuracy_contributionvalue_5learners}
  }
  \subfloat[DVW-MacroAccuracy]{
    \includegraphics[width=0.33\linewidth]{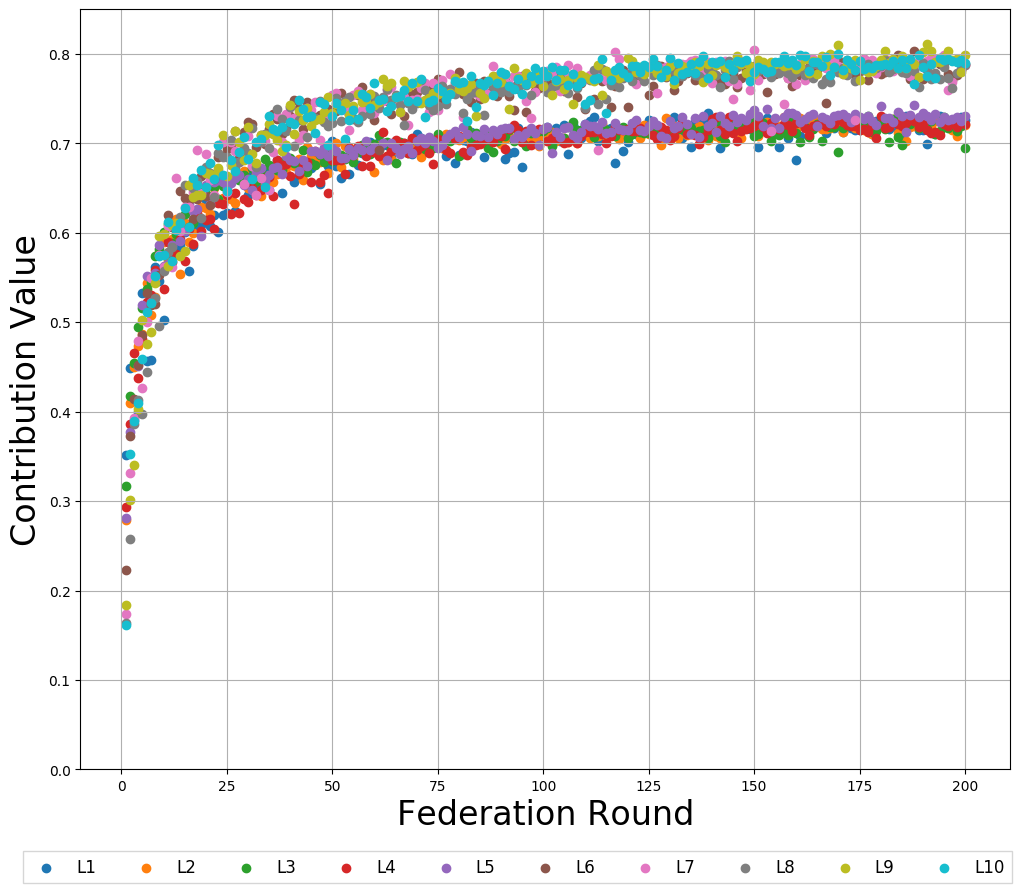}
    \label{subfig:cifar10_powerlawiid_targetedlabelflipping_macroaccuracy_contributionvalue_5learners}
  }
  \subfloat[DVW-GMean]{
    \includegraphics[width=0.33\linewidth]{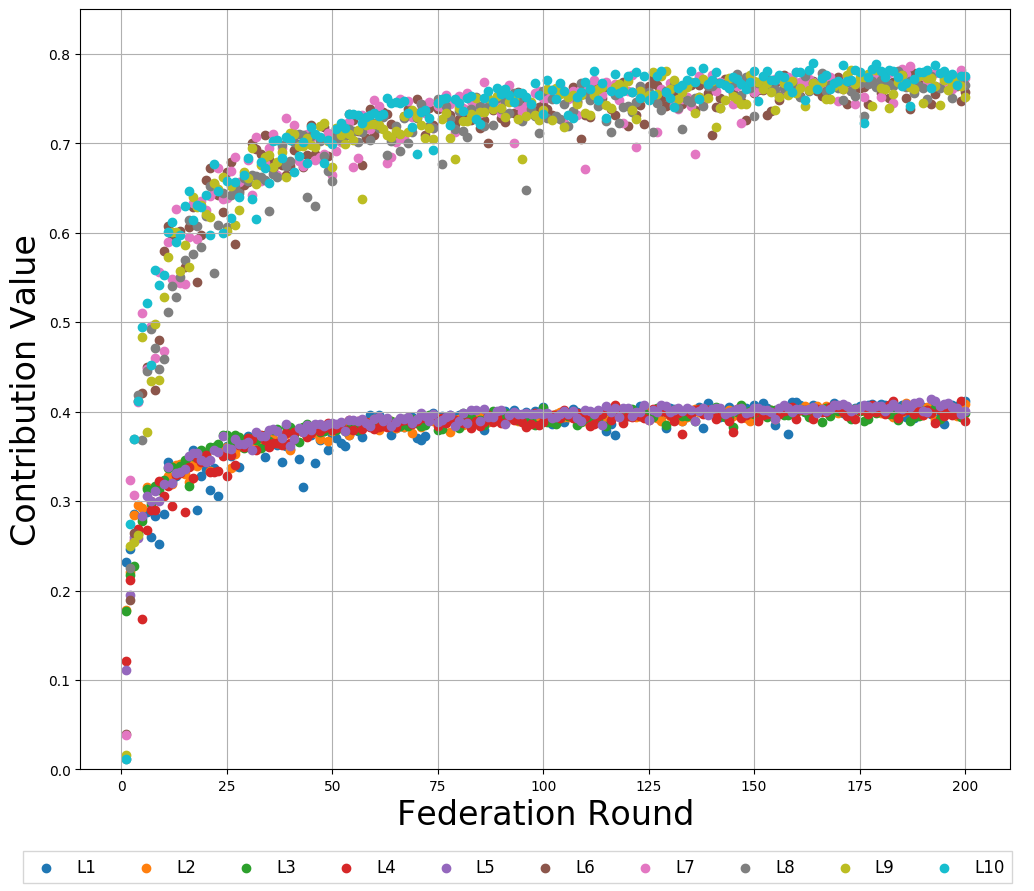}
    \label{subfig:cifar10_powerlawiid_targetedlabelflipping_gmean_contributionvalue_5learners}
  }
  \caption{Learners contribution value for the \textbf{Targeted Label Flipping} attack in the \textbf{PowerLaw \& IID} learning environment, \textbf{with 5 corrupted learners.}}
  \label{fig:cifar10_powerlawiid_targetedlabelflipping_contributionvalue_5learners}
\end{figure}

In \cref{fig:cifar10_powerlawiid_targetedlabelflipping_contributionvalue_1learner,fig:cifar10_powerlawiid_targetedlabelflipping_contributionvalue_3learners,fig:cifar10_powerlawiid_targetedlabelflipping_contributionvalue_5learners} we demonstrate the contribution/weighting value of each learner in the federation for every learning environment for the three different performance metrics. As it was also observed in the uniform and iid learning environments, compared to Macro-Accuracy, the Micro-Accuracy scheme cannot detect corrupted from non-corrupted learners and always considers them in the global model with an equally weighted contribution value. However, Geometric Mean is shown to have a clear advantage and penalize more the corrupted learners.

\newpage %

\subsubsection{Learners Performance Per Class} \label{appendix:targeted_label_flipping_powerlawiid_learners_performance_per_class}
\begin{figure}[htpb]

  \subfloat[DVW-MicroAccuracy]{
    \includegraphics[width=0.33\linewidth]{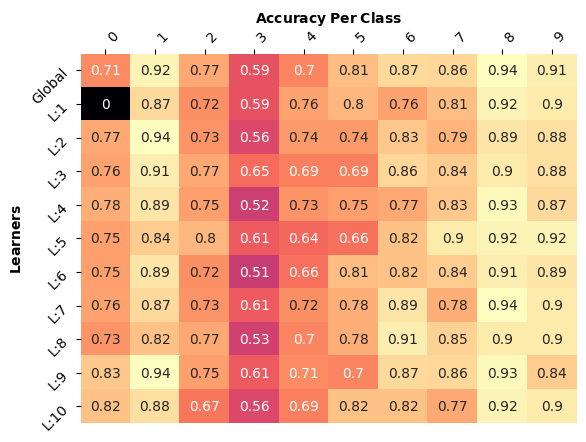}
    \label{subfig:cifar10_powerlawiid_targetedlabelflipping_microaccuracy_performanceperclass_1learner}
  }
  \subfloat[DVW-MacroAccuracy]{
    \includegraphics[width=0.33\linewidth]{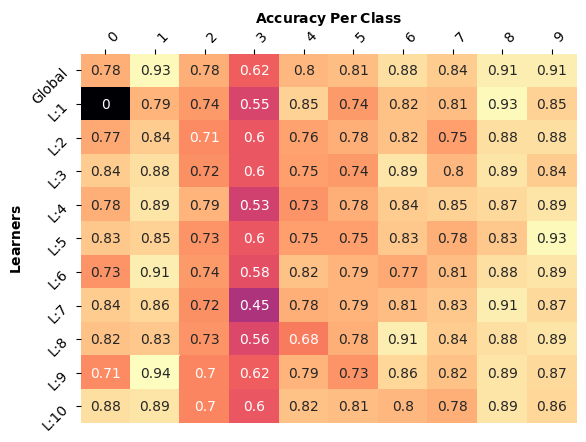}
    \label{subfig:cifar10_powerlawiid_targetedlabelflipping_macroaccuracy_performanceperclass_1learner}
  }
  \subfloat[DVW-GMean]{
    \includegraphics[width=0.33\linewidth]{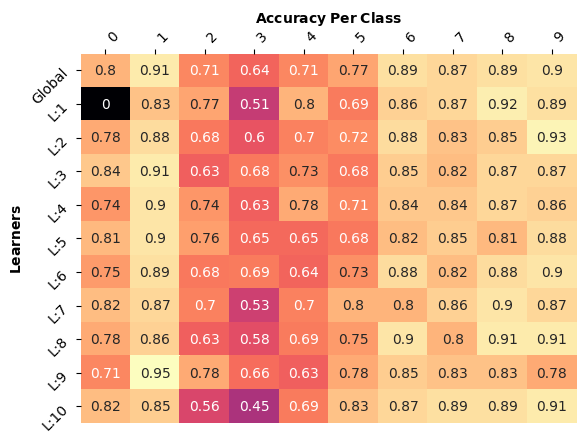}
    \label{subfig:cifar10_powerlawiid_targetedlabelflipping_gmean_performanceperclass_1learner}
  }
  \caption{Accuracy per class in the last federation round for the community (global) model and each learner for the \textbf{Targeted Label Flipping} attack in the \textbf{PowerLaw \& IID} learning environment, with \textbf{1 corrupted learner.}}
  \label{fig:cifar10_powerlawiid_targetedlabelflipping_performanceperclass_1learner}
\end{figure}

\begin{figure}[htpb]  
  \subfloat[DVW-MicroAccuracy]{
    \includegraphics[width=0.33\linewidth]{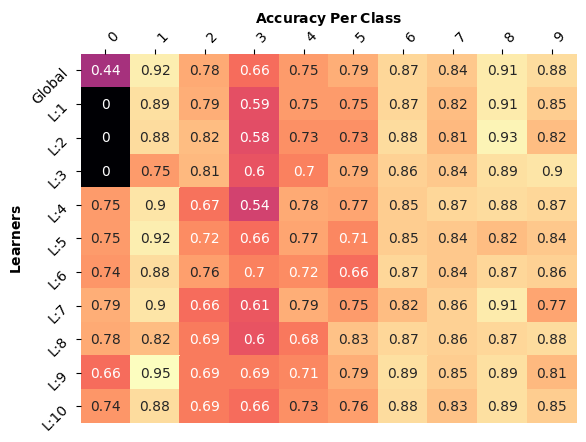}
    \label{subfig:cifar10_powerlawiid_targetedlabelflipping_microaccuracy_performanceperclass_3learners}
  }
  \subfloat[DVW-MacroAccuracy]{
  \centering
    \includegraphics[width=0.33\linewidth]{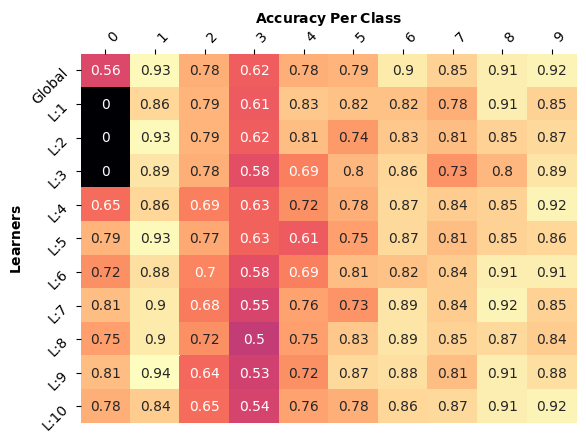}
    \label{subfig:cifar10_powerlawiid_targetedlabelflipping_macroaccuracy_performanceperclass_3learners}
  }
  \subfloat[DVW-GMean]{
  \centering
    \includegraphics[width=0.33\linewidth]{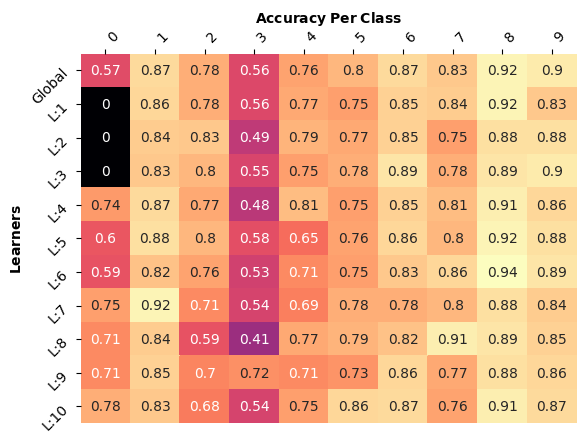}
    \label{subfig:cifar10_powerlawiid_targetedlabelflipping_gmean_performanceperclass_3learners}
  }
  \caption{Accuracy per class in the last federation round for the community (global) model and each learner for the \textbf{Targeted Label Flipping} attack in the \textbf{PowerLaw \& IID} learning environment, with \textbf{3 corrupted learners.}}
  \label{fig:cifar10_powerlawiid_targetedlabelflipping_performanceperclass_3learners}
\end{figure}

\begin{figure}[htpb]
  \subfloat[DVW-MicroAccuracy]{
  \centering
    \includegraphics[width=0.33\linewidth]{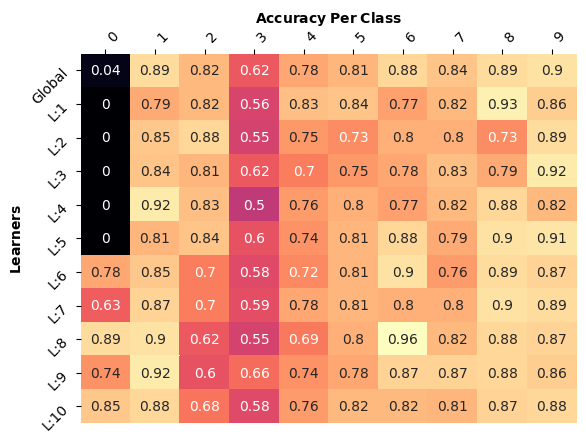}
    \label{subfig:cifar10_powerlawiid_targetedlabelflipping_microaccuracy_performanceperclass_5learners}
  }
  \subfloat[DVW-MacroAccuracy]{
  \centering
    \includegraphics[width=0.33\linewidth]{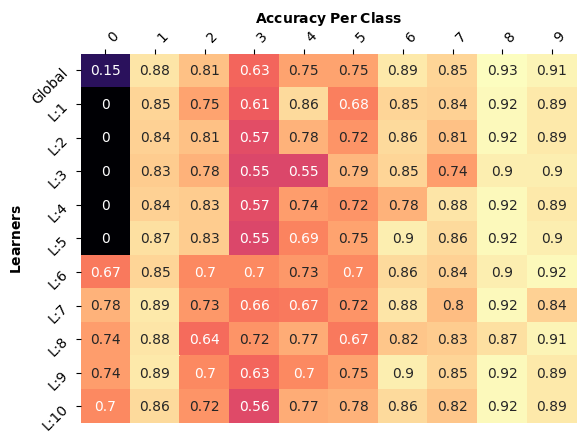}
    \label{subfig:cifar10_powerlawiid_targetedlabelflipping_macroaccuracy_performanceperclass_5learners}
  }
  \subfloat[DVW-GMean]{
  \centering
    \includegraphics[width=0.33\linewidth]{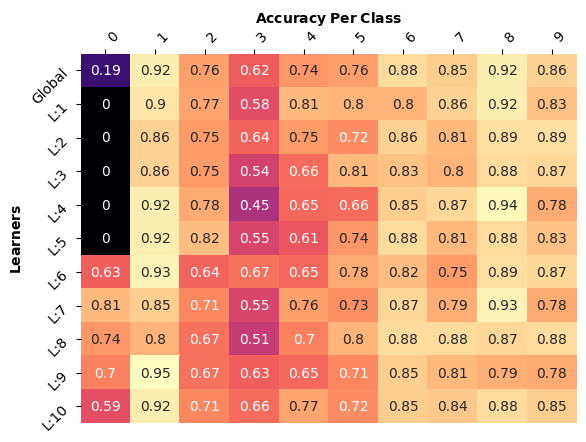}
    \label{subfig:cifar10_powerlawiid_targetedlabelflipping_gmean_performanceperclass_5learners}
  }
  \caption{Accuracy per class in the last federation round for the community (global) model and each learner for the \textbf{Targeted Label Flipping} attack in the \textbf{PowerLaw \& IID} learning environment, with \textbf{5 corrupted learners.}}
  \label{fig:cifar10_powerlawiid_targetedlabelflipping_performanceperclass_5learners}
\end{figure}

In \cref{fig:cifar10_powerlawiid_targetedlabelflipping_performanceperclass_1learner,fig:cifar10_powerlawiid_targetedlabelflipping_performanceperclass_3learners,fig:cifar10_powerlawiid_targetedlabelflipping_performanceperclass_5learners} we present the per-community (global) and per-learner model accuracy for every class of the distributed validation dataset at the final federation round. The higher the corruption is the harder it is for the community model to learn class 0.


\end{document}